\documentclass{article} % For LaTeX2e
\usepackage{iclr2025_conference,times}

\usepackage{amsmath,amsfonts,bm}

\def\eqref#1{equation~\ref{#1}}
\def\1{\bm{1}}

\DeclareMathAlphabet{\mathsfit}{\encodingdefault}{\sfdefault}{m}{sl}
\SetMathAlphabet{\mathsfit}{bold}{\encodingdefault}{\sfdefault}{bx}{n}

\usepackage{hyperref}
\usepackage{url}
\usepackage{graphicx} 
\usepackage{subcaption}
\usepackage{xcolor}
\usepackage{wrapfig}
\usepackage{booktabs}
\usepackage[misc]{ifsym}
\hypersetup{
colorlinks=true,
linkcolor=black,
citecolor=black
}

\title{Scaling Law with Learning Rate Annealing}

\author{Howe Tissue \small{\Letter}\\ 
\texttt{h-sun20@tsinghua.org.cn} \\
\And 
Venus Wang \\
\texttt{wangxxing12@gmail.com} \\
\And 
Lu Wang \\
\texttt{wangluloveslezhi@gmail.com} \\
}

\iclrfinalcopy % Uncomment for camera-ready version, but NOT for submission.
\begin{document}
\maketitle

\begin{abstract}
We find that the cross-entropy loss curves of neural language models empirically adhere to a scaling law with learning rate (LR) annealing over training steps:
$$L(s) = L_0 + A\cdot S_1^{-\alpha} - C\cdot S_2,$$
where $L(s)$ is the validation loss at step $s$, $S_1$ is the area under the LR curve, $S_2$ is the LR annealing area, and $L_0$, $A$, $C$, $\alpha$ are constant parameters.
This formulation takes into account two factors: (1) power-law scaling over data size, and (2) the additional loss reduction during LR annealing. 
Therefore, this formulation can describe the full loss curve at each step, rather than the single loss point at the end of training.
Applying the scaling law with LR annealing and fitting only one or two training curves, we can accurately predict the loss at any given step across any learning rate scheduler (LRS).
This approach significantly reduces computational cost in formulating scaling laws while providing more accuracy and expressiveness for training dynamics.
Extensive experiments demonstrate that our findings hold across a range of hyper-parameters and model architectures, and our equation can extend to scaling effect of model sizes.
Moreover, our formulation provides accurate theoretical verification and explanation for empirical results observed in numerous previous studies, particularly those focusing on LR schedule and annealing.
We believe that this work is promising to enhance the understanding of LLM training dynamics while greatly democratizing scaling laws, and it can guide researchers in refining training strategies (e.g. critical LRS) for further LLMs~\footnote{We welcome any feedback, comment, and discussion at \href{https://www.alphaxiv.org/abs/2408.11029}{AlphaXiv} or \href{https://huggingface.co/papers/2408.11029}{HuggingFace}, where we also provide our implementation codes.}.

\end{abstract}

\section{Introduction}
In recent years, large language models (LLMs) have garnered significant academic and industrial attention \citep{browngpt3_2020,touvron2023llama}. The scaling law suggests that the validation loss of language models follow a power-law pattern as model and data sizes increase~\citep{hestness2017deep,kaplan2020scaling,henighan2020scaling}. This law provides a powerful framework for forecasting LLM performances before large scale training by fitting losses at smaller scales \citep{achiam2023gpt,deepseek-ai2024deepseek,dubey2024llama}. Numerous studies have explored on the formulation to model the scaling effect of LLMs under various different settings~\citep{bahri2021explaining,hernandez2021scaling,caballero2022broken,Quantization-scaling,muennighoff2023scaling}.

However, typical scaling law formulations focus only on the final loss at the end of training~\citep{hoffmann2022training}. Specifically, previous approaches generally rely on a set of training runs and fit the scaling law curve solely on the final loss from each run. Essentially, the middle points with different degrees of LR annealing fail to follow typical scaling laws, which do not consider local loss drop brought by LR annealing. 
The previous approach under-utilizes the training compute and fails to capture the \emph{training dynamics} within each run. Further, the application of scaling laws in LLM developments is limited since the loss curve through the whole training process is not modeled. An expressive formulation that models full loss curves enables prediction of future training dynamics and also offers insights on understanding the learning process of LLMs.
%In other words, the data points used for fitting scaling laws and predicting are the losses incurred after the completion of learning rate scheduler (LRS). 
% Essentially, the middle loss points with different LR annealing degrees fail to follow typical scaling laws, which do not consider local loss drop brought by LR annealing.
% Consequently, these laws are unable to fit or predict a full loss curve. Till this work, we do not have an appropriate formulation that accurately describes the dynamics during the training process, which is crucial to deeply understand and improve the training process.

\begin{figure}[tbp]
\centering
\includegraphics[width=0.8\textwidth]{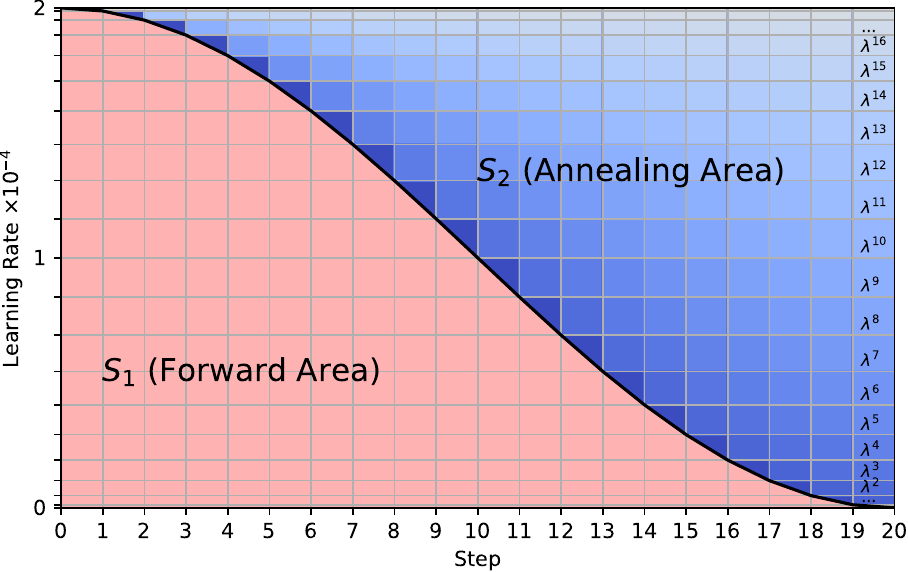}
\caption{Visualization of $S_1$ and $S_2$ at the 20-th step of a cosine LR scheduler. $S_1$ is the forward area, i.e., sum of red grid areas, which can be approximately regarded as the total amount of movement for neural network parameters; $S_2$ is the decayed annealing area, i.e., weighted sum of blue grid areas, where lighter shades indicate smaller weights.
%  (the lighter the color, the greater the degree of decay)
Both $S_1$ and $S_2$ contribute to loss reduction, and balancing their values is crucial for achieving the lowest possible final loss.
}
\label{fig:definition}
\vspace{-10pt}
\end{figure}

In this study, we propose a scaling law that models the full loss curve within a complete LLM training run. 
Specifically, we dive deeper into the training dynamics during LR annealing, and incorporate a LR annealing factor into the traditional scaling law formula to formulate the process.
This design is motivated by the observed correlation between LRS and loss curves, where loss gradually decreases as we consume more training steps~\footnote{In this paper, we use training steps to quantify the amount of consumed data, as they are typically proportional, with data amount calculated as training steps multiplied by batch size.} and then sharply declines when the LR undergoes significant annealing~\citep{loshchilov2016sgdr,dontdecay,simple-scalable-cpt2024,hu2024minicpm}. 
Fig.~\ref{fig:similarity} depicts how loss curves change over different learning rate schedules.
Overall, we discover that the validation loss of a language model at any step is determined by two factors: the forward area $S_1$ under the LR curve and the degree of LR annealing $S_2$ at that step. Formally, the expectation of loss $L$ at step $s$ of a language model follows::
\begin{equation}
\label{eq:scaling}
\begin{aligned}
L(s) &= L_0 + A\cdot S_1^{-\alpha} - C\cdot S_2, \\
S_1 &= \sum\limits_{i=1}^{s}\eta_i, \\
S_2 &= \sum\limits_{i=1}^{s}\sum\limits_{k=1}^{i}(\eta_{k-1} - \eta_{k})\cdot \lambda^{i-k},
\end{aligned}
\end{equation}
where $\eta_i$ is the learning rate at step $i$, and $\lambda$ is a hyper-parameter representing the decay factor for LR annealing momentum (see Sec.~\ref{sec:theorem} in detail),which typically ranges from $0.99$ to $0.999$. $L_0$, $A$, $C$, $\alpha$ are undetermined positive constants. $S_1$ is also known as the summed learning rate~\citep{kaplan2020scaling}, and $S_2$ represents the LR annealing area. A visualization of $S_1$ and $S_2$ is provided as Fig.~\ref{fig:definition}. 

\begin{figure}[tbp]
    \centering
    \begin{subfigure}[b]{0.32\textwidth}
        \includegraphics[width=\textwidth]{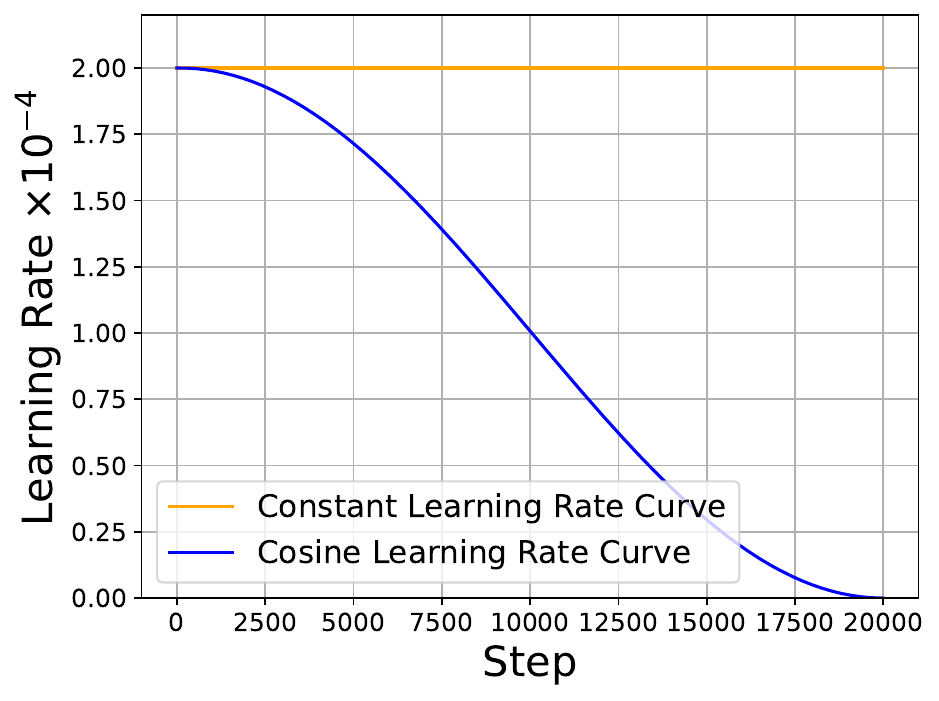}
        \caption{LR curves.}
    \end{subfigure}
    \hfill
    \begin{subfigure}[b]{0.32\textwidth}
        \includegraphics[width=\textwidth]{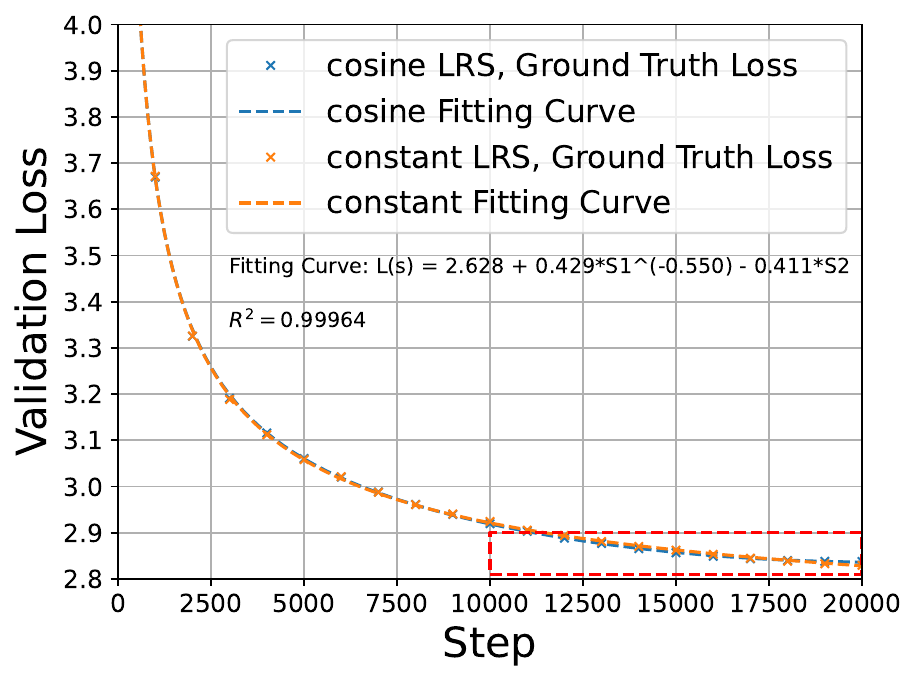}
        \caption{Loss curves.}
    \end{subfigure}
    \hfill
    \begin{subfigure}[b]{0.32\textwidth}
        \includegraphics[width=\textwidth]{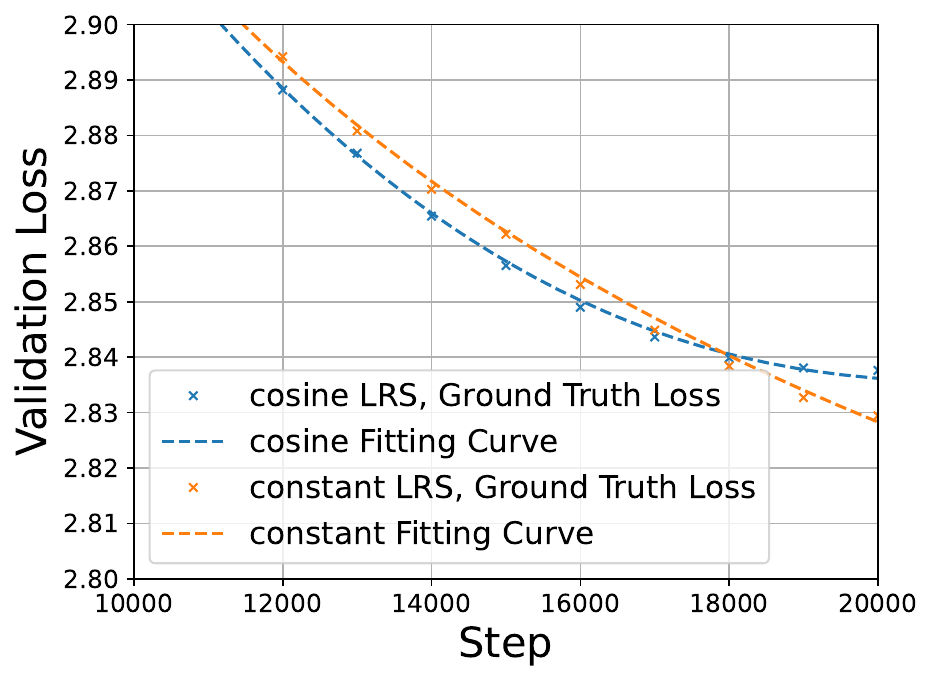}
        \caption{Zoomed-in view.}
    \end{subfigure}
    \caption{Using Eq.~\ref{eq:scaling} to \textbf{fit} full loss curves yield by constant and cosine LRS. Total steps = 20K, $\eta_{max}=2\times10^{-4}$, $\eta_{min}=0$. The fitted equation is $L(s) = 2.628 + 0.429\cdot S_1^{-0.550} - 0.411\cdot S_2$.
    }
\label{fig:fit}
\end{figure}

\begin{figure}[tbp]
    \centering
    
    \begin{subfigure}[b]{\textwidth}
        \centering
        \includegraphics[width=0.32\textwidth]{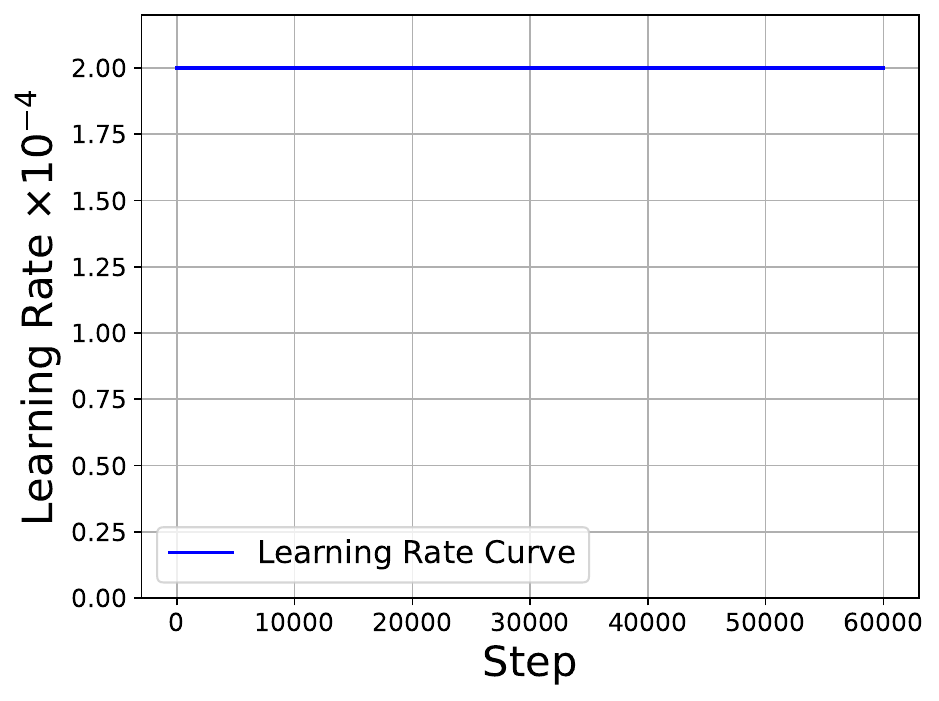}
        \includegraphics[width=0.32\textwidth]{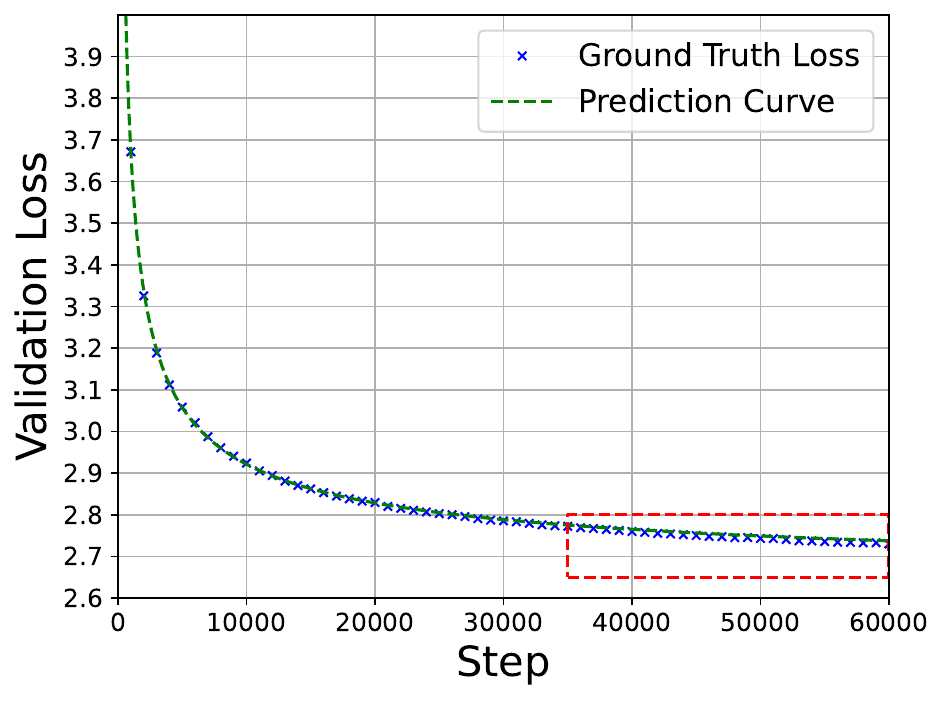}
        \includegraphics[width=0.32\textwidth]{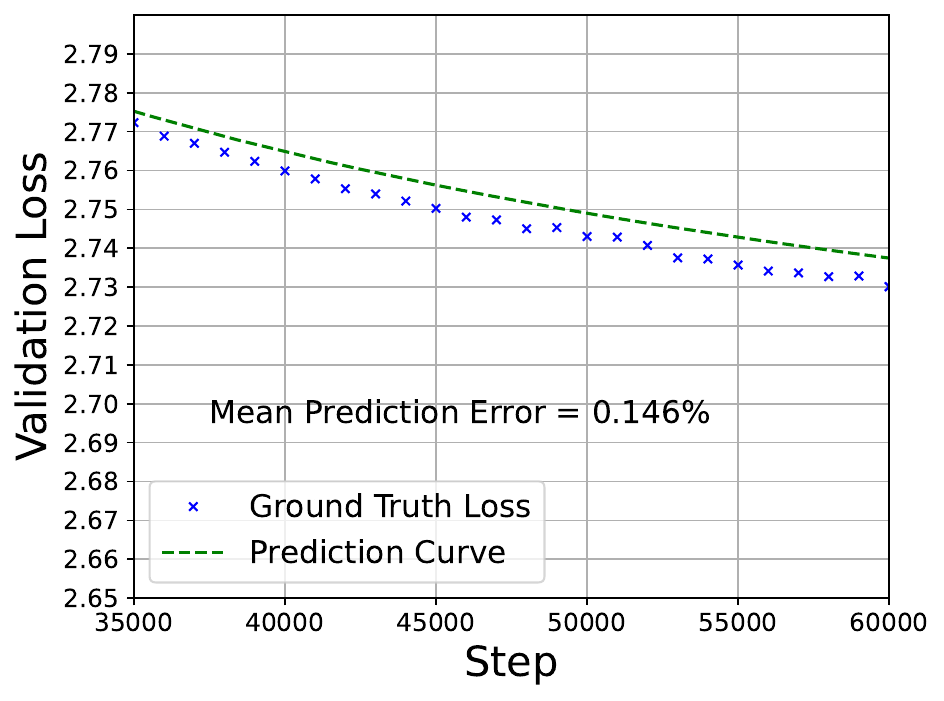}
        \caption{Full curve prediction of constant LRS.}
        \label{fig:prediction-constant}
    \end{subfigure}

    \begin{subfigure}[b]{\textwidth}
        \centering
        \includegraphics[width=0.32\textwidth]{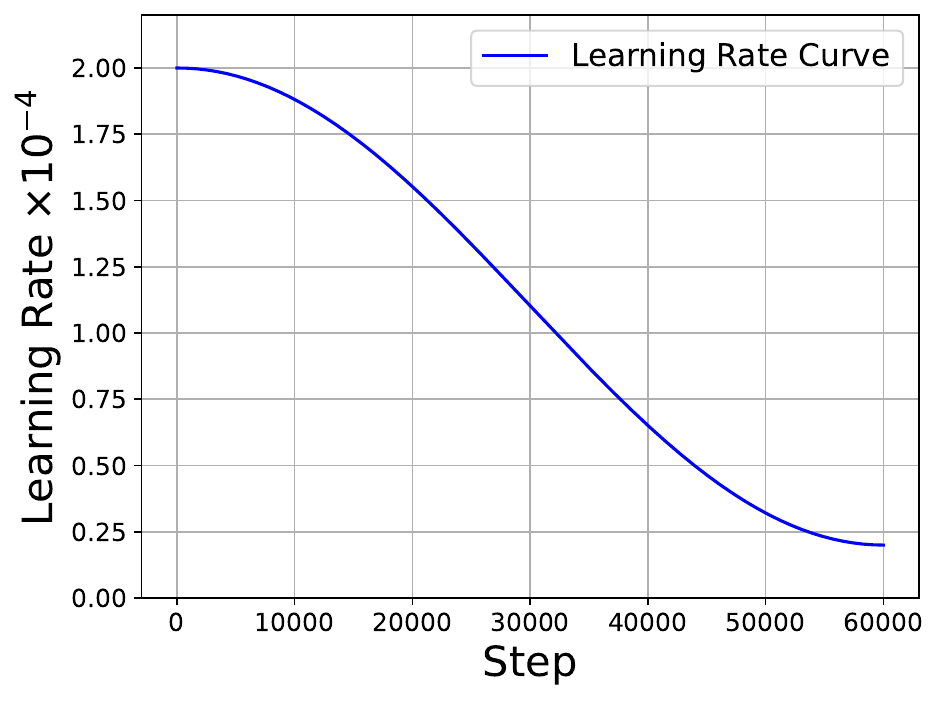}
        \includegraphics[width=0.32\textwidth]{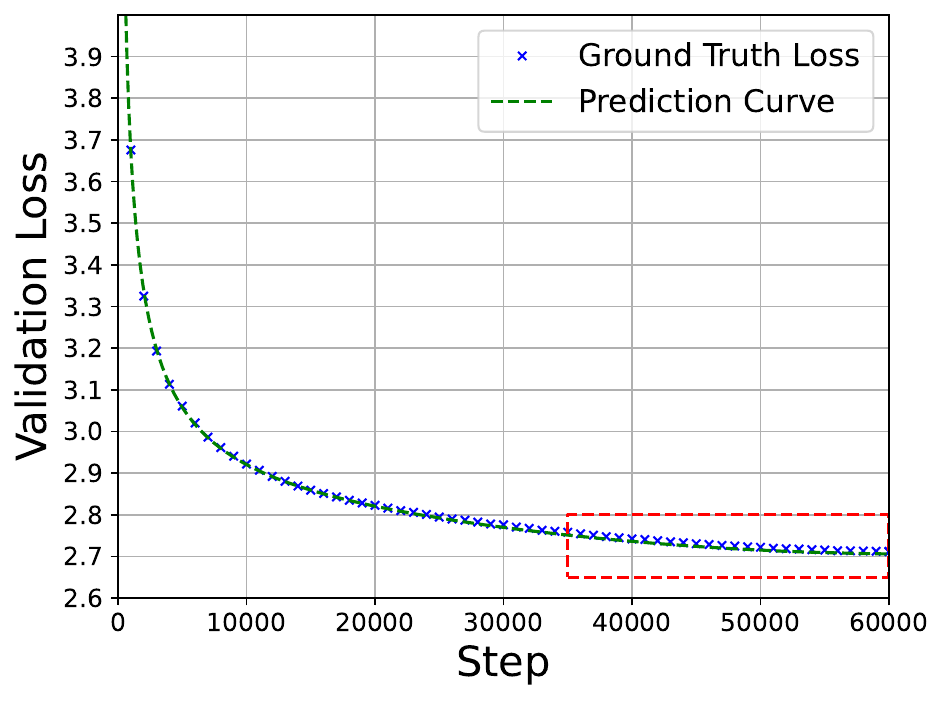}
        \includegraphics[width=0.32\textwidth]{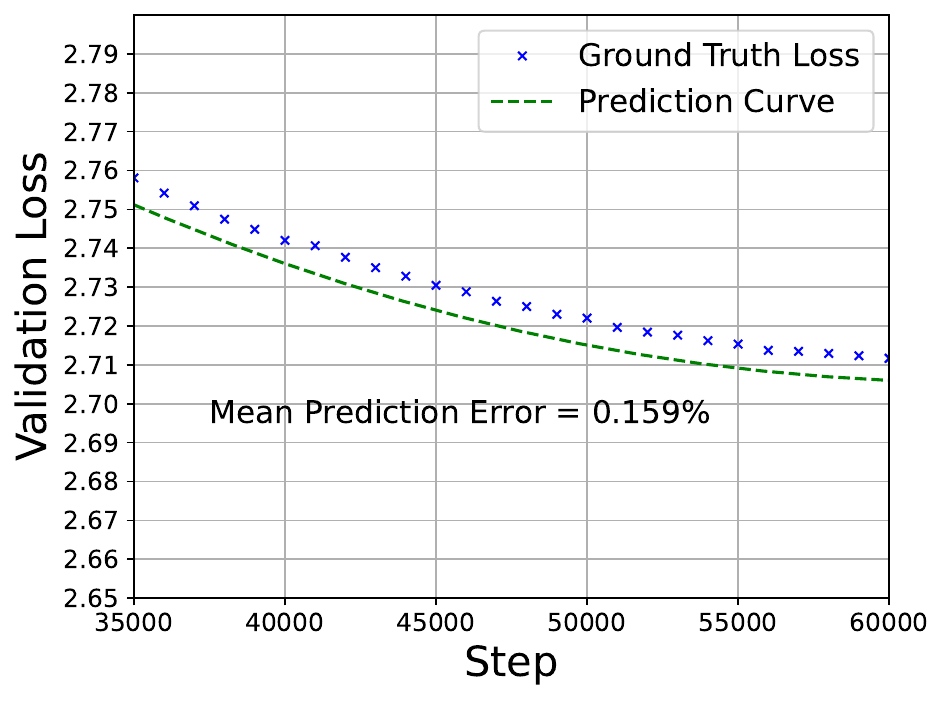}
        \caption{Full loss curve prediction of the cosine LRS (60K steps, $\eta_{min} = 0.1\cdot\eta_{max}$).}
        \label{fig:prediction-cosine}
    \end{subfigure}

    \begin{subfigure}[b]{\textwidth}
        \centering
        \includegraphics[width=0.32\textwidth]{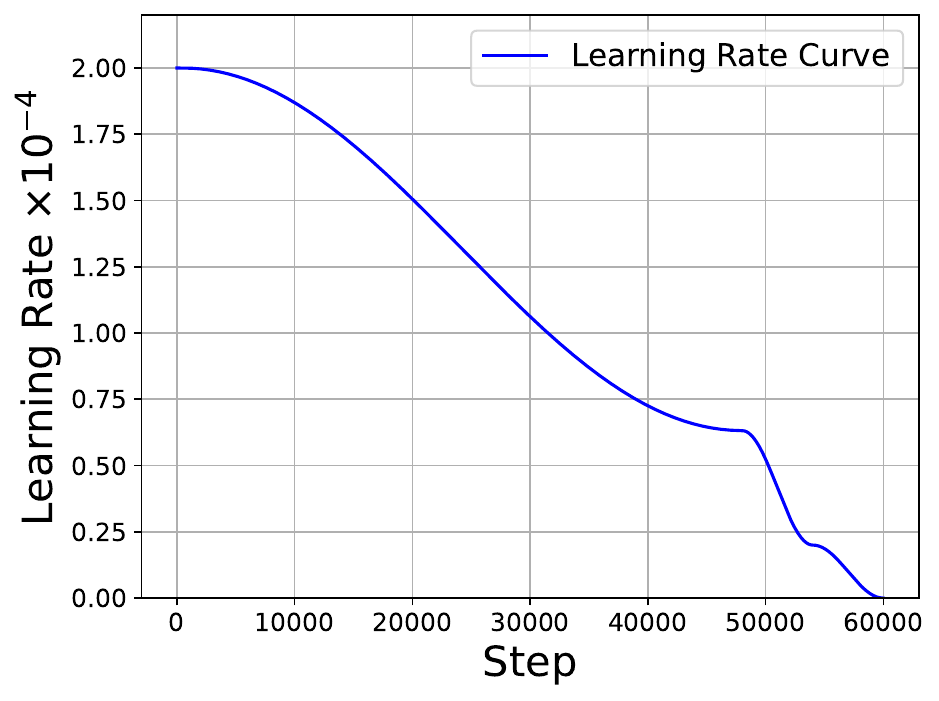}
        \includegraphics[width=0.32\textwidth]{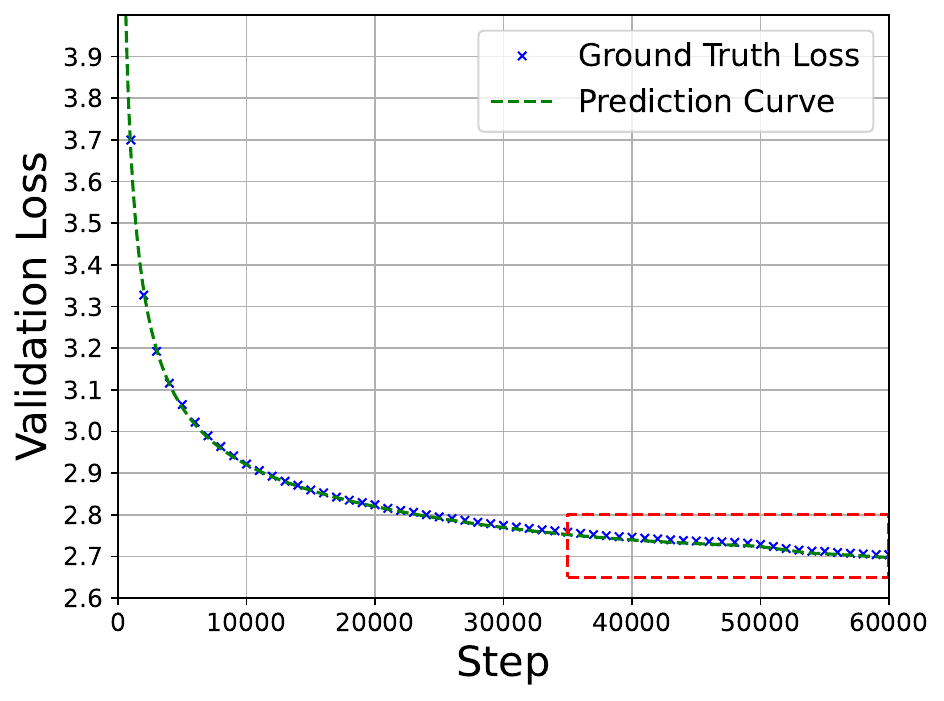}
        \includegraphics[width=0.32\textwidth]{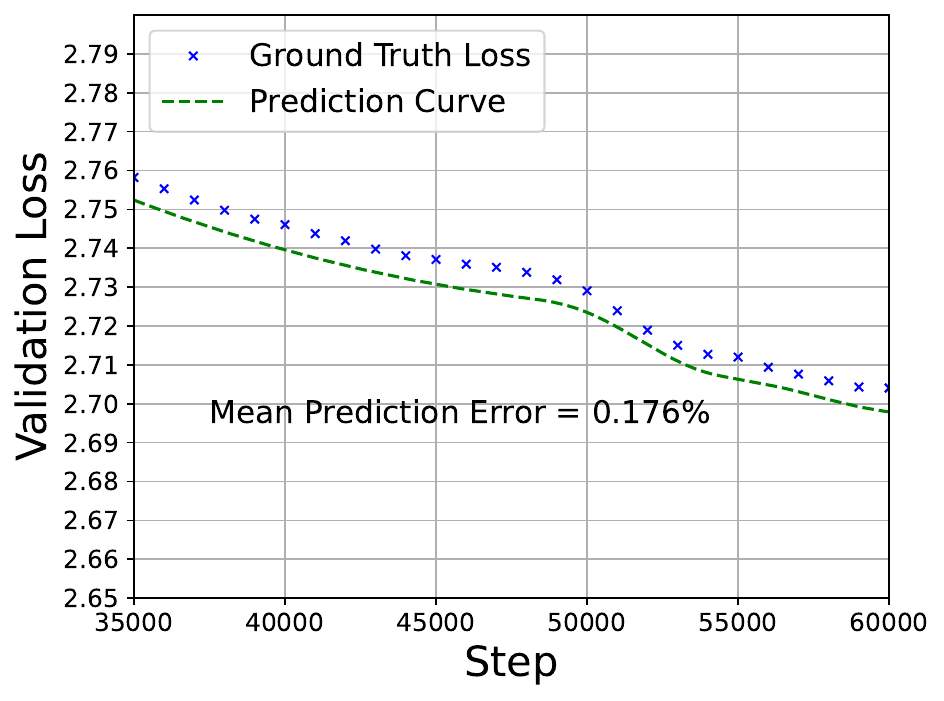}
        \caption{Full loss curve prediction of the multi-step cosine LRS (80\% + 10\% + 10\%)~\citep{deepseek-ai2024deepseek}.}
        \label{fig:prediction-multi-cosine}
    \end{subfigure}

    \begin{subfigure}[b]{\textwidth}
        \centering
        \includegraphics[width=0.32\textwidth]{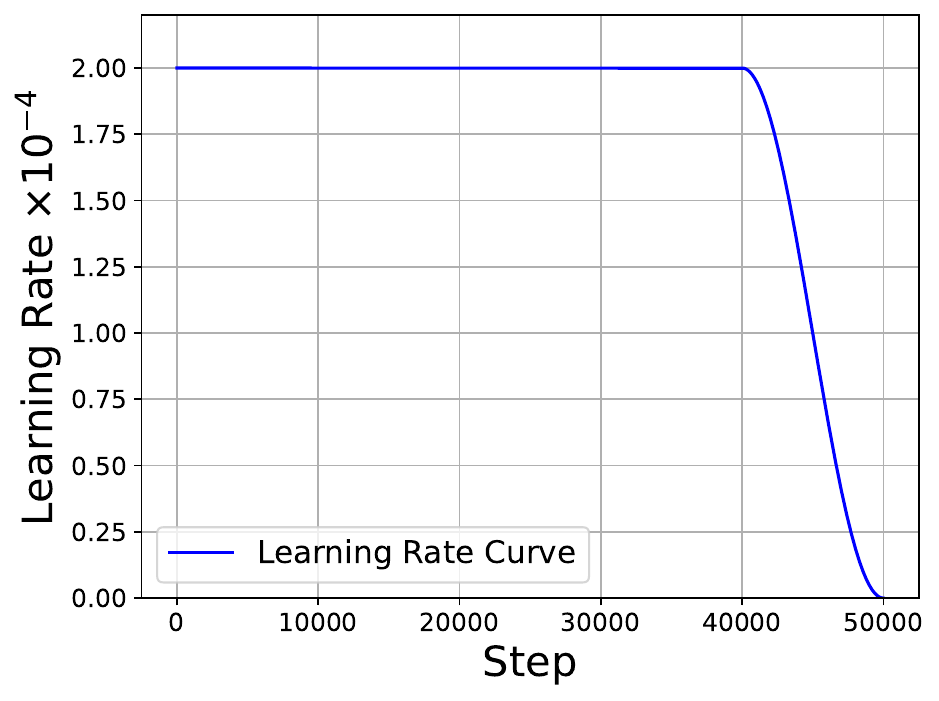}
        \includegraphics[width=0.32\textwidth]{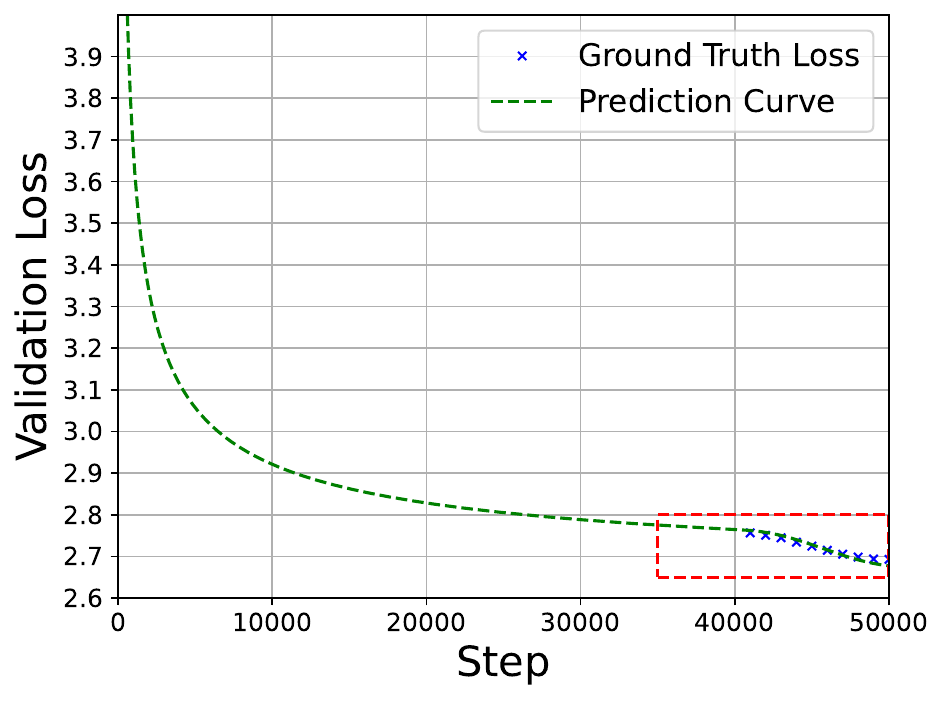}
        \includegraphics[width=0.32\textwidth]{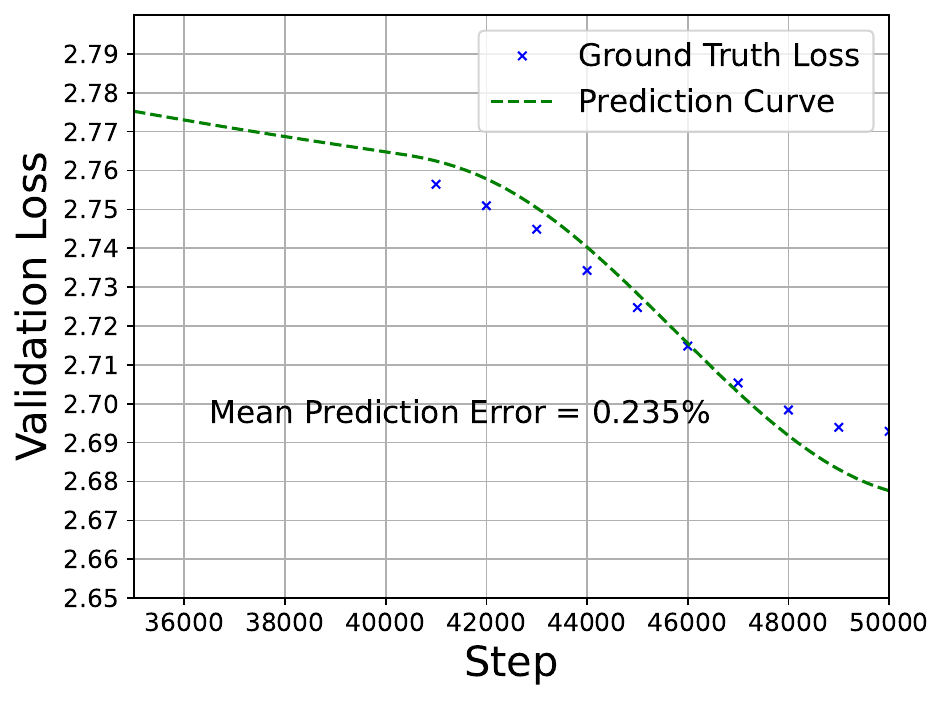}
        \caption{Full Loss curve prediction of the WSD LRS (20\% cosine annealing to $\eta_{min} = 0$)~\citep{hu2024minicpm}.}
        \label{fig:prediction-wsd}
    \end{subfigure}
     \begin{subfigure}[b]{\textwidth}
        \centering
        \includegraphics[width=0.32\textwidth]{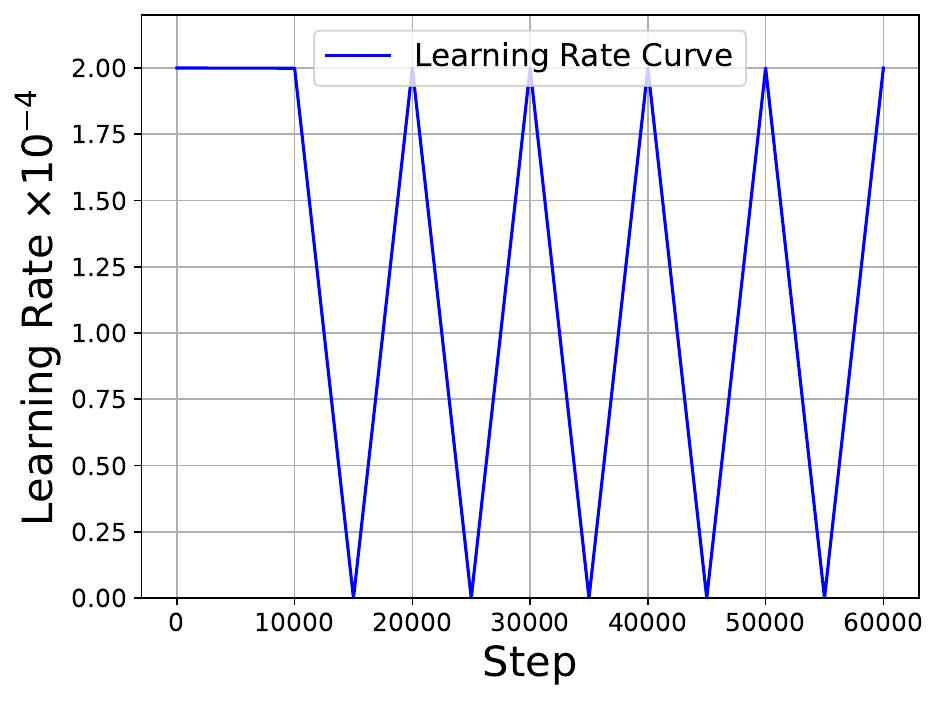}
        \includegraphics[width=0.32\textwidth]{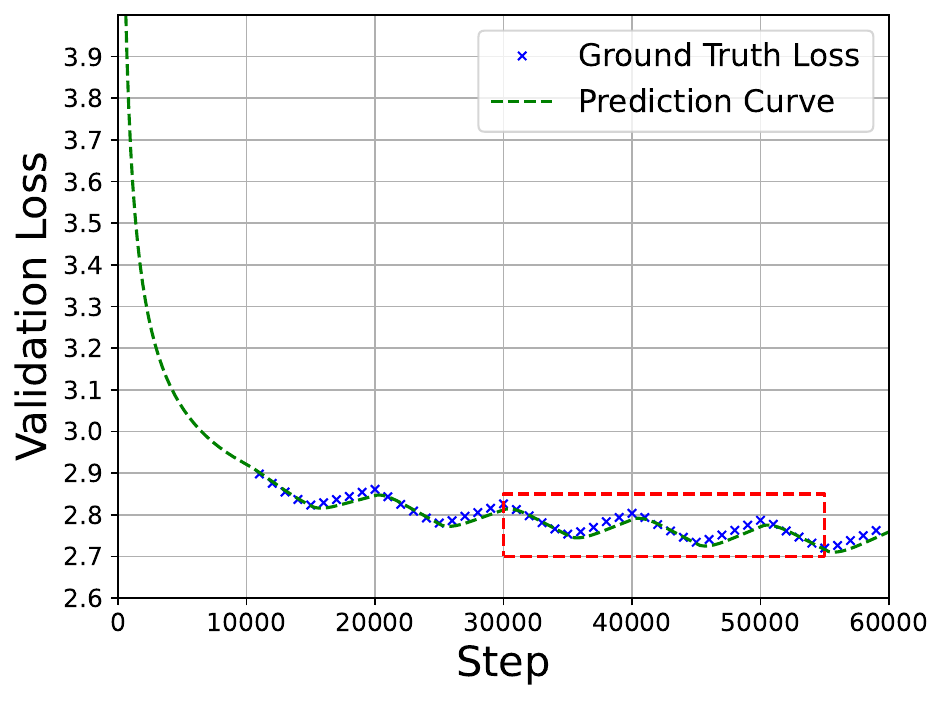}
        \includegraphics[width=0.32\textwidth]{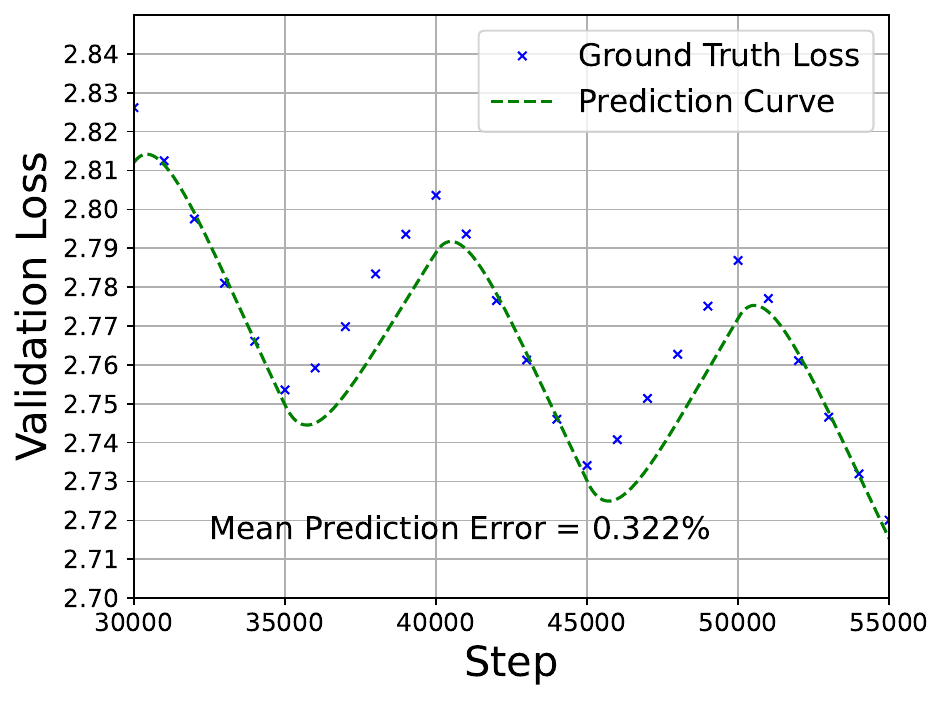}
        \caption{Full Loss curve prediction of the Cyclic LRS \citep{smith2017cyclical} including multiple unseen LR re-warmup stages.}
        \label{fig:prediction-strange}
    \end{subfigure}
    
    \caption{Using the fitted equation from Fig.~\ref{fig:fit} to \textbf{predict} full loss curves for unseen LRS with 60K total steps. The left, middle, and right columns present the LR curve, the loss curve, and a zoomed-in view of loss curve, respectively. 
%     The red rectangle means the zoomed-in zone. 
    Warmup steps (500) are not shown in this figure.
    The fitted equation accurately predicts each loss curve, particularly for capturing the trend of loss changes as the LR varies. 
    Notable, all LRS and loss curves shown here were \textbf{unseen} during the fitting in Fig.~\ref{fig:fit}. The mean prediction errors across different LRS is as low as $\sim 0.2\%$.
    }
\label{fig:prediction}
\vspace{-10pt}
\end{figure}

Eq.~\ref{eq:scaling} describes how loss changes for each step in a full loss curve during training. In Eq.~\ref{eq:scaling}, the term $L_0 + A\cdot S_1^{-\alpha}$ represents a rectified scaling law that captures the expected loss decreases as a power-law function of the number of training steps. The new term $-C\cdot S_2$ accounts for the further loss drop due to learning rate annealing. Remarkably, this simple formulation accurately describes the validation loss at any training step across various LRS and even allows us to predict the loss curve for unseen LRS. For example, we can fit Eq.~\ref{eq:scaling} to the full loss curve of constant and cosine LRS with 20K total steps (Fig.~\ref{fig:fit}), and then predict the full loss curve for various unseen LRS with longer total steps (e.g. 60K) (Fig.~\ref{fig:prediction}).

% \begin{wrapfigure}{r}{0.5\linewidth}
% \centering
% \includegraphics[width=\linewidth]{images/definition}
% \caption{The definition of $S_1$ and $S_2$, taking a 20-step cosine learning rate schedule as a toy example. $S_1$ (forward area) is the area enclosed by the learning rate curve and the X-axis step, which can be approximately regarded as the total amount of movement of neural network parameters; $S_2$ is the decayed annealing area of the learning rate curve (the lighter the color, the greater the degree of decay), that is, the weighted sum of the areas of the small blue squares. 
% %Both the increase of $S_1$ and $S_2$ can help to reduce the loss, but they exist in a delicate balance or trade-off relationship. To minimize the final loss, it's necessary to discover optimal balance between $S_1$ and $S_2$.
% }
% \label{fig:definition}
% \vspace{-10pt}
% \end{wrapfigure}

We validate our proposed equation through extensive experiments and find that: (1) Our formulation performs consistently well across various hyper-parameters and model architectures;  (2) Eq.~\ref{eq:scaling} can be extended to incorporate other scaling factors, such as model sizes; (3) Our proposed equation accurately fits the loss curves of open-sourced models;
%(e.g. BLOOM~\citep{workshop2022bloom} and OLMo~\citep{groeneveld2024olmo}); 
(4) Our formulation can be used to verify and explain numerous previous findings regarding LR annealing and scheduling.

In Sec.~\ref{sec:theorem}, we derive the scaling law formulation with LR annealing and elucidate the potential theory underpinning our formulation. Extensive experiments are conducted to validate the formulation.
In Sec.~\ref{sec:takeaways}, we apply our formulation to verify and explain the empirical results from various previous studies. 
% These conclusions, which previously required substantial computational resources,
% can now be consolidated in a cost-free manner using our equation.
Our approach offers theoretical insights into the crux of loss drop, LR schedule, and LR annealing,
%(i.e., the art of balance between forward area and annealing area).
enabling LLM participants to better understand training dynamics of LLM and select optimal training recipes in advance.
In Sec.~\ref{sec:comparison}, we compare our approach to typical scaling law formula, such as the Chinchilla scaling law~\citep{hoffmann2022training}. We show that 
%chinchilla scaling law actually describes special cases (the endpoint of one full loss curve) of our equation, which means that 
our formulation is more general and requires significantly less compute (less than 1\%) to fit,
% Furthermore, we compare the computational costs required by the chinchilla scaling law and our equation, revealing that our equation offers significant savings in computational costs. 
which greatly democratizes the development of LLMs and scaling laws.

\section{Preliminary}
\subsection{Scaling Laws}
Cross-entropy loss of language models on the validation set is a reliable indicator of LLMs' performance on downstream tasks~\citep{caballero2022broken,du2024understanding}. \citet{kaplan2020scaling} empirically discovered a power-law relationship between validation loss $L$ and three factors: model size $N$, dataset size $D$, and training compute.
As an application of scaling law, \citet{hoffmann2022training} developed Chinchilla, a compute-optimal LLM, by balancing model size and dataset size. They used a simplified and intuitive scaling law equation:
\begin{equation}
\begin{aligned}
    L (D,N) = L_0 + A\cdot D^{-\alpha} + B\cdot N^{-\beta},
\end{aligned}
\label{eq:chinchilla}
\end{equation}
where $L_0$, $A$, $B$, $\alpha$, $\beta$ are positive constants. 
Traditional scaling law formulations fit only the loss at the final training step, while ignoring losses from other steps. Note that collecting a new loss value of data size requires launching a another training run with the same LRS, which is resource-intensive.
Previous works have conducted some preliminary studies on the impact of the learning rate on the scaling laws.
For example, OpenAI and chinchilla scaling laws both report that the choice of learning
rate schedule does not influence the power-law format~\citep{kaplan2020scaling,hoffmann2022training}. 
Also, OpenAI's experiments suggest that the specific choice of learning rate schedule has minimal impact on the final validation loss, provided that the total summed learning rate is adequately large and the schedule incorporates both a warmup stage and a final annealing stage, reducing the learning rate to nearly zero at the end of training~\citep{kaplan2020scaling}.

\subsection{Learning Rate Annealing}
Learning rate annealing is a widely-used technique in training neural networks, where the learning rate is progressively reduced from a maximum to a minimum value following a pre-defined LRS. Various LRS schemes have been proposed to improve the performance and stability of model training. For example, the popular cosine LRS~\citep{loshchilov2016sgdr} reduces the LR in a cosine-like pattern over full training steps. WSD LRS~\citep{hu2024minicpm} keeps a constant LR for the majority of training, and applies annealing only in the final (e.g. $10\%\sim 20\%$) steps. During the training of LLMs, it has been widely observed
that a more pronounced decrease in the learning rate often results in a more precipitous drop in the
validation loss~\citep{loshchilov2016sgdr,simple-scalable-cpt2024,deepseek-ai2024deepseek,hu2024minicpm}. However, to the best of our knowledge, all previous studies end with providing a rough and
qualitative description of how loss changes during LR annealing, while our work provides an accurate equation to
quantitatively formulate the loss changes during LR annealing.

%~\citep{loshchilov2016sgdr,hu2024minicpm}. 
%However, to the best of our knowledge, all previous studies end with providing a rough and qualitative description of how loss changes during LR annealing, while we provide an equation to formulate the loss changes during LR annealing in this work.

% \citet{hu2024minicpm} proposes a warmup-stable-decay (WSD) LRS including three learning rate stages, which could help get a lower validation loss compared to the typical cosine LRS. The format is like 
% \begin{equation}
% \label{eq:wsd}
% \begin{aligned}
% WSD(s)=\begin{cases}&
% \frac{s}{T_{warmup}}\eta_{max},\quad s\leq T_{warmup}\\&
% \eta_{max},\quad T_{warmup}<s\leq T_{stable}\\&
% f(s-T_{stable})\eta_{max},\quad T_{stable}<s\leq T_{total}
% \end{cases}
% \end{aligned}
% \end{equation}
% Where $0 \leq f(s - T_{stable}) \leq 1$ is a decreasing function about step $s$, and $\eta_{max}$ is the maximal learning rate.~\citet{hägele2024scaling} consolidates the effectiveness of WSD scheduler by many empirical experiments. Moreover,~\citet{hägele2024scaling} also finds that using 1-sqrt annealing and a moderate
% annealing ratio (e.g. 20\%) can further decrease the final loss.
\section{Theory}
\label{sec:theorem}
In this section, we elaborate the origin, the intuition, and the experimental basis behind Eq.~\ref{eq:scaling}. We then validate our formula through extensive experiments.

\subsection{Similarity between Learning Rate, Gradient Norm, and Loss}

\begin{figure}[tbp]
\centering
\includegraphics[width=0.95\linewidth]{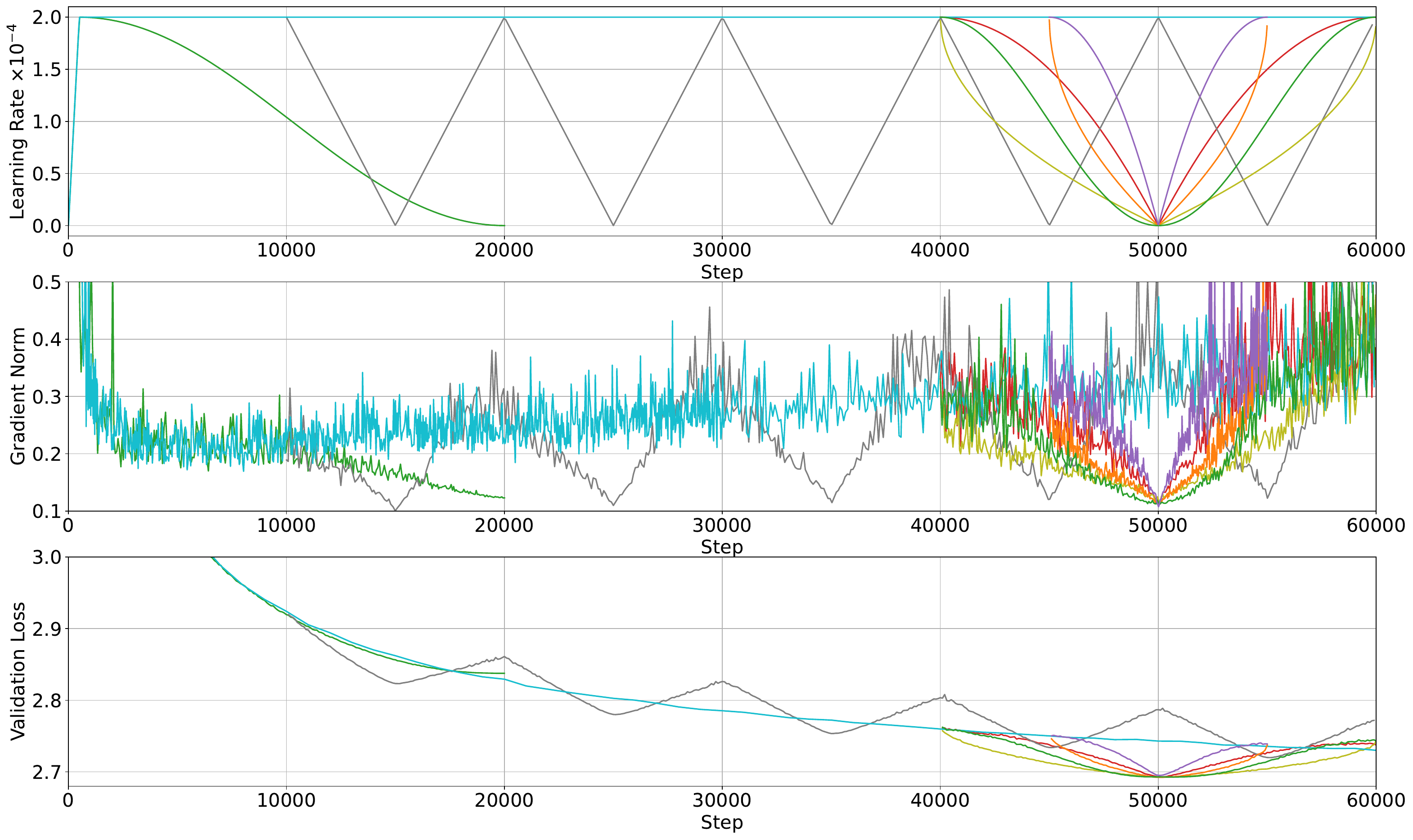}
\caption{The shapes of LR (top), gradient norm (medium), and validation loss (bottom) curves exhibit high similarity across various LRS (labeled as different colors).}
\label{fig:similarity}
\end{figure}

The first key observation is that the shapes of LR curve, gradient norm curve, and validation loss curve are quite similar across various LRS when training LLMs (Fig.~\ref{fig:similarity}). This suggests an implicit connection between learning rate and loss, where gradient norm could be the bridge.

\paragraph{Scaling Laws for Constant LRS.}
A constant LRS is a special LRS, where every training step can be viewed as an endpoint of the LRS. Notably, the Chinchilla scaling law~\citep{hoffmann2022training} exactly fits losses of last steps, i.e., LRS endpoints. 
Therefore, the expectation of validation loss of all steps under in constant LRS adheres to a power-law over training step $s$.

\paragraph{Extra Loss Changes in LR Annealing.}
Unlike a constant LRS, LR annealing (or re-warmup) brings significant local changes in the loss (see Fig.~\ref{fig:similarity}), causing the full loss curve to deviate from the traditional power-law formulation that consider only the training steps $s$. We hypothesis that such loss changes can be captured by an additional LR ($\eta$) related term, i.e.,
\begin{equation}
\begin{aligned}
L(s) = {\color{cyan} \underbrace{L_0 + A\cdot s^{-\alpha}}_{\text{Traditional scaling law}}} 
\quad
{\color{red} \underbrace{-f(\eta)}_{\text{LR annealing term}}},
\end{aligned}
\label{eq:guess-1}
\end{equation}
where the first two terms (blue part) follow traditional scaling laws, while the last term (red part) denotes the extra loss change brought by LR annealing. Recall the similarity between learning rate and loss curves, we can form a naive guess for $f(\eta)$ as $f(\eta) = C\cdot\eta_s$, where $C$ is a positive constant.

\paragraph{Training Discount in Annealing.} The form of Eq.~\ref{eq:guess-1} is still imperfect. 
Note that the gradient norm $\Vert \mathbf{g} \Vert$ decreases along with LR during the annealing process (shown in Fig.~\ref{fig:similarity}).
Thus, the amount of parameter movement (approximately $\eta \cdot \Vert \mathbf{g} \Vert$ per step) in the LR annealing stage declines at an almost quadratic rate compared to stages before annealing.
As the parameter movement become smaller, the change in loss also slows down accordingly. Therefore, the loss drop brought by the power law term (i.e., the first two terms in Eq.~\ref{eq:guess-1}) should also diminish during LR annealing. 
This consideration leads to an improved equation:
\begin{equation}
\begin{aligned}
L(s) &= {\color{cyan} L_0 + A\cdot S_1^{-\alpha}} {\color{red} - f(\eta)} \\
S_1 &=  \sum\limits_{i=1}^{s}\eta_i,
\end{aligned}
\label{eq:guess-2}
\end{equation}
where $S_1$ is the forward area, i.e., the area under the LR curve (as visualized in Fig.~\ref{fig:definition}), which could be approximately interpreted as the total amount of parameter updates.

\subsection{LR Annealing Momentum}
\begin{figure}[tbp]
    \centering
    \begin{subfigure}[b]{0.48\textwidth}
        \includegraphics[width=\textwidth]{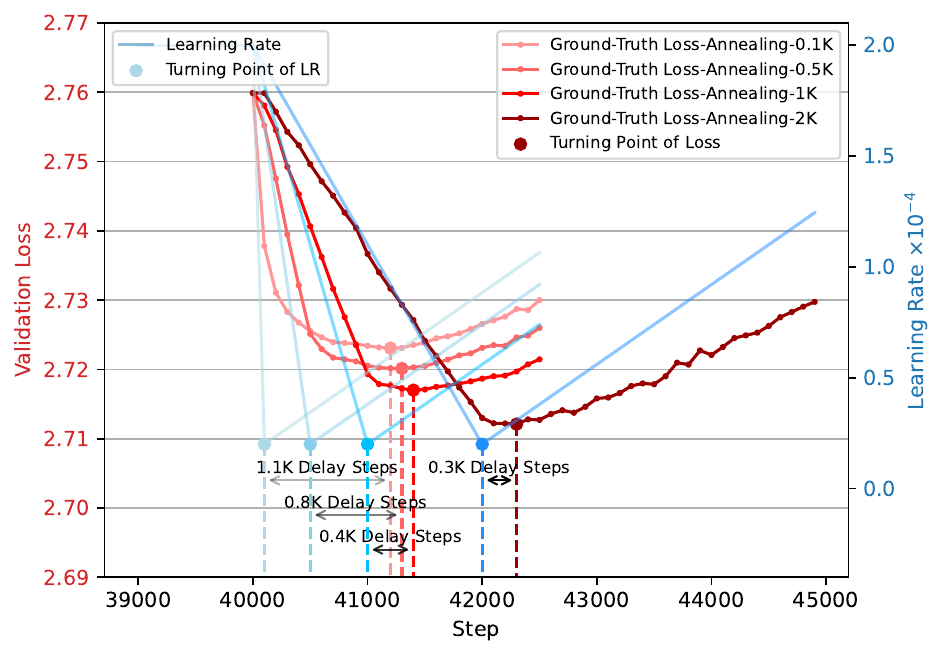}
        \caption{Different delay steps in the annealing process associated with different annealing steps (0.1K, 0.5K, 1K and 2K).}
        \label{fig:delay-anneal-then-rewarmup}
    \end{subfigure}
    \hfill
    \begin{subfigure}[b]{0.48\textwidth}
        \includegraphics[width=\textwidth]{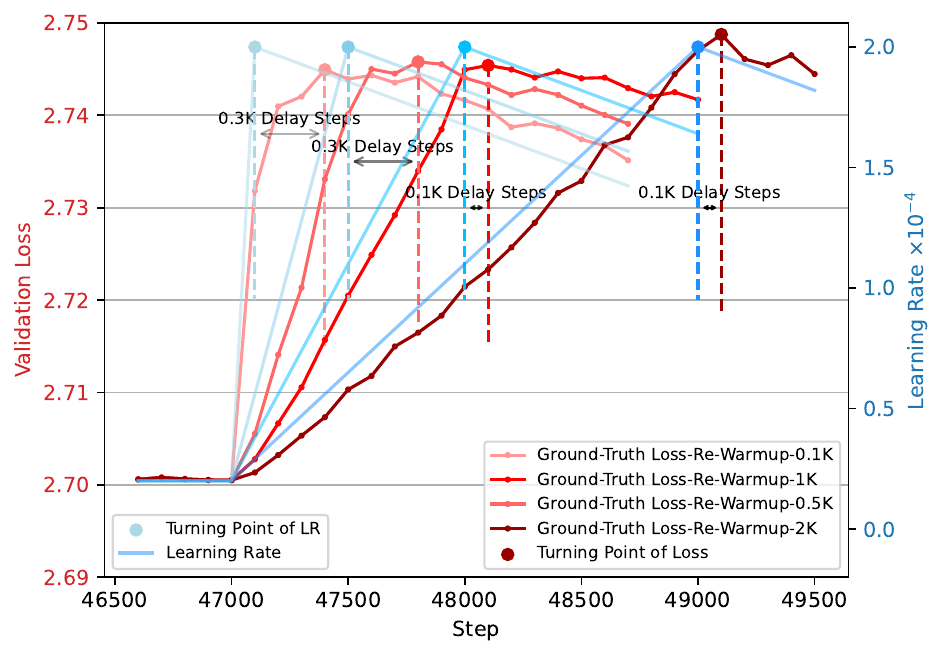}
        \caption{Different delay steps in the re-warmup process associated with different re-warmup steps (0.1K, 0.5K, 1K and 2K).}
        \label{fig:delay-rewarmup-then-anneal}
    \end{subfigure}
    \caption{The delay phenomenon between the LR and validation loss curves. This phenomenon suggests that LR annealing (re-warmup) has momentum.}
\label{fig:delay}
\end{figure}

Another key observation is that LR annealing has momentum.
To refine the formulation of $f(\eta)$,
we design a special LRS where the LR decreases linearly from $\eta_{max}$ to $\eta_{min}$ and then increases.
The increasing stage always has a fixed slope, reaching the maximum value in 5K steps, while the slope of the decreasing stage is varied, with durations of 0.1K, 0.5K, 1K, and 2K. 
Symmetrically, we design another LRS where the LR increases linearly from $\eta_{min}$ to $\eta_{max}$ and then decreases. Fig.~\ref{fig:delay} shows the corresponding LR and loss curves.

We observe a \textbf{delay phenomenon} between the LR and the validation loss. Firstly, the turning point of the validation loss curve consistently lags behind the turning point of the LR curve, indicating that the validation loss continuous along its previous trajectory for some steps even after the LR changes direction. Secondly, the steeper the slope of the LR annealing (or re-warmup), the more pronounced the delay phenomenon becomes. 
Thirdly, given the same LR slope, the left figure (where LR decreases then increases) consistently shows a longer delay compared to the right figure (where LR increases then decreases). We discuss some possible root reasons of delay phenomenon in Sec.~\ref{sec:discsussion-delay}.

Interestingly, this phenomenon closely resembles the physical experiment of a small ball rolling down a slope.
The steeper the slope, the faster the ball accelerates. 
When the ball lands, the accumulated momentum causes the ball to slide further.
Inspired by this delay phenomenon, we hypothesize that $f(\eta)$, the loss reduction induced by LR annealing, has cumulative historical formation so that the past change of learning rate will affect the following
loss curve for a few steps.
In summary, \emph{learning rate annealing exhibits momentum}. 
To capture this, we define $f(\eta) = C\cdot S_2$, where $S_2$ is calculated as:
\begin{equation}
\label{eq:definition-s2}
\begin{aligned}
    m_i &= \lambda \cdot m_{i-1} + (\eta_{i-1} - \eta_i), \\
    S_2 &= \sum\limits_{i=1}^{s}m_i = \sum\limits_{i=1}^{s}\sum\limits_{k=1}^{i}(\eta_{k-1} - \eta_{k})\cdot \lambda^{i-k},
\end{aligned}
\end{equation}
where $m_i$ is the LR annealing momentum at step $i$ ($m_1=0$), and $\Delta \eta = \eta_{i-1} - \eta_i$ denotes the LR annealing amount at step $i$. $\lambda$ is the decay factor that signifies how much historical information is retained. 
We find that $\lambda$ values between $0.99$ and $0.999$ generally works well. 
In contrast, $\lambda=0$ implies no momentum effect, reducing $f(\eta)$ to $C\cdot\eta_s$, which degenerate to the initial form mentioned above. 
Note that $S_2$ applies not only to LR annealing ($S_2>0$), but also to LR re-warmup ($S_2<0$). This means that our equation is applicable to scenarios like continual pre-training, where LR re-warmup serves
as an important factor for better outcomes (see Sec.~\ref{sec:takeaways}). 
More intuitively, the definition of $S_2$ can be visualized
in Fig.~\ref{fig:definition}, as the weighted sum of blue grid areas.

\subsection{Final Formulation}
\paragraph{Scaling Law with LR Annealing.}
\emph{Given the same training and validation dataset, the same model size, the same training hyper-parameters such as warmup steps, max learning rate $\eta_{max}$ and batch size, the language modeling loss at training step $s$ empirically follows the equation $L(s) = L_0 + A\cdot S_1^{-\alpha} - C\cdot S_2$, where $S_1$ and $S_2$ are defined in Eq.~\ref{eq:scaling}. $L_0$, $A$, $C$, $\alpha$ are positive constants.}

Our formulation describes the loss of each training step across different LRS. It allows fitting based on a simpler LRS with shorter training steps and enables the prediction of validation losses for more complex LRS with longer training steps.

\begin{wrapfigure}{r}{0.45\linewidth}
    \centering
    \includegraphics[width=\linewidth]{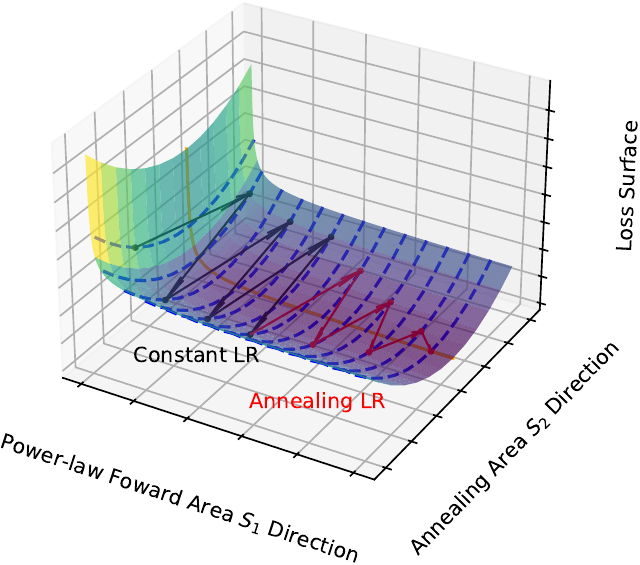}
    \caption{Loss surface of language models as a \emph{slide} after simplification. Optimization direction could be decomposed into two directions: power-law scaling direction ($S_1$, sliding down) and annealing direction ($S_2$, inner height of the slide).}
\label{fig:slide}
\end{wrapfigure}

\paragraph{Only {\color{red} One Extra} Parameter.} Notably, we add only one parameter, the coefficient of $S_2$ term $C$, compared to Chinchilla scaling law~\citep{hoffmann2022training}. 
%Theoretically, enough parameters can always fit any curve but cannot generalize to unseen curve prediction. 
That is, we utilize the fewest extra parameters but model the essential training dynamics in LR annealing to the greatest extent.

\paragraph{Loss Surface as a Slide.} To better understand our formulation, we view the loss surface of language models as a slide in Fig.~\ref{fig:slide}. The optimization process can be seen as sliding down the slide according to the power-law scaling ({\color{orange} orange line}), while oscillating on the inner wall ({\color{blue} blue dashed line}). When the learning rate anneals ({\color{red} red line}), the amplitude of the oscillation decreases, resulting in a reduction in loss.

% \begin{figure}[tbp]
%     \centering
%     \includegraphics[width=0.8\linewidth]{images/slide-crop.pdf}
%     \caption{Loss surface of language models as a \emph{slide} after simplification. Optimization direction could be decomposed into two directions: power-law scaling direction ($S_1$, sliding down) and annealing direction ($S_2$, inner height of the slide).}
% \label{fig:slide}
% \end{figure}

\paragraph{Balance between $S_1$ and $S_2$.}
Note that in Eq.~\ref{eq:scaling}, $\frac{\partial L}{\partial S_1}<0$ and $\frac{\partial L}{\partial S_2}<0$ always hold, indicating that increases in both $S_1$ and $S_2$ help to reduce the loss.
However, as shown intuitively in Fig.~\ref{fig:definition}, there exists delicate balance between $S_1$ and $S_2$. When LR begins to anneal and $S_2$ starts to increase, the forward area $S_1$ of subsequent steps starts to diminish instead. 
Our equation aptly describes this delicate balance.
In Sec.~\ref{sec:takeaways}, we elaborate this topic in detail.
\subsection{Experiments}
\paragraph{LR Warmup.} Different warmup steps can result in different loss curves in training from scratch.
During the warmup stage, neural networks are prone to random optimization, resulting in \emph{unpredictable} outcomes~\citep{hestness2017deep}.
Various studies, along with our own pilot experiments (Appendix~\ref{apx:warmup}), show that LR warmup significantly accelerates model convergence.
High gradient norms are usually observed during the LR warmup stage, especially in the initial steps of training (see Fig.~\ref{fig:similarity}). This indicates that model parameters undergo substantial updates during this stage.
% shows that the gradient norm decreases sharply in the first few training steps.  
% is necessary for convergence. Our pilot experiment (refer to Appendix~\ref{apx:warmup}) shows that warmup indeed significantly accelerates convergence. 
%, a finding also noted by~\citet{warmup-evidence,kosson2024analyzing}. 
Therefore, in all our experiments, we linearly warmup LR to reach $\eta_{max}$ but compute $S_1$ and $S_2$ assuming a constant LR value $\eta_{max}$ in the warmup stage.

\paragraph{Experimental Setups.}
We use standard experimental setups for LLM pre-training. In our main experiments,
the training dataset is Fineweb~\citep{penedo2024fineweb} and the validation
dataset is RedPajama-CC~\citep{together2023redpajama}. We train a $594$M non-embedding parameters LLAMA
architecture-like model~\citep{touvron2023llama} from scratch. We use AdamW optimizer~\citep{loshchilov2017decoupled} with $\beta_1 = 0.9$ and $\beta_2 = 0.95$. The weight decay is set as $0.1$ and gradient clip is set as $1.0$. We set maximal learning rate as $2\times10^{-4}$ and batch size as $4$M tokens.
To verify the robustness of our formulation across different experimental settings, we have four other distinct experimental setups (see Appendix~\ref{apx:exp-sets}). 
We adopt the decay factor or learning rate annealing $\lambda=0.999$ in our all experiments and discuss the impact of $\lambda$ in Sec.~\ref{sec:discsussion-lambda}.

\paragraph{Fitting Details.}
Given a LRS of total step $s$, i.e., a learning rate value sequence $\{\eta_1, \eta_2\, \cdots, \eta_s\}$, the value of $S_1$ and $S_2$ for each step can be computed using Eq.~\ref{eq:scaling} in advance.
To estimate $(L_0, A, C, \alpha)$, we adopt a similar fitting method as used by~\citet{hoffmann2022training}. Specifically, we minimize the Huber loss~\citep{huber-loss} between the predicted and the observed log loss values using the L-BFGS algorithm~\citep{L-BFGS}:
\begin{equation}
\label{eq:huber}
\begin{aligned}
\min_{L_0,A,C,\alpha}\quad\sum_{\mathrm{Step~} i}\mathrm{Huber}_\delta\left(\log\hat{L}(i)-\log L(i)\right).
\end{aligned}
\end{equation}
We use the implementation of the \texttt{minimize} method provided by the \texttt{scipy} library. Huber loss is to enhance the robustness of the fitting process and we set the $\delta$ value of Huber loss to $1.0\times 10^{-3}$. To address the potential issue of local minima during fitting, we select the optimal fit by testing across a range of initial conditions. Note that in practice, we can also fit the loss curves generated by multiple LRS using the same set of parameters $(L_0, A, C, \alpha)$. In this situation, we sum the Huber losses from Eq.~\ref{eq:huber} of all fitted LRS.

\paragraph{Fitting and Prediction Results.}
We fit Eq.~\ref{eq:scaling} on the loss curves under constant and cosine LRS with 20K total steps (see Fig.~\ref{fig:fit}), and then predict the full loss curves under several unseen LRS with 60K total steps (see Fig.~\ref{fig:prediction}). The results show an almost perfect fit, achieving a coefficient of determination ($R^2$) greater than $0.999$. This underscores the robust capability of our equation to accurately fit loss curves across diverse LRS using a single parameter tuple.

The prediction results in Fig.~\ref{fig:prediction} indicate that our formulation is broadly applicable and  generalizes robustly across four unseen LRS, with a mean prediction error as low as 0.2\%. Moreover, our equation can accurately predict losses even for complex LRS that include multiple LR re-warmup stages (Fig.~\ref{fig:prediction-strange}), despite that the loss curves used for fitting do not contain any LR re-warmup stages.

\paragraph{Extensive Experiments on Different Setups.}
To demonstrate the broad applicability of our proposed equation, we conduct additional fitting and prediction experiments using various setups. (1) We use an alternative set of training hyper-parameters (Appendix~\ref{apx:exp-2}); (2) We test our equation on the Mixture of Experts (MoE) architecture (Appendix~\ref{apx:moe}); (3) We apply our equation to predict loss curves for a much longer ($10\times$) training run involving a 1.7B parameter model trained on 1.4T tokens (Appendix~\ref{apx:long-exp}).
(4) We fit the loss curves of open-sourced models, including BLOOM-176B trained on 300B tokens \citep{workshop2022bloom} and OLMo-1B trained on 2T tokens ~\citep{groeneveld2024olmo} (Appendix~\ref{apx:open-source}).
All experiments produce excellent results, indicating that our equation is effective across diverse experimental setups, including different training hyper-parameters, architecture, model sizes, and dataset scales.

\begin{figure}[tbp]
    \centering
    \begin{subfigure}[b]{0.48\textwidth}
        \includegraphics[width=\textwidth]{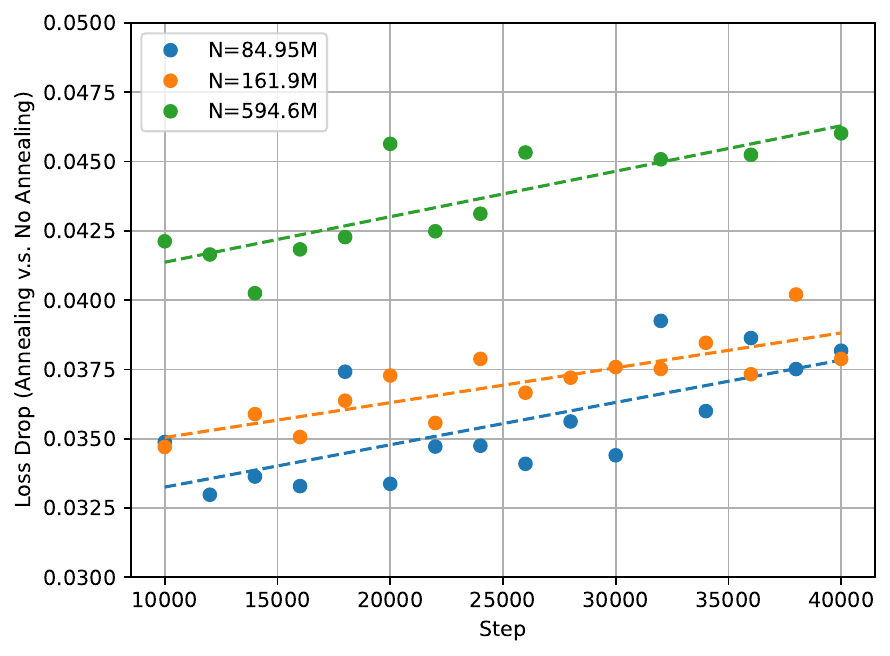}
        \caption{The loss drop brought by LR annealing for different model size $N$. Dashed lines represent trends over steps. The loss drop brought by LR annealing scales with data size and model size.}
        \label{fig:delta-N}
    \end{subfigure}
    \hfill
    \begin{subfigure}[b]{0.48\textwidth}
        \includegraphics[width=\textwidth]{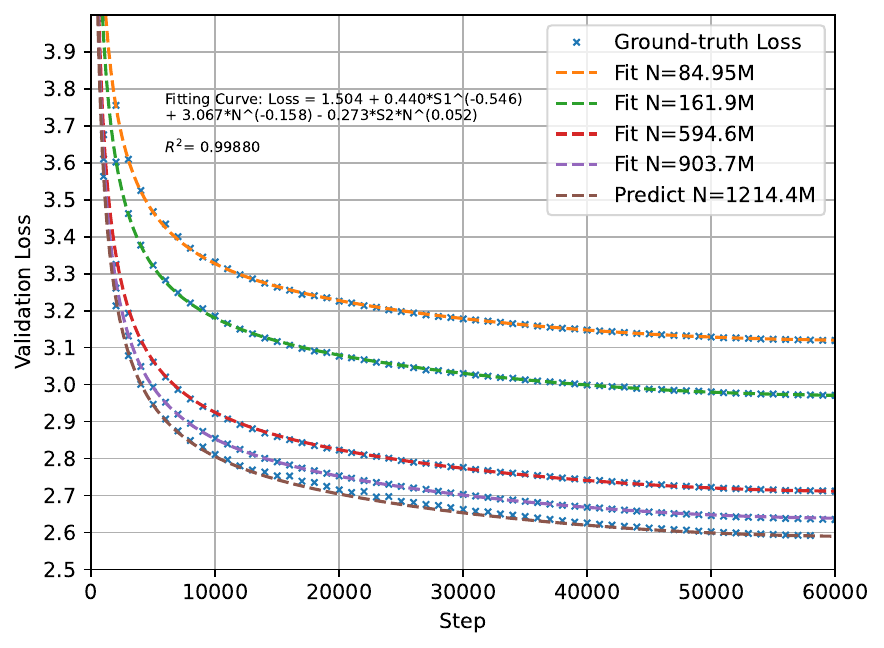}
        \caption{Curve fitting and prediction on cosine LRS (60K steps to $\eta_{min}=0.1\cdot\eta_{max}$) of different model sizes using our $N$-extended scaling law. Results for N=1214.4M are predicted.}
        %Substituting specific values of $N$ can simplify to Eq.~\ref{eq:scaling}.}
        \label{fig:scaling-N-fit}
    \end{subfigure}
    \caption{The loss drop brought by LR annealing (left) and the $N$-extended full loss curve fitting and prediction (right). It suggests that $S_2 \propto N^\gamma$ is reasonable from both experimental phenomena and accurate curve prediction.}
\end{figure}

\subsection{Extension to Model Size Scaling}
\paragraph{Loss Drop During Annealing Scales with Model Size $N$.}
We explore the effect of model size $N$ on the loss drop during the annealing stage. Specifically, we compare the final losses obtained with a constant LRS and a WSD LRS (10\% cosine annealing to $\eta_{min}=0$) to estimate the loss drop due to LR annealing. We conduct this experiment on different total steps and different model sizes. The experimental results are shown in Fig.~\ref{fig:delta-N}. It suggests that the loss drop from LR annealing scales with both annealing steps and model sizes. This implies that the annealing area $S_2$ in our equation should also increase as the model size $N$ increases. We suppose there is a simple relationship of $S_2 \propto N^\gamma$ where $\gamma$ is a positive constant.

\paragraph{Model Size Scaling.} Building on the experiments and analysis above, we extend our proposed Eq.~\ref{eq:scaling} to incorporate model size scaling, based on traditional scaling laws (Eq.~\ref{eq:chinchilla}):
\begin{equation}
\begin{aligned}
& L(s, N) = L_0 + A\cdot S_1^{-\alpha} + B\cdot N^{-\beta}- C\cdot S_2 \cdot N^{\gamma},
\end{aligned}
\label{eq:scaling-N}
\end{equation}
where $N$ is the number of non-embedding model parameters, and $B$, $\beta$, $\gamma$ are positive constants related to $N$.
We parameterize $S_2 \propto N^{\gamma}$ via a multiplier $N^{\gamma}$ to the original annealing term $-C\cdot S_2$.

\paragraph{Fitting and Prediction with Model Size.}
We validate Eq.~\ref{eq:scaling-N} by fitting the full loss curves of models with varying sizes.
We then apply the obtained equation to predict full loss curve on the unseen largest model size.
Results in Fig.~\ref{fig:scaling-N-fit} show an almost perfect fit ($R^2>0.998$) and prediction for entire training dynamics of larger-scale models. This indicates the effectiveness and robustness of our proposed $N$-extended equation.
Additional $N$-extended experiments with other setups further confirm the robustness of our formulation (see detail in Appendix~\ref{apx:exp-2-N}).

\section{Takeaways: Experimental Findings Verification and Explanation}
\label{sec:takeaways}

We apply our proposed formulation to validate and provide a theoretical explanation for numerous existing experimental findings regarding the training dynamics of language models. These key insights also guide researchers in selecting critical LRS before initiating model training. An interesting summary is that
\begin{quotation}
{\color{magenta}\textit{The art of learning rate schedule lies in the delicate balancing act between forward area and annealing area.}}
\end{quotation}

\subsection{\color{cyan} It verifies and explains why loss drops more sharply when LR anneals.}
\label{apx:takeaway-sharpdrop}
\begin{figure}[tbp]
\centering
\includegraphics[width=0.7\textwidth]{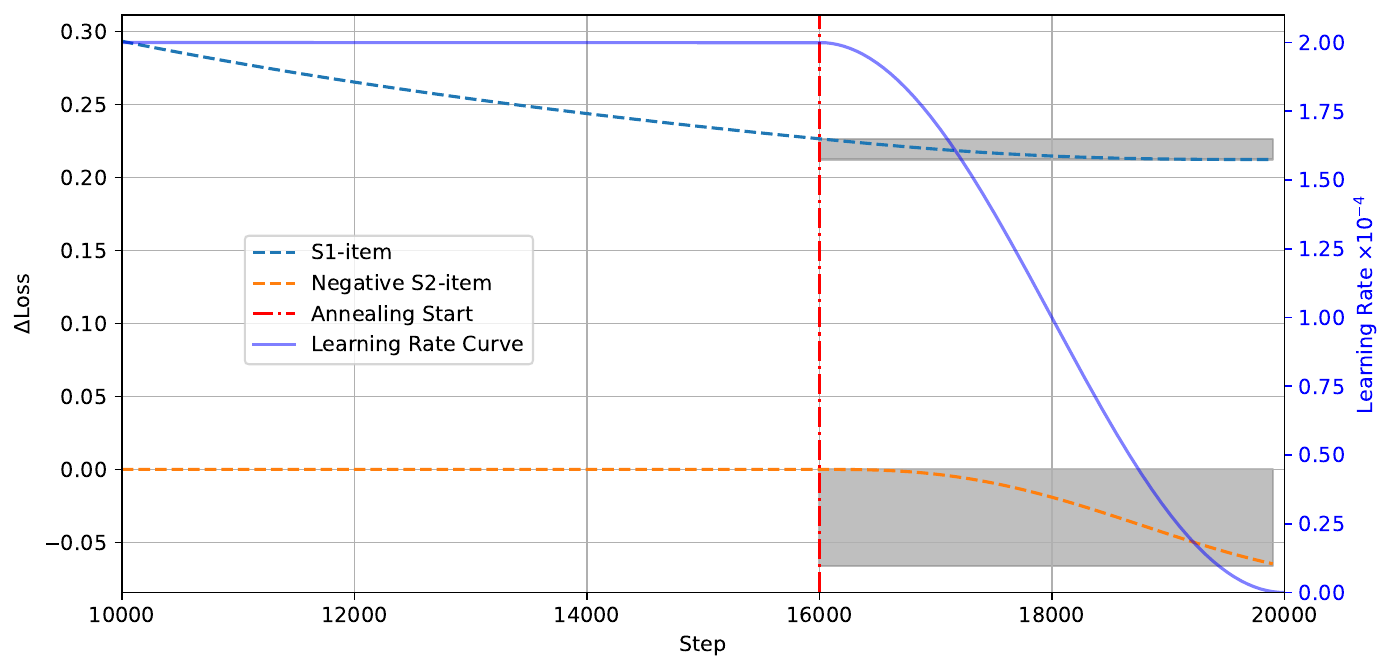}
\caption{How S1-item and negative $S_2$-item changes in a WSD LRS. The upper and lower shaded area indicate the loss drop brought by $S_1$ and $S_2$, respectively, in the annealing stage.}
\label{fig:phe-sharp}
\end{figure}

Our equation helps researchers understand why loss values drop more sharply when LR anneals. This phenomenon has been widely observed in many previous studies.
Without loss of generality, consider the fitted equation from Fig.~\ref{fig:fit}.
We analyze how the $S_1$-item ($A\cdot S_1^{-\alpha}$) and the negative $S_2$-item ($-C\cdot S_2$) impact the loss values (notated as ``$\Delta$Loss'') obtained across a WSD scheduler (see Fig.~\ref{fig:phe-sharp}). It can be seen that in the annealing stage, the negative $S_2$-item has a much more significant impact on the overall loss than the $S_1$-item. Therefore, the loss drops more sharply compared to the stage with constant LR.
In conclusion, LR annealing increases the annealing area, which further leads to a drastic decrease for the validation loss.

\subsection{\color{cyan} Determining cosine cycle length and minimum LR in cosine LRS.}
\begin{figure}[htbp]
    \centering
    \includegraphics[width=0.7\linewidth]{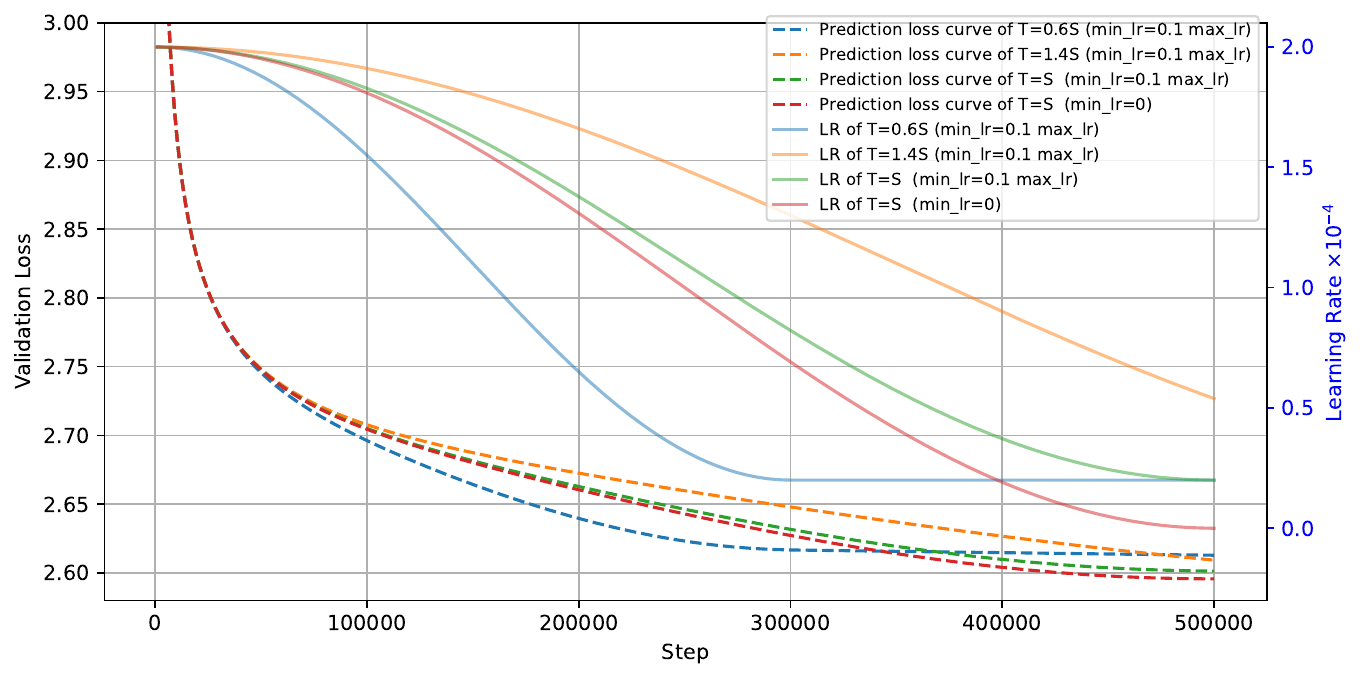}
    \caption{Predicted loss curves of different cycle length $T$ and min LR in cosine LRS. The results well align with previous studies.}
\label{fig:phe-cosine}
\end{figure}

Many papers have found that in LLM pre-training using cosine LRS, setting the cosine cycle length $T$ as the total steps $S$, and setting min LR as nearly 0 (rather than 10\% max LR) can lead to the optimal loss~\citep{hoffmann2022training,hu2024minicpm,hägele2024scaling,parmar2024reuse}. 
% Actually, the settings above have been a factual standard in LLM pre-training using cosine LRS.
We theoretically validate this observation using our equation in Fig.~\ref{fig:phe-cosine}. The predicted loss curve with $T=S$ and a minimum LR of 0 indeed achieves the optimal loss in the final step.
Moreover, our equation gives a quite intuitive explanation: setting $T>S$ leads to incomplete annealing, while $T<S$ leads to a small forward area $S_1$ due to early annealing. Thus, the optimal configuration is to set $T$ equal to $S$. Also, setting the minimum LR to $0$ maximizes the annealing amount, thereby increasing the annealing area $S_2$, which facilitates lower final loss.

\subsection{\color{cyan} It verifies and explains the phenomenon, where constant LRS gets a lower loss than cosine LRS if setting small total steps, and vice versa.}
\label{apx:takeaway-constant-cosine}
\begin{figure}[ht]
    \centering
    \begin{subfigure}[b]{0.32\textwidth}
        \includegraphics[width=\textwidth]{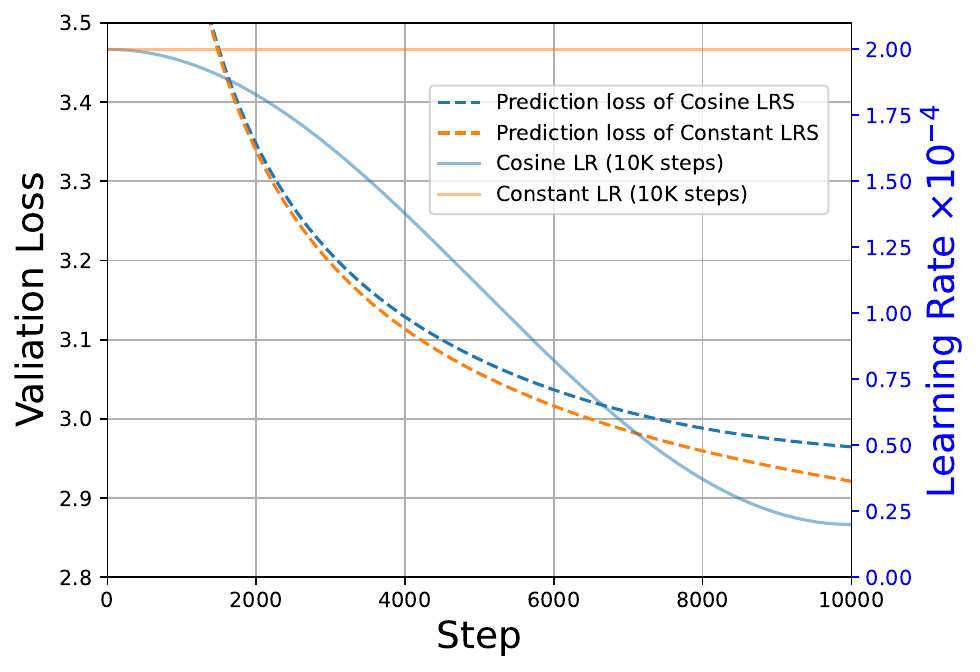}
        \caption{The predicted loss curve of 10K total steps.}
    \end{subfigure}
    \hfill
    \begin{subfigure}[b]{0.32\textwidth}
        \includegraphics[width=\textwidth]{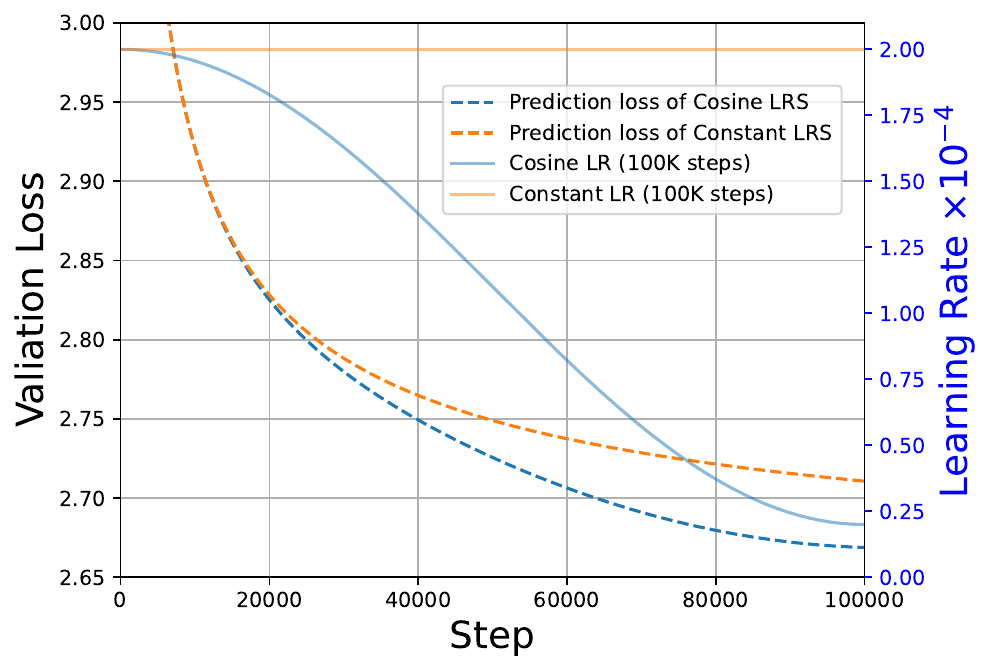}
        \caption{The predicted loss curve of 100K total steps.}
    \end{subfigure}
    \hfill
    \begin{subfigure}[b]{0.288\textwidth}
        \includegraphics[width=\textwidth]{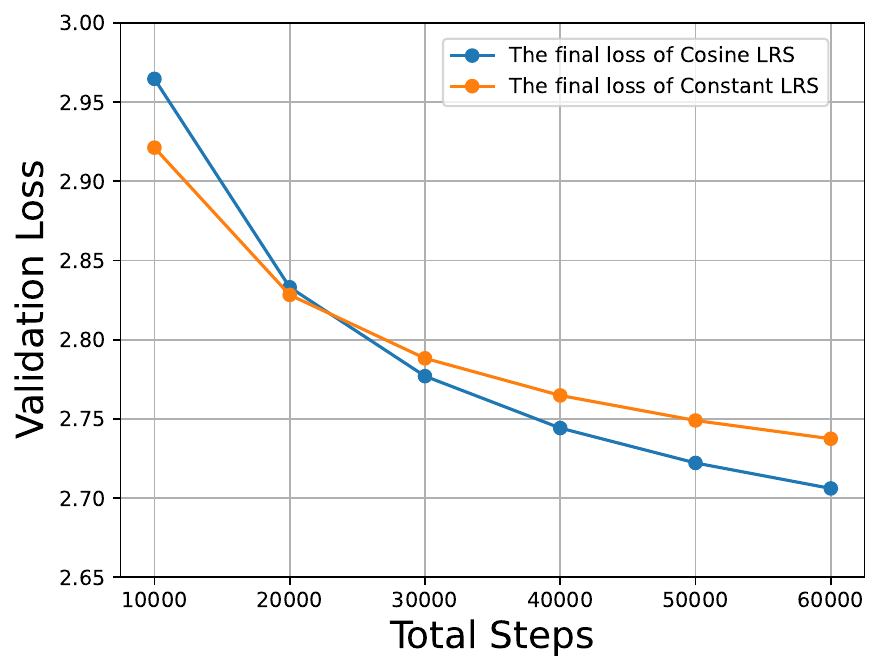}
        \caption{The predicted final loss of different total steps.}
        \label{fig:phe-constant-cosine-final}
    \end{subfigure}
    \caption{Predicted loss curves and final losses for the constant and cosine LRS.}
\label{fig:phe-constant-cosine}
\end{figure}

In our experiments shown in Fig.~\ref{fig:fit}, we observe that the constant LRS can sometimes yield lower final loss values than the cosine LRS.
To validate this phenomenon, we use our equation to predict the loss curve of a constant and cosine LRS for 10K and 100K total steps in Fig.~\ref{fig:phe-constant-cosine}. It can be seen that with more training steps, the cosine LRS outperform the constant LRS, while with less training steps, the constant LRS is better.
Moreover, Fig.~\ref{fig:phe-constant-cosine-final} shows the predicted final loss of different total steps using constant and cosine LRS. It further convincingly suggests that constant LRS indeed gets a lower loss if setting small total steps, but the scaling slope is smaller than cosine LRS's, resulting in higher loss in more steps.

We can further analyze this phenomenon by taking the derivative of $S_1$ and $S_2$ of Eq.~\ref{eq:scaling}. Specifically, $|\frac{\partial L}{\partial S_1}| = \alpha A \cdot S_1^{-\alpha-1}$ is a power-law decreasing function while $|\frac{\partial L}{\partial S_2}| = C$ is a constant.
When the training step $s$ is small, $|\frac{\partial L}{\partial S_1}|$ is larger than $|\frac{\partial L}{\partial S_2}|$ since we have a small $S_1$. Therefore $S_1$ plays a dominant role over $S_2$. In this case, increasing $S_1$ by maintaining a large LR, i.e. using a constant LRS rather than a cosine LRS, leads to more sharply decrease of the loss value.

When the training step $s$ is large,
$|\frac{\partial L}{\partial S_1}|$ becomes smaller than $|\frac{\partial L}{\partial S_2}|$. Therefore, $S_2$ plays a dominant role over $S_1$. In this case, increasing $S_1$ does not contribute much for decreasing the loss. It is time to start LR annealing to increase $S_2$. 
Interestingly, this formulation aligns with the idea of WSD LRS~\citep{hu2024minicpm}: 
In the early stages, the neural network is exploring globally and it is a suitable time to use a larger LR; In the later stages, the neural network is exploring locally and it is a suitable time to use a smaller LR.
We will delve further into WSD LRS in the following subsections.

\subsection{\color{cyan} It verifies and explains WSD and multi-step cosine LRS have matched or even lower loss than cosine LRS.}
\begin{figure}[htbp]
    \centering
    \begin{subfigure}[b]{0.48\textwidth}
        \includegraphics[width=\textwidth]{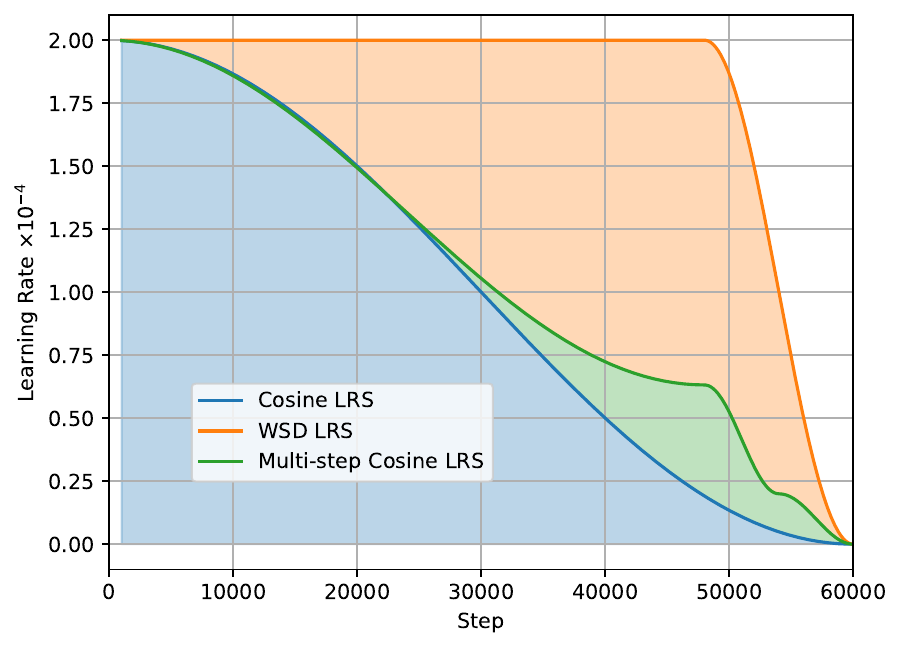}
        \caption{Learning rate curves of three types of LRS.}
    \end{subfigure}
    \hfill
    \centering
    \begin{subfigure}[b]{0.48\textwidth}
        \includegraphics[width=\textwidth]{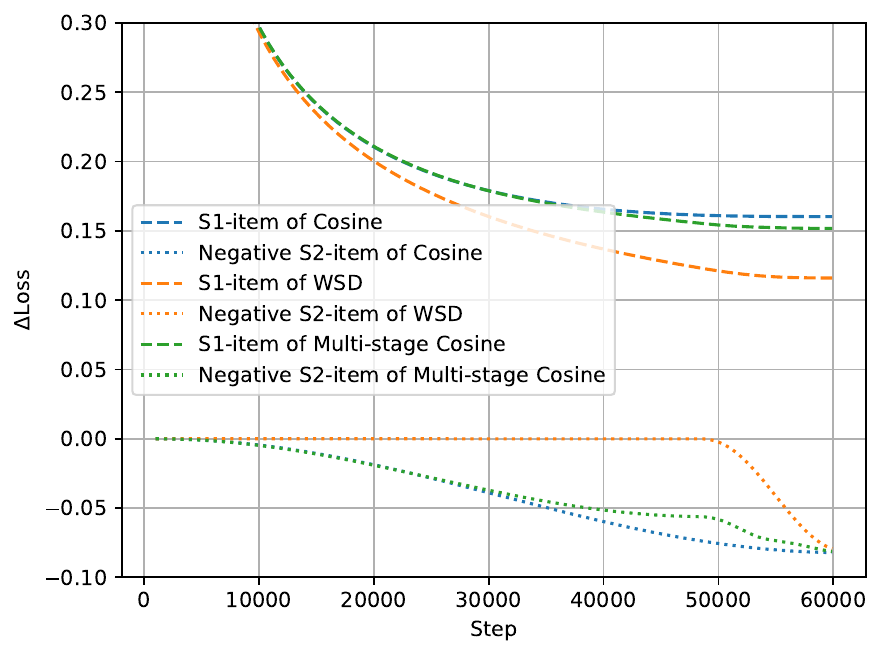}
        \caption{$S_1$-item and negative $S_2$-item of different LRS.}
    \end{subfigure}
    \caption{The comparison between $S_1$-item and negative $S_2$-item in different LRS.}
\label{fig:phe-match}
\end{figure}

Recent studies have shown that WSD LRS~\citep{hu2024minicpm} and multi-step cosine LRS~\citep{deepseek-ai2024deepseek} result in lower loss compared to the traditional cosine LRS. 
Our experiments also support this finding (refer to the ground-truth loss in Fig.~\ref{fig:prediction-cosine},~\ref{fig:prediction-multi-cosine},~\ref{fig:prediction-wsd}). 
We validate and elucidate this finding using our proposed equation. Fig.~\ref{fig:phe-match} shows the learning rate curve (left) and the predicted loss drop (right) for different LRS.
The results suggest that for WSD and multi-step cosine LRS, the negative $S_2$-item ($-C\cdot S_2$) is slightly larger than that of the cosine LRS, whereas the $S_1$-item ($A\cdot S_1^{-\alpha}$) is significantly lower.
Specifically, both the WSD LRS and multi-step cosine LRS unintentionally employ strategies that marginally reduces $S_2$ but substantially increases $S_1$, leading to an overall decrease in validation loss.

\subsection{\color{cyan} Determining Optimal Annealing Ratio of WSD LRS.}
In the case of WSD LRS, it is crucial to ascertain the optimal annealing ratio for training steps. 
~\citet{hägele2024scaling} found that there is an optimal annealing ratio for WSD LRS, i.e. both excessively high or low annealing ratios lead to sub-optimal model performance. 
This phenomenon can be further elucidated through our proposed equation. 
Specifically, a high annealing ratio results in a significant reduction of the forward area $S_1$ while a low annealing ratio leads a diminished annealing area $S_2$. 
Our scaling law equation describes the trade-off between the forward area $S_1$ and the annealing area $S_2$ about the annealing ratio.

\begin{figure}[tbp]
    \centering
    \begin{subfigure}[b]{0.48\textwidth}
        \includegraphics[width=\textwidth]{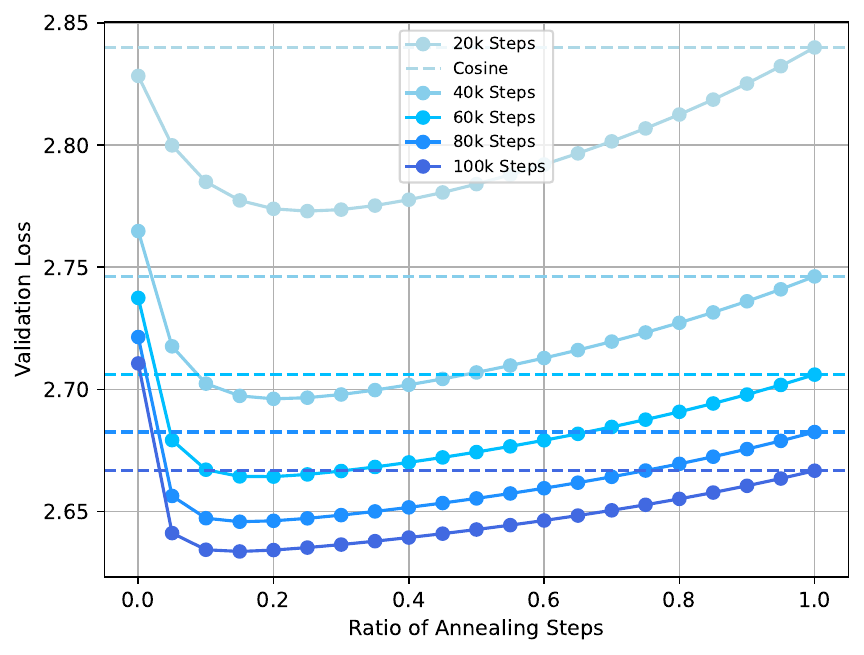}
        \caption{The relationship between the predicted final loss and the ratio of annealing steps under the condition of different total steps.}
    \end{subfigure}
    \hfill
    \begin{subfigure}[b]{0.48\textwidth}
        \includegraphics[width=\textwidth]{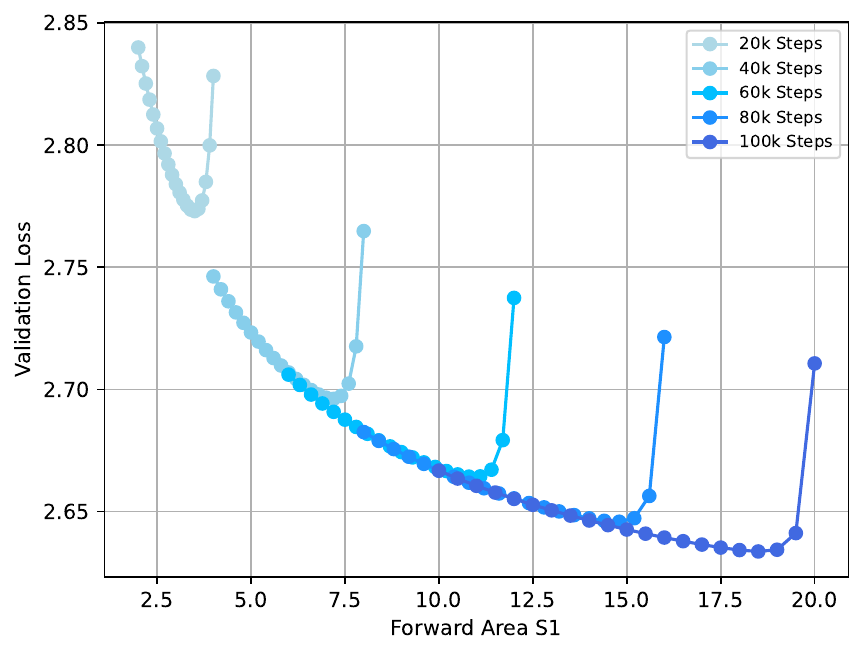}
        \caption{The relationship between predicted final loss and the forward area $S_1$ of different total steps. Different points denote different annealing ratios.}
    \end{subfigure}
    \caption{Illustration of the predicted loss in relation to the ratio of annealing steps and the forward area in WSD LRS (cosine annealing), presenting parabola-like curves, with a distinct optimal loss.}
\label{fig:phe-match:1}
\end{figure}

Fig.~\ref{fig:phe-match:1} depicts the final loss predicted by our equation for various annealing ratios and total training steps. 
The predictions form parabola-like curves, and align well with the actual experimental results reported in previous studies.
This suggests that a moderate annealing ratio, typically around 10\% to 20\%, is optimal, as it balances $S_1$ and $S_2$ to maximize their combined effect, thereby minimizing the overall validation loss.
Moreover, our equation can directly predict the optimal annealing ratio for different total steps without large-scale experiments, which saves a lot of resources.

\subsection{\color{cyan} Determining the optimal annealing function in WSD LRS.}
\label{apx:takeaway-anneal-func}
\begin{figure}[tbp]
    \centering
    \begin{subfigure}[b]{0.48\textwidth}
        \includegraphics[width=\textwidth]{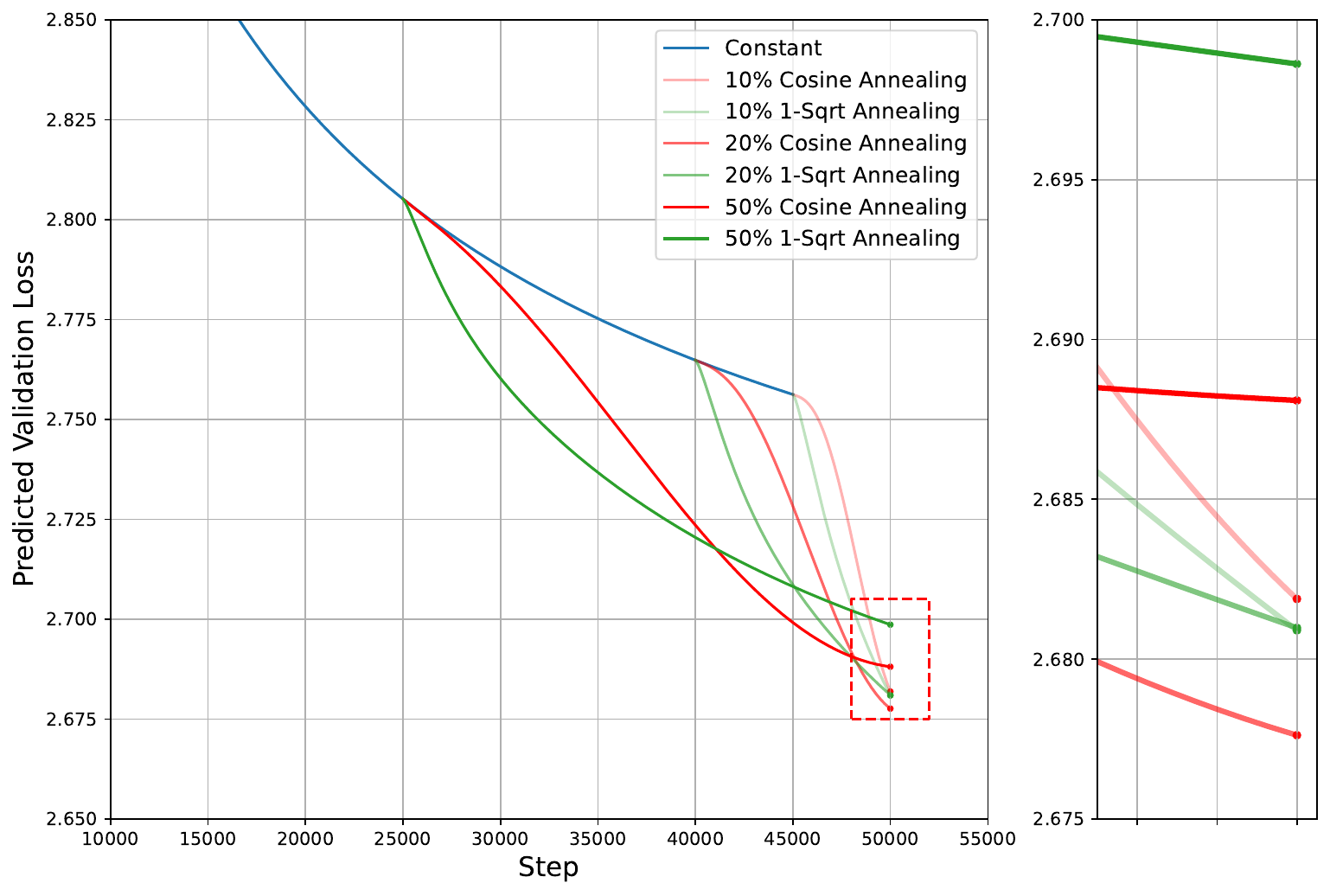}
        \caption{The \textbf{predicted} loss curve of cosine and 1-sqrt annealing method of different annealing ratio.}
        \label{fig:phe-diff-anneal-predict}
    \end{subfigure}
    \hfill
    \begin{subfigure}[b]{0.48\textwidth}
        \includegraphics[width=\textwidth]{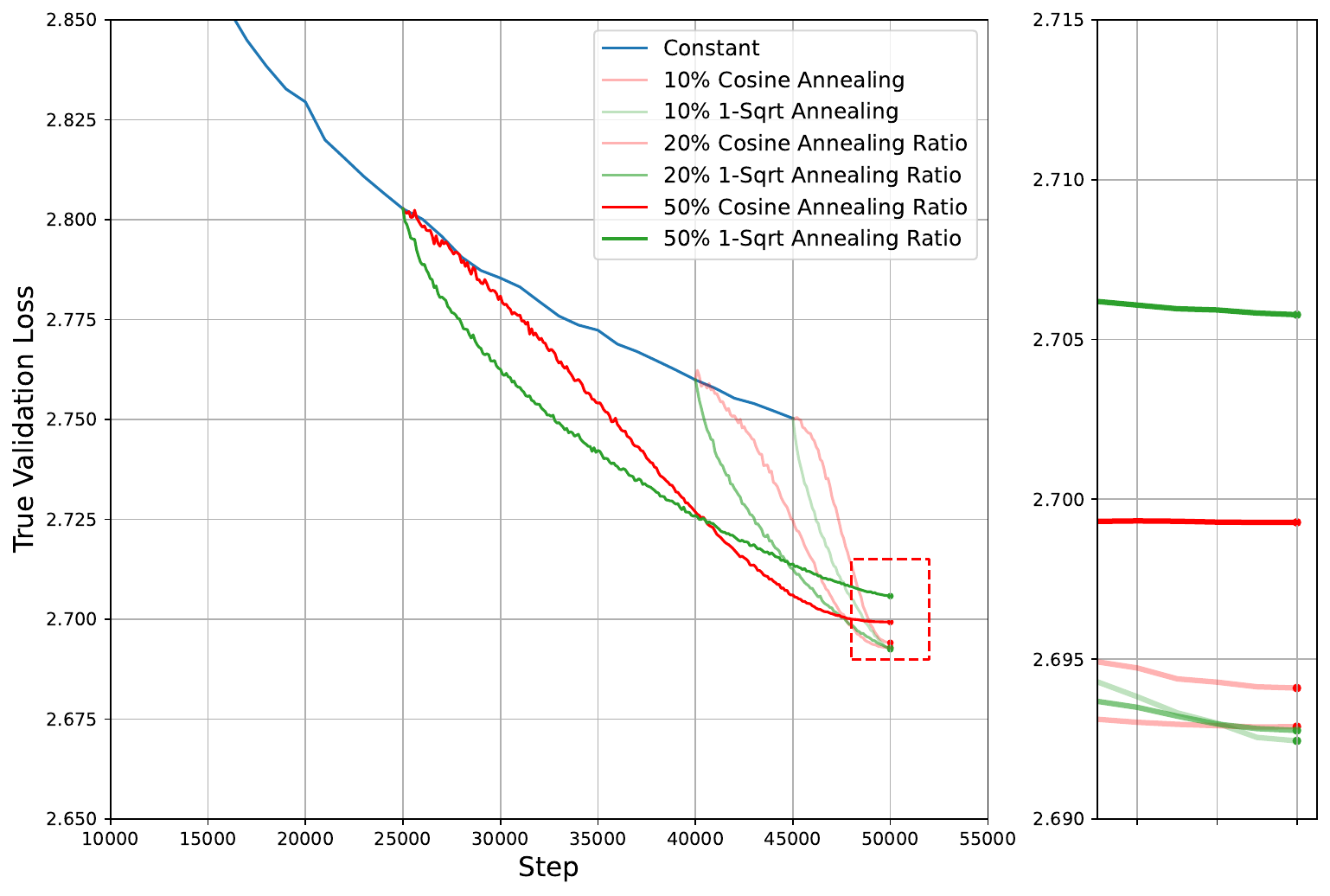}
        \caption{The \textbf{true} loss curve of different annealing ratios with cosine and 1-sqrt annealing methods.}
        \label{fig:phe-diff-anneal-true}
    \end{subfigure}
    \caption{The predicted (left) and true loss (right) of cosine and 1-sqrt annealing method at different annealing ratios. Experimental results (right), aligned with our prediction (left), refute the claim ``the order and results of different annealing hold across settings''~\citep{hägele2024scaling}.}
\label{fig:phe-diff-anneal}
\end{figure}

For the WSD LRS, the selection of the annealing function is another important factor for obtaining further lower loss values. \citet{hägele2024scaling} conclude that the 1-sqrt annealing (see Appendix~\ref{apx:1-sqrt} for the detailed formula) yields lower final losses compared to the other annealing methods (e.g. cosine). 
They claim that this conclusion is true across different annealing ratios.

However, our equation suggests different results (see Fig.~\ref{fig:phe-diff-anneal-predict}). The 1-sqrt annealing approach does get a lower loss than the cosine annealing approach when using small annealing ratios (e.g. 10\%), but it performs much worse than the cosine annealing approach when using 50\% annealing ratio.

To verify our prediction, we conduct experiments by training models using different annealing methods and ratios within 50K total steps. As illustrated in Fig.~\ref{fig:phe-diff-anneal-true}, at a 10\% annealing ratio, the 1-sqrt annealing method obtains lower final losses compared to the cosine annealing method, whereas at a 50\% annealing ratio, the cosine annealing method obtains a lower final loss compared to the 1-sqrt annealing method. The experimental results align quite well with our prediction, which also overturns some of the claims made by previous works. We conclude that the optimal annealing function in WSD LRS depends on the annealing ratio.

Our scaling law equation provides an explanatory framework for these empirical observations. We draw the LR curves associated with the 1-sqrt and cosine annealing approaches in Appendix~\ref{apx:1-sqrt}.
When using an small annealing ratio, the forward area \(S_1\) of the cosine annealing method is slightly larger than that of the 1-sqrt annealing method, while the annealing area \(S_2\) of the 1-sqrt method plays a more dominate role in decreasing the final loss. As the annealing ratio increases, the difference of $S_1$ between two annealing methods becomes larger and larger. Therefore the forward area \(S_1\) gradually takes the dominate role. The delicate balance between $S_1$ and $S_2$ breaks at 50\% annealing ratio, resulting in a lower final loss for the cosine annealing function.

The above analysis underscores the importance of carefully selecting the annealing strategy for the WSD LRS to optimize model training outcomes. Our equation can help predict a better annealing function without experiments, which saves a lot of resources. 

\subsection{\color{cyan} It verifies and explains that in continual pre-training, a higher max learning rate during re-warmup leads to a higher initial peak in the loss, followed by a more rapid decrease.}
\label{apx:takeaway-cpt-peaklr}

In continual pre-training (CPT) settings, the LRS typically involves re-warmup to a new max LR at the start of training. 
Through numerous experiments,~\citet{gupta2023continualpretraininglargelanguage} concludes that a higher max LR during re-warmup results in a higher initial peak loss, followed by a more rapid decrease. 

According to our scaling law formulation~\footnote{Strictly speaking, continual pre-training process often include LR re-warmup as well as data distribution shift. Here we primarily research on the setting that there is no distribution shift between two training stages. The conclusions transfer across most cases because the loss change brought by LR re-warmup is significantly larger than the loss change brought by data distribution shift~\citep{gupta2023continualpretraininglargelanguage,simple-scalable-cpt2024}.}, in the LR re-warmup process, the annealing area $S_{2}$ will reduce to a negative value ($S_2<0$) and thus the validation loss increases. A higher max LR in re-warmup lead to a lower annealing area $S_{2}$, and thus there would be a higher peak loss. 
Moreover, a higher max LR also leads to a faster growth of the forward area $S_{1}$ after re-warmup. We use the fitted equation to predict the continual pre-training process with different max LR as shown in Fig.~\ref{fig:phe-match-cpt-lr}. The predicted loss curves reproduce a quite similar phenomenon with previous works~\citep{gupta2023continualpretraininglargelanguage}. 

There is a more profound strategy using our equation in CPT. As shown in Fig.~\ref{fig:phe-match-cpt-lr}, after the total steps of CPT is determined, we can apply our equation to predict a better max LR and scheduler to get the lowest final loss without experiments, which saves a lot of resources.

\subsection{\color{cyan} It verifies and explains that in continual pre-training, the steps of re-warmup have little impact on the final loss.}
\label{apx:takeaway-cpt-step}

\begin{figure}[tbp]
    \centering
    \begin{subfigure}[b]{0.48\textwidth}
        \includegraphics[width=\textwidth]{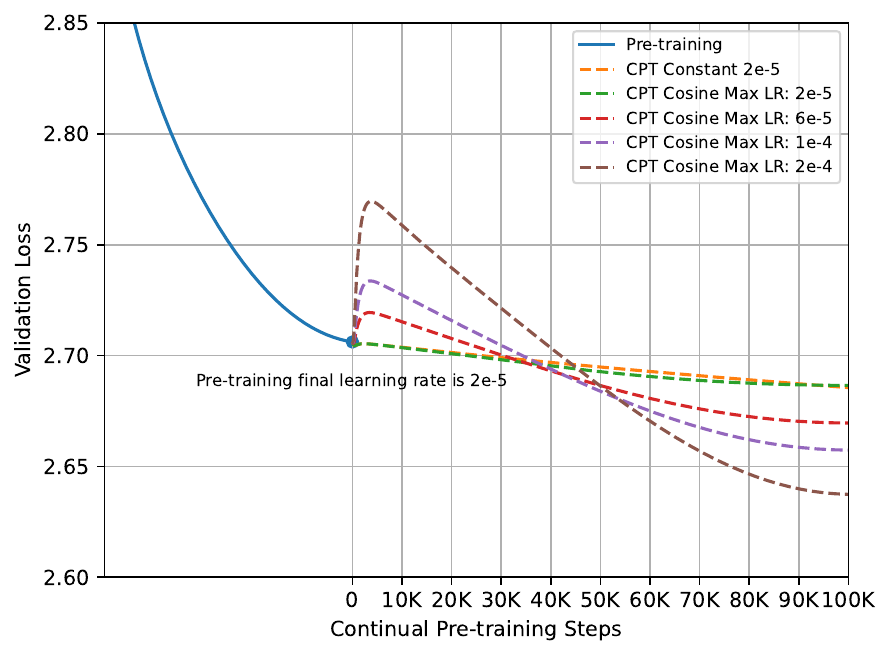}
        \caption{The predicted validation loss with different re-warmup max LR in the continual pre-training process. All the re-warmup steps are 500 steps. }
        \label{fig:phe-match-cpt-lr}
    \end{subfigure}
    \hfill
    \begin{subfigure}[b]{0.48\textwidth}
        \includegraphics[width=\textwidth]{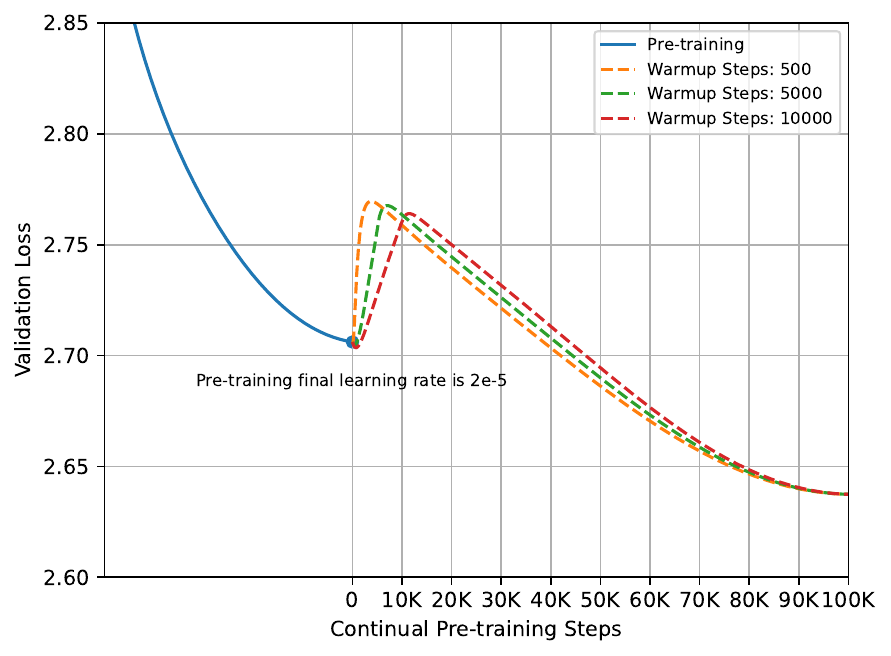}
        \caption{The predicted validation loss with different re-warmup max learning rate in the continual pre-training process. }
        \label{fig:phe-match-cpt-step}
    \end{subfigure}
    \caption{The predicted validation loss with different re-warmup max learning rate and re-warmup steps in the continual pre-training process. The LRS of continual pre-training is cosine ($T$=100K) and the min learning rate is 0.}
\label{fig:phe-match:2}
\end{figure}

Meanwhile, how many steps to re-warmup is another important issue in the continual pre-training.
~\citet{gupta2023continualpretraininglargelanguage} find that longer re-warmup steps smooth the transition of loss curve but the number of re-warmup steps does not significantly influence the final validation loss. 
We use the fitted equation to predict the continual pre-training dynamics with different re-warmup steps. The results, shown in Fig.~\ref{fig:phe-match-cpt-step}, present a good alignment with previous observations~\citep{gupta2023continualpretraininglargelanguage}. 

Based on our theory, given a fixed max LR, longer re-warmup steps lead to a slower decreases of the annealing area, resulting in a smoother raise in the loss curve. However, both the final values of the forward area ($S_1$) and the annealing area ($S_2$) remain relatively stable across different re-warmup steps. The annealing area ($S_2$) corresponding to different re-warmup steps are very close since the max and min LR remain the same. 
Besides, although different re-warmup steps lead to temporary distinct loss fluctuations, re-warmup only accounts for a small portion of the whole training process. Thus, the forward area $S_1$ is also close across different re-warmup steps, resulting in the close overall loss across different steps of re-warmup.

\section{Comparison with Chinchilla Scaling Law}
\label{sec:comparison}

\subsection{Reduction to Chinchilla Scaling Law}
Our scaling law equation can predict the full loss curve across any given LRS. In this section, we show    
that our equation has no contradiction with traditional scaling laws, and it is a generalized form of the Chinchilla scaling law~\citep{hoffmann2022training}. Specifically, all the final loss values for different total training steps following our equation should also follow a power-law relationship. We prove this by dividing two conditions: (1) constant LRS, and (2) other LRS.

\noindent \textbf{Constant LRS.} In the case of a constant LRS, the annealing area $S_2$ is always zero and the forward area $S_1 = \eta_{max} \cdot s$, where $s$ is the step, and $\eta_{max}$ is the constant maximal LR. Thus, the whole loss curve formula becomes:
% $L(s) = L_0 + (A\cdot \eta_{max}^{-\alpha})  \cdot s^{-\alpha} =  L_0 + A^{\prime} \cdot s^{-\alpha}$,
\begin{equation}
L(s) = L_0 + (A\cdot \eta_{max}^{-\alpha})  \cdot s^{-\alpha} =  L_0 + A^{\prime} \cdot s^{-\alpha} 
\end{equation}
which aligns with the Chinchilla scaling law equation.

\begin{figure}[tbp]
    \centering
    \begin{subfigure}[b]{0.48\textwidth}
        \includegraphics[width=\textwidth]{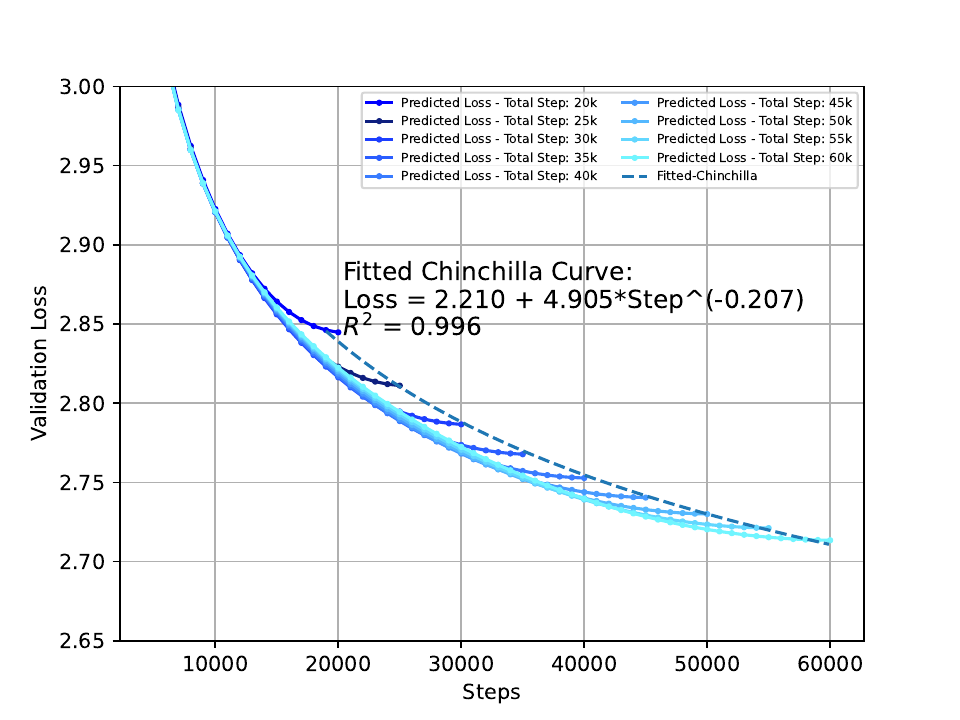}
        \caption{The predicted loss of different total steps with cosine LRS and the fitted chinchilla curve.}
    \end{subfigure}
    \hfill
    \begin{subfigure}[b]{0.48\textwidth}
        \includegraphics[width=\textwidth]{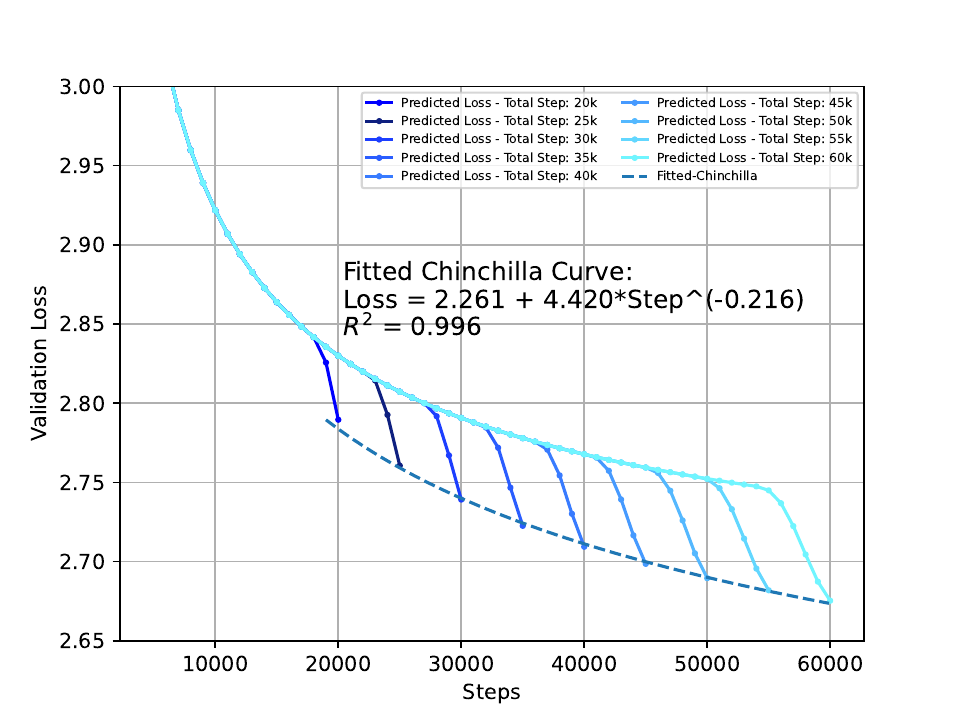}
        \caption{The predicted loss of different total steps with WSD LRS and the fitted chinchilla curve.}
    \end{subfigure}
    \caption{The Chinchilla scaling law equation fits well on the final validation losses predicted by our formulation, taking cosine LRS (on the left) and WSD LRS (on the right) as examples.}
\label{fig:phe-reduction}
\end{figure}

\noindent \textbf{Other LRS.} For non-constant LRS, we use a statistical approach to show that our equation can be reduced to the Chinchilla scaling law. 
Specifically, we verify whether the Chinchilla scaling law adequately fits the endpoints of loss curves predicted by our equation.
The parameter tuple of our equation is $(L_{0}, A, C, \alpha)$. We then randomly sample 1000 sets of parameter tuples from some uniform distributions:
$L_0\sim U(1,3)$, $A\sim U(0.3,0.5)$, $C\sim U(0.2,0.6)$,  $\alpha\sim U(-0.6,-0.4)$.
Each parameter tuple could be seen as the fitting result of a distinct set of experimental setups~\footnote{It's worth noting that some of these sampled parameter tuples might not be reasonable or likely to be observed in real experiments, but we choose to keep them nonetheless.}. (e.g. dataset, batch size, model size, etc.).
For each generated parameter tuple, we apply our equation to predict the final loss of different total training steps on two LRS including cosine and WSD (10\% annealing ratio), 
ranging from 5K to 60K steps. 
The predicted loss values on final steps are used to fit the chinchilla equation. 
The fitting results are shown in Fig.~\ref{fig:phe-reduction}.

\begin{table}[htbp]
\begin{center}
\caption{Mean and standard deviation of $R^2$, as well as the mean Huber loss of 1,000 randomly
generated parameter tuples.}
\label{tab:reduction}
\begin{tabular}{lccc}
\toprule
\textbf{LR Scheduler} & \textbf{mean}($R^2$) $\uparrow$ & \textbf{std}($R^2$) $\downarrow$ & \textbf{Huber Loss} $\downarrow$ \\
\midrule
Cosine         &0.972 &0.056 & 0.00017 \\
WSD             &0.979  &0.053 & 0.00013 \\
\bottomrule
\end{tabular}
\end{center}
\end{table}

Each parameter tuple represents a synthetic fitting result corresponding to a distinct set of experimental setups (e.g. dataset, model size, etc.).
For each sampled parameter tuple, we apply our equation to predict the final loss for different total training steps with both cosine and WSD LRS, and then employ the predicted losses to fit the Chinchilla scaling law.
We calculate the mean and standard deviation of $R^{2}$ values for each fit. 
The results in Table~\ref{tab:reduction} demonstrate that Chinchilla scaling law fits well on the data predicted by our scaling law equation. Thus, our equation can be considered a generalization that can be reduced to the Chinchilla scaling law. 

\subsection{Scaling Law Fitting and Prediction Democratization}

Our scaling law equation allows us to utilize all loss values from a full loss curve during training, while traditional scaling laws can only collect a single data point from the full loss curve. This feature allows us to fit scaling laws with much less cost.
For a direct comparison, we compare the computational efficiency of our approach and the Chinchilla scaling law~\citep{hoffmann2022training}.
Specifically, we assume and evaluate the computational cost of obtaining 100 fitting points for each scaling law equation with an step interval of $K$: 
% \begin{itemize}
%     \item Fitting the \textbf{Chinchilla scaling law equation with the cosine LRS} requires a total number of parameter updates of $1K + 2K + 3K + \dots + 100K = 5050K$;
%     \item Fitting the \textbf{Chinchilla scaling law equation with the WSD LRS} requires a total number of parameter updates of $(1K + 2K + 3K + \dots + 100K)r + 100K(1-r) = (100 + 4950r)K$, in which $r$ is the annealing ratio for the WSD LRS;
%     \item Fitting \textbf{our scaling law equation} only need one training curve with a moderate number of parameter updates, such as one curve under the cosin LRS with $50K$ training steps. Further, the number of loss values we can get for fitting our equation is far more than $100$. Note that more fitting points generally achieve more accurate fitting results. Our equation can also utilize loss curves from various different LRS (e.g. $20K$ constant + $30K$ cosine).
% \end{itemize}

\begin{itemize}
    \item Adopting Chinchilla scaling law, typical cosine LRS requires total steps of at least $1K + 2K + 3K + \cdots + 100K = 5050K$;
    \item Adopting Chinchilla scaling law, WSD LRS (notating annealing ratio as $r$) requires total steps of at least $(1K + 2K + 3K + \cdots + 100K)r + 100K(1-r) = (100 + 4950r)K$.
    \item Adopting our scaling law, all we need is only one or two training curves with moderate total steps, such as one curve with $50K$ steps under cosine LRS. Further, the number of loss values we can get for fitting our equation is far more than $100$. Note that more fitting points generally achieve more accurate fitting results. Our equation can also utilize loss curves from various different LRS (e.g. $20K$ constant + $30K$ cosine)
\end{itemize}

\begin{table}[tbp]
\begin{center}
\caption{The comparison of computational cost for fitting different scaling law equations.}
\label{tab:green}
\begin{tabular}{cccc}
\toprule
\textbf{Equation}  & \textbf{LRS}  & \textbf{Computational cost} & \textbf{Applicable to other LRS?}
\\
\midrule
Chinchilla  & Cosine & 100\% &  No \\
Chinchilla  &  WSD (20\% annealing)   & 21.6\% & No \\
Chinchilla  &  WSD (10\% annealing)   & 11.8\% & No \\
Ours & Any (except constant) & $<$1.0\% & Yes \\
\bottomrule
\end{tabular}
\end{center}
\end{table}

We present a comparison of the computational costs associated with different laws and LRS in Table~\ref{tab:green}. The results indicate that our proposed equation uses less than 1\% of the computational cost required by the Chinchilla scaling law. Further, our scaling law with LR annealing, can be universally applied to predict full loss curves for unseen LRS, thus conserving even more computational resources. This approach significantly democratizes the study of scaling laws in LLM pre-training, paving the way for a more environmentally friendly and carbon-efficient methodology.

\section{Discussion}
\label{sec:discussion}
\subsection{The impact of Decay Factor $\lambda$}
\label{sec:discsussion-lambda}
\begin{figure}[htbp]
    \centering
    \begin{subfigure}[b]{0.32\textwidth}
        \includegraphics[width=\textwidth]{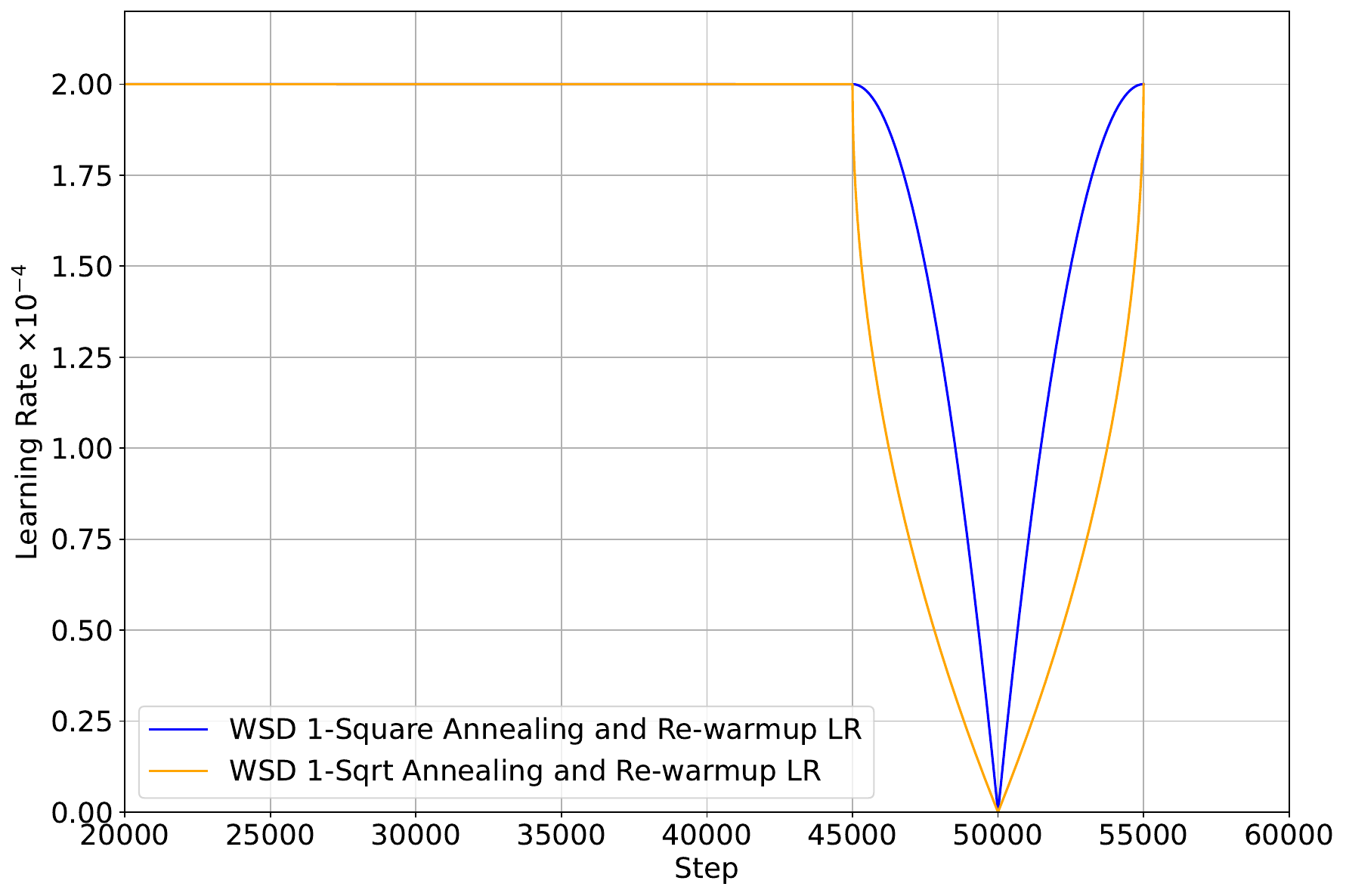}
        \caption{Learning rate curves. }
    \end{subfigure}
    \hfill
    \begin{subfigure}[b]{0.32\textwidth}
        \includegraphics[width=\textwidth]{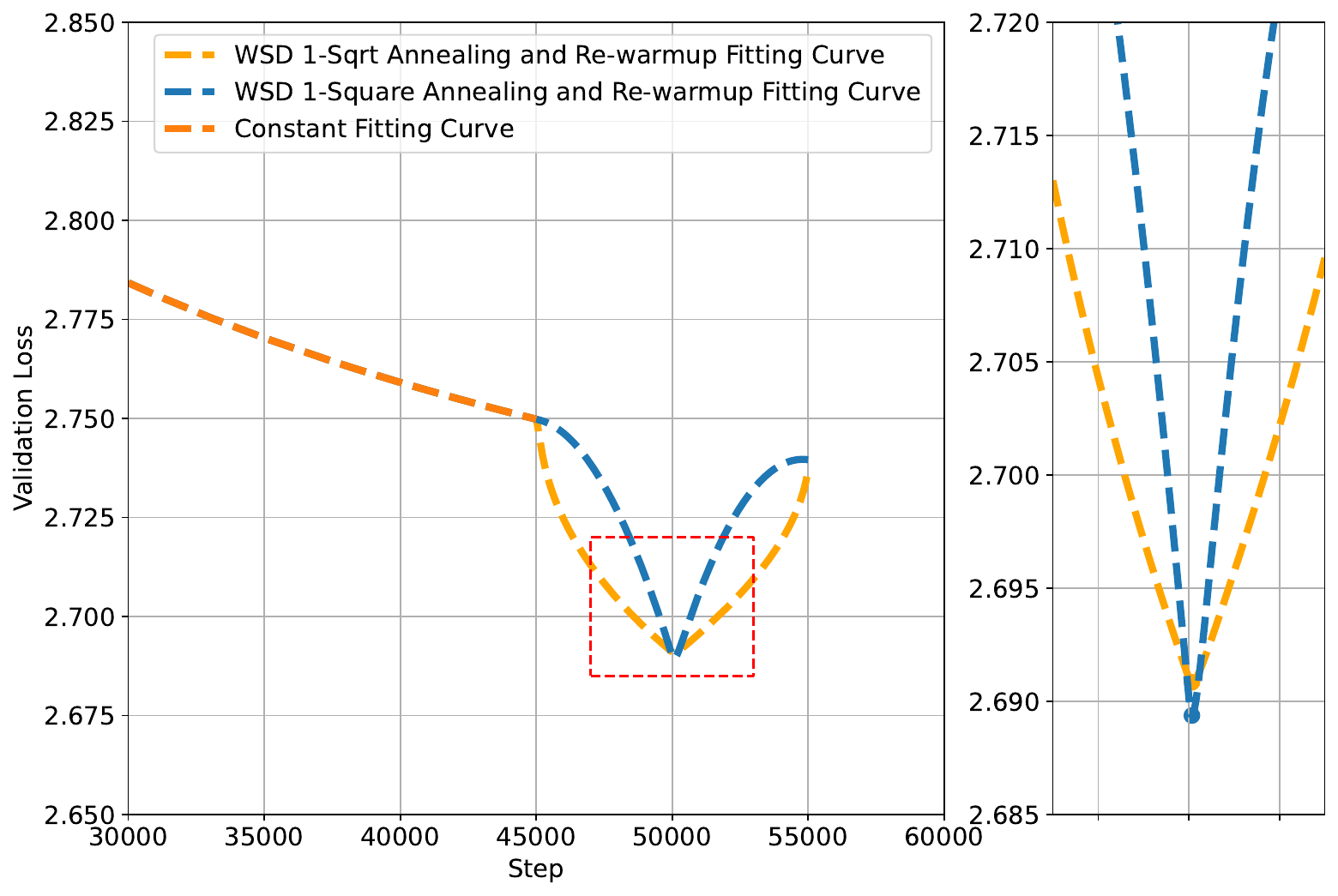}
        \caption{The loss curve of $\lambda = 0.99$.}
        \label{fig:lambda99}
    \end{subfigure}
    \hfill
    \begin{subfigure}[b]{0.32\textwidth}
        \includegraphics[width=\textwidth]{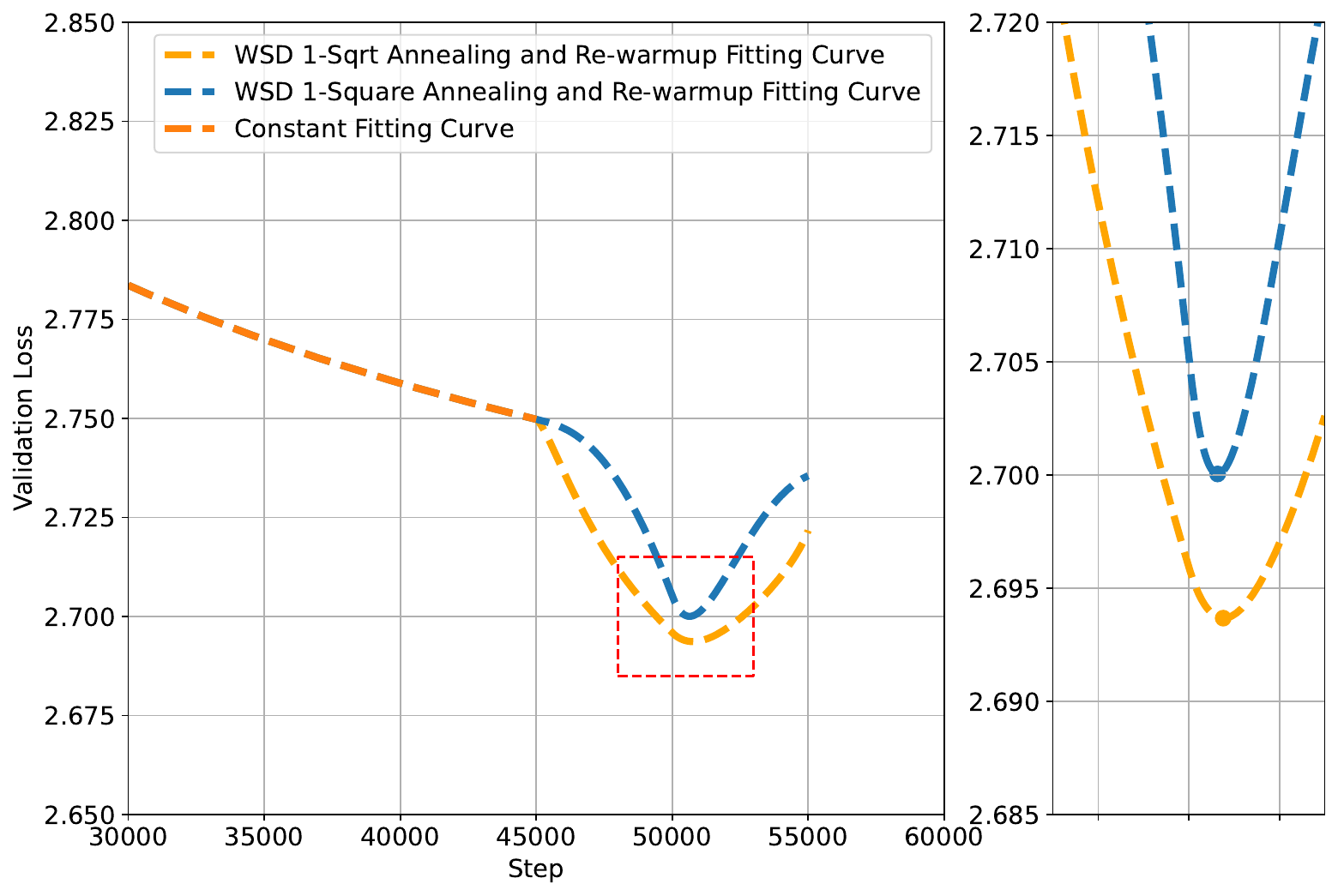}
        \caption{The loss curve of $\lambda = 0.999$. }
        \label{fig:lambda999}
    \end{subfigure}
    \caption{The comparison of fitting effect of different decay factor $\lambda$.}
    
\label{fig:lambda}
\end{figure}

\begin{figure}[tbp]
    \centering
    \begin{subfigure}[b]{0.6\textwidth}
        \includegraphics[width=\textwidth]{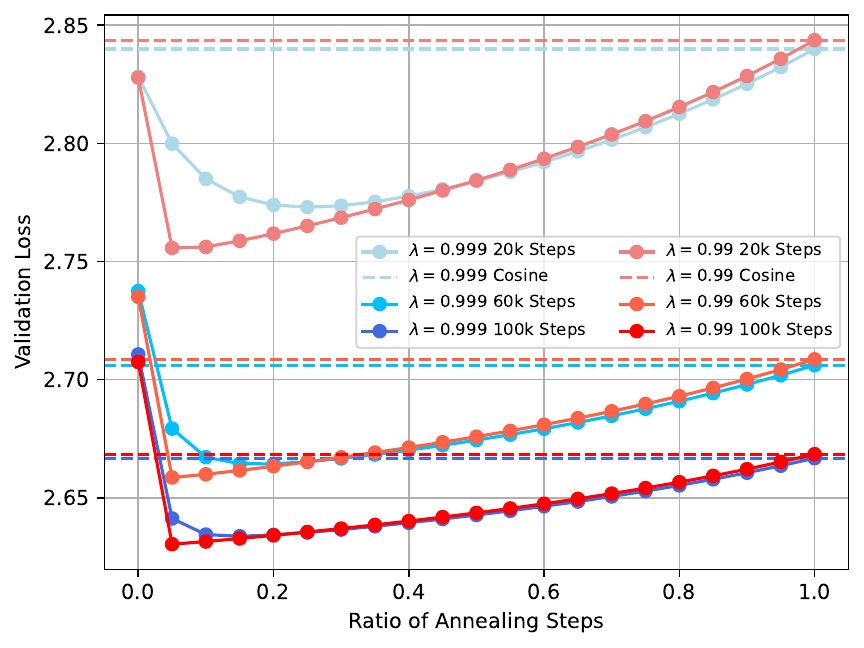}
    \end{subfigure}
    \caption{The predicted loss in different annealing ratios of WSD LRS for $\lambda=0.99$ and $\lambda=0.999$.}
\label{fig:wsd_lambda}
\end{figure}

The decay factor $\lambda$ in our equation indicates the information retaining degree in LR annealing. We set $\lambda$ as $0.999$ in our all experiments. We explore the difference from another decay factor $\lambda=0.99$. We fit and get different equations for different $\lambda$. We compare them (1) on the predicted loss curves for 1-square and 1-sqrt annealing methods, and (2) on the predicted loss curves in different annealing ratios of WSD LRS (cosine annealing).

The results, illustrated in Fig.~\ref{fig:lambda} and~\ref{fig:wsd_lambda}, reveal several key insights into the impact of decay factor:

\paragraph{Delay Steps.} A larger decay factor results in longer delay steps. Comparing 
Fig.~\ref{fig:lambda99} and Fig.~\ref{fig:lambda999}, $\lambda=0.999$ introduces a more obvious delay phenomenon, which is consistent across both the 1-square and 1-sqrt annealing methods.
The root reason is simple: larger $\lambda$ can retain more LR historical momentum, causing longer delay steps after LR finish annealing.

\paragraph{Optimal Annealing Ratio.} a larger decay factor tends to favor a higher annealing ratio. As shown in Fig.~\ref{fig:wsd_lambda},  
The optimal annealing ratio of $\lambda=0.999$
is larger than that of $\lambda=0.99$. Meanwhile, due to the similar reason, $\lambda=0.999$ favors 1-sqrt annealing method while $\lambda=0.99$ favors 1-square annealing method, as shown in Fig.~\ref{fig:lambda}.

\paragraph{Balance Point between $S_1$ and $S_2$.} More essentially, the selection of $\lambda$ decides the balance point of $S_1$ and $S_2$.
For example,
$\lambda=0.999$ means that LR annealing only retain the information of previous approximately $\frac{1}{1-\lambda}=1000$ steps, which can be seen as the window size of LR annealing momentum. The window size could be very close to the optimal annealing steps if LR changes smoothly over steps. After reaching window size, $S_2$ increases very slowly, with the cost of large decrease of $S_1$.

The analyses above highlights the importance of selecting a decay factor that aligns closely with empirical data to ensure the accuracy of predictions. 
We recommend that the future developers try different $\lambda$ for their own setups~\footnote{Actually, $\lambda$ can be fitted as a parameter, instead of a hyper-parameter requiring manual tuning. We regard $\lambda$ as a hyper-parameter because $\lambda=0.999$ performs well in our all experiments. Besides, fitting with $\lambda$ could bring in additional time complexity due to the recomputation of $S_2$.}.

\subsection{Possible Root Reasons of Delay Phenomenon in Learning Rate Annealing}
\label{sec:discsussion-delay}

In Sec.~\ref{sec:theorem}, we discover the delay phenomenon, which proves that LR annealing has momentum. 
We discuss possible root reasons of the phenomenon in this section.

\paragraph{Adam Optimizer? No.} We notice that Adam optimizer~\citep{adam-optim} also has the first-order momentum decay factor $\beta_1$ and the second-order momentum decay factor $\beta_2$, which presents the possible connection to the the delay phenomenon.

\begin{figure}[tbp]
    \centering
    \begin{subfigure}[b]{0.48\textwidth}
        \includegraphics[width=\textwidth]{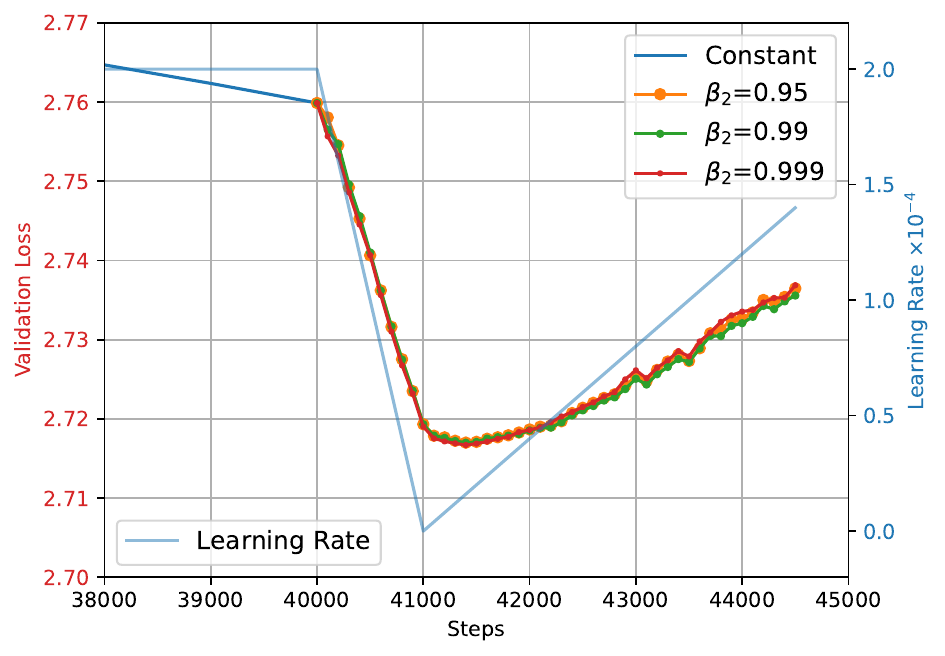}
        \caption{The comparison of true loss curve with setting different $\beta_2$ of Adam optimizer.}
        \label{fig:adam_diff}
    \end{subfigure}
    \hfill
    \begin{subfigure}[b]{0.48\textwidth}
        \includegraphics[width=\textwidth]{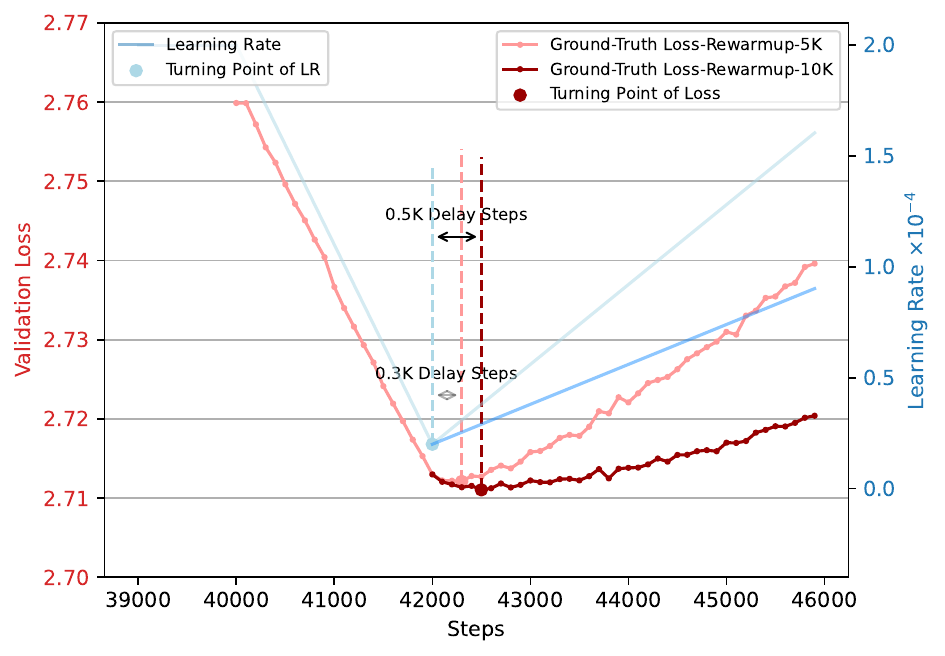}
        \caption{The comparison of delay steps of  different re-warmup steps (and thus different $S_1$).}
        \label{fig:delay-exp-apx}
    \end{subfigure}
    \caption{The possible root reason analysis (left: Adam optimizer, right: $S_1$) of delay phenomenon.}
\label{fig:root-reason}
\end{figure}

We keep $\beta_1=0.9$, and conduct delay experiments
on different $\beta_2 \in \{0.95, 0.99, 0.999\}$ (default: $0.95$) of AdamW optimizer~\citep{loshchilov2017decoupled} to observe whether larger $\beta_2$ causes a more longer delay steps.
The learning rate and ground-true loss curve are shown in Fig.~\ref{fig:adam_diff}. It suggests that the ground-truth loss curves of different $\beta_2$ almost coincide with each other, and their delay steps are also the same. Therefore, we believe that Adam optimizer has little to do with the delay phenomenon, despite its momentum form seeming very related to our experiments. Speaking of which, we even once tried to mimic the form of Adam Optimizer to describe LR annealing momentum, attempting to discover a connection between them, but the fitting results were a mess.

% \begin{figure}[tbp]
%     \centering
%     \begin{subfigure}[b]{0.7\textwidth}
%         \includegraphics[width=\textwidth]{images/adma_diff.pdf}
%     \end{subfigure}
%     \caption{The comparison of true loss curve with setting different $\beta_2$ of Adam optimizer.}
% \label{fig:adam_diff}
% \end{figure}

% \begin{figure}[tbp]
%     \centering
%     \begin{subfigure}[b]{0.7\textwidth}
%         \includegraphics[width=\textwidth]{images/decrease_increase.pdf}
%     \end{subfigure}
%     \caption{The comparison of delay steps of  different re-warmup steps.}
% \label{fig:delay-exp-apx}
% \end{figure}

\paragraph{Forward Area $S_1$? Not Really.} No matter how LR changes, $S_1$ is always increasing over steps, resulting in consistently reducing the validation loss brought from $S_1$. Therefore, the forward area, $S_1$ would lengthen delay steps in LR annealing then re-warmup, but would shorten delay steps in LR re-warmup then annealing. The delay phenomenon is indeed related to $S_1$.

But still, $S_1$ is not all the reasons of delay phenomenon. We prove this by Fig.~\ref{fig:delay-rewarmup-then-anneal}, which suggests that even though in LR re-warmup then annealing, the delay phenomenon, while not that pronounced, still exists. Moreover, we conduct delay experiments by adjusting the slope of LR after tuning point of LR. As shown in Fig.~\ref{fig:delay-exp-apx}, We find that more smooth slope of LR re-warmup, with smaller $S_1$, but still causes longer delay steps. Therefore, we conclude that $S_1$ indeed influences the specific delay length, but is not the root reason.

\paragraph{Other Possible Reasons?} The delay phenomenon could be intuitive in some cases. For example, suppose that learning rate decreases directly from 2e-4 to 2e-5 in one step, and then maintains 2e-5. In this case, although the loss would decrease to a lower value but the parameter changes in one step is too small in neural networks. Given a sudden low LR, neural networks still require some steps to gradually optimize to a local minimum, incurring delay phenomenon. 
But still, analysis above still ends with a rough description, and we have not figured out the root reasons and look forward to completing this part in future work.

\subsection{Other Possible Scaling Law Formats with LR annealing}
\label{apx:discsussion-format}
\paragraph{Adding a LR-weighted Coefficient to $S_2$?}
Imagine that when LR anneals to nearly $0$, the neural network's parameters almost do not change and the validation loss should not change, either.
However, as defined in our equation, Eq.~\ref{eq:scaling}, $S_2$ still has historical momentum even if LR is nearly $0$, making the loss continue to decrease and misalign with observed training dynamics.

To cover this corner case, we try a revision to our equation and add a LR-weighted coefficient to $S_2$. Specifically, we adjust $S_2$ to more approach $0$ when $\eta$ is close to $0$, counteracting the original formulation's tendency to overestimate loss reduction when $\eta\approx0$. 

% This modification aims to more accurately reflect the diminishing effectiveness of learning rate changes as they near zero, aligning the predicted loss curve more closely with observed training behaviors.

The revised equation for the annealing area \(S_2\) in our scaling law function is as follows:
\begin{equation}
\label{eq:scaling_withlr}
\begin{aligned}
m_i &= \lambda\cdot m_{i-1} + (\eta_{i-1} - \eta_{i}) \\
&= \sum\limits_{k=1}^{i}(\eta_{k-1} - \eta_{k})\cdot \lambda^{i-k}, \\
S_2 &= \sum\limits_{i=1}^{s} m_i \color{red}\cdot \eta_{i}^{\epsilon},
\end{aligned}
\end{equation}
where the red part is the added LR-weighted coefficient and $\epsilon$ is a undetermined positive constant. $\epsilon$ could be very small in practice.

We have tried the revised function to fit data. We find that the fitting results are quite similar and $\epsilon$ is very close to 0, showing little use in practical effect. Hence, we adopt the original format in our experiments. However, we still recommend future developers to try this format if possible.

\paragraph{$L \propto S_2^{\zeta}$ rather than {$L \propto S_2$}?}

\begin{figure}[tbp]
    \centering
    \begin{subfigure}[b]{0.32\textwidth}
        \includegraphics[width=\textwidth]{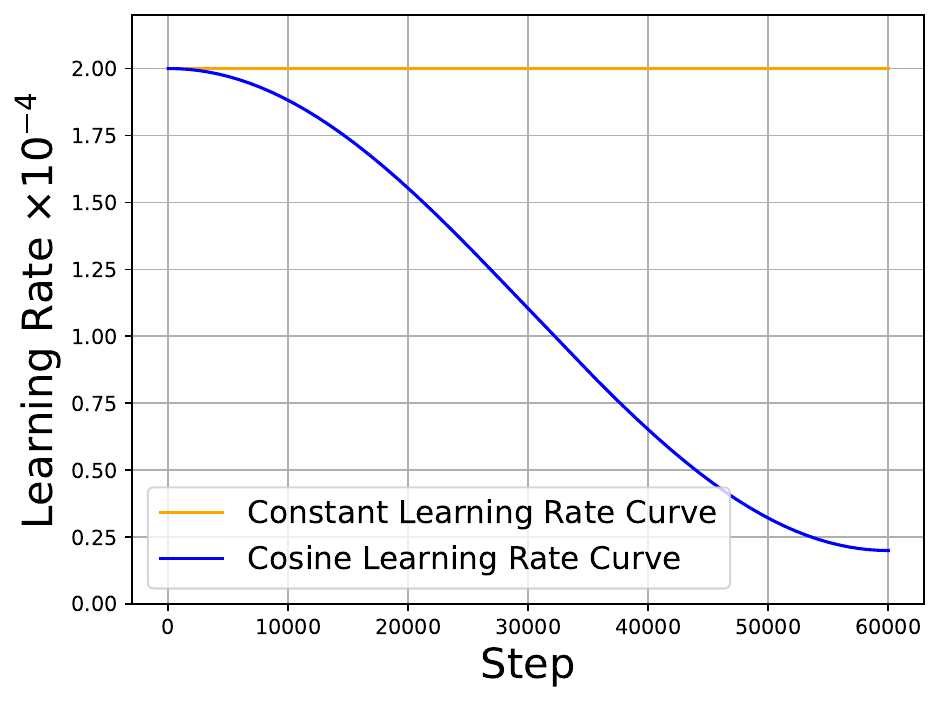}
        \caption{LR of cosine and constant. }
    \end{subfigure}
    \hfill
    \begin{subfigure}[b]{0.32\textwidth}
        \includegraphics[width=\textwidth]{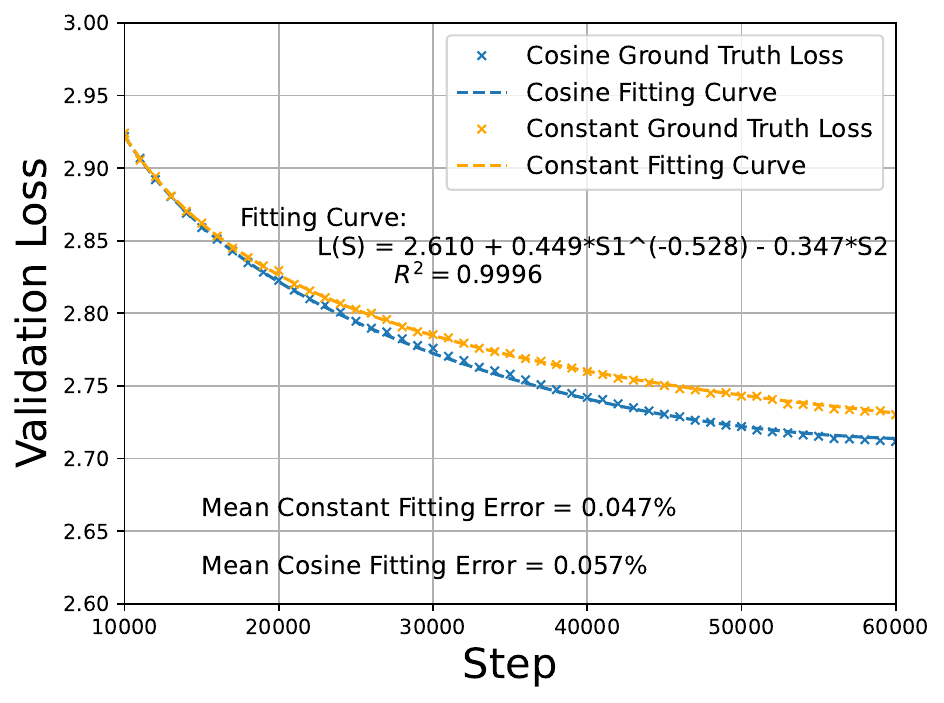}
        \caption{The fitting curve of $L \propto S_2$}
    \end{subfigure}
    \hfill
    \begin{subfigure}[b]{0.32\textwidth}
        \includegraphics[width=\textwidth]{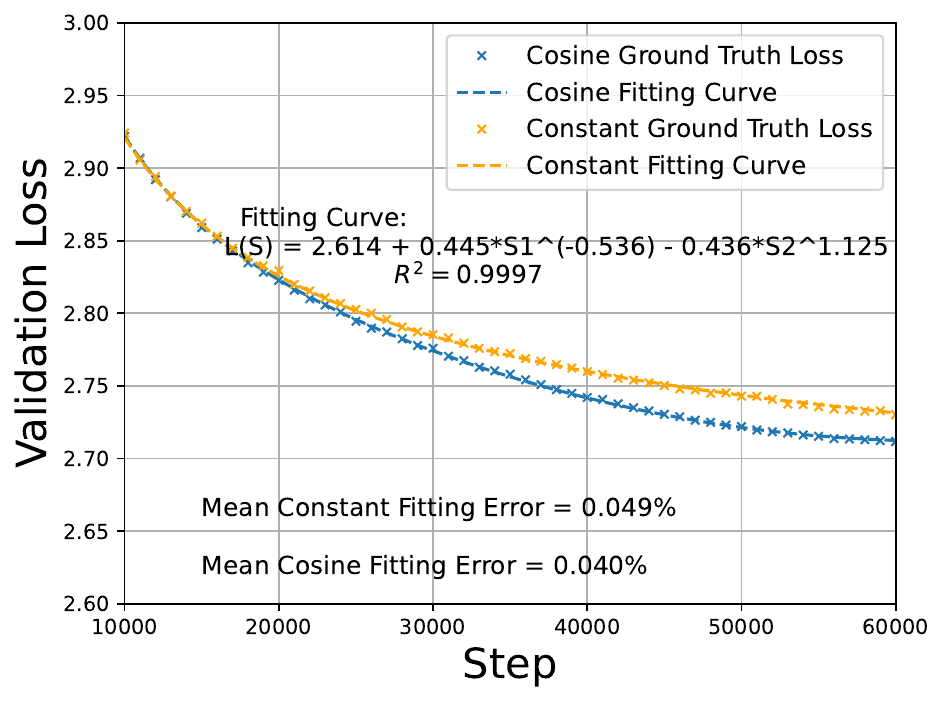}
        \caption{The fitting curve of $L \propto S_2^{\zeta}$. }
        \label{fig:have-zeta}
    \end{subfigure}
    \caption{The comparison of fitting effect between $L \propto S_2^{\zeta}$ with $L \propto S_2$.}
    
\label{fig:zeta}
\end{figure}

Actually, all we know is that $L$ and $S_2$ have a positive correlation. Thus $L\propto S_2^\zeta$ rather than $L \propto S_2$ might be a more reasonable assumption. 
That is, our equation would be changed to 
$L(s) = L_0 + A\cdot S_1^{-\alpha} - C\cdot S_{2}^{{\color{red} \zeta}} $.
Theoretically, the introduction of \(\zeta\) as an additional fitting parameter is expected to provide a more nuanced control over how changes in the learning rate annealing affect validation loss, potentially leading to improve the accuracy of our equation. 

However, the empirical results, as depicted in Fig.~\ref{fig:zeta}, demonstrate that the fitting improvement with the inclusion of \(\zeta\) is quite marginal when compared to the version without this parameter. This slight enhancement does not justify the additional complexity introduced by managing negative values of \(S_2\). Furthermore, the empirical observation that \(\zeta\) tends to converge close to $1$ (e.g. $1.125$ in Fig.~\ref{fig:have-zeta}) reinforces the notion that the original formulation of the function, without the power term \(\zeta\), is adequately robust. This finding suggests that the direct influence of the learning rate annealing area, as initially modeled, sufficiently captures the essential dynamics without the need for this additional complexity. Another additional complexity lies in that $S_2^{\zeta}$ becomes incalculable when $S_2<0$ in LR re-warmup.

\section{Future Works}
\subsection{Tuning Scaling Law Format}
We have tested a variety of equation formats to enhance the accuracy of the entire training process. As a result, the final equation format, as presented in Eq.~\ref{eq:scaling}, proves to be optimal so far
across a range of scenarios. 
We add only one extra parameter but obtain a very good fitting and prediction result.
The formulation has achieved a level of practicality that enables the prediction of future loss when scaling training steps and model sizes. We expect more following researches to explore the format of the scaling law with learning rate annealing.

\subsection{More Applications via Our Scaling Law}
In Sec.~\ref{sec:takeaways}, we present many instances to apply our scaling law with LR annealing, to predict future training dynamics with a cost-free manner.
We believe that our equation can help analyze and select more training recipes in specific scenarios.

\subsection{Extension to Post-training}
In this work, we research primarily on the scope of pretraining of LLM. We also show how to apply our equation to guide the LR re-warmup strategy in continual pre-training. We will continue researching on how to extend our equation to post-training, which might include data distribution shift, data mixture, model alignment, and specific downstream evaluations.

\section{Conclusion}
In conclusion, we propose that the loss curves of neural language models empirically adhere to a scaling law with learning rate annealing over training steps $s$: $L(s) = L_0 + A\cdot S_1^{-\alpha} - C\cdot S_{2}$.  
This equation can accurately predict full loss curves across unseen learning rate schedulers.
We present the underlying intuition and theory for deriving our equation and demonstrate that our approach can be extended to capture the scaling effect of model sizes.
Extensive experiments demonstrate that our proposed scaling law has good accuracy, scalability, and holds under various experimental setups.
It also offers accurate theoretical insights to the training dynamics of LLMs,
and explains numerous phenomena observed in previous studies. We believe that the scaling law with LR annealing is promising to reshape the understanding of researchers for LLM training and scaling laws.

\bibliography{iclr2025_conference}
\bibliographystyle{iclr2025_conference}

\newpage
\appendix
\section{Impact of Warmup Steps}
\label{apx:warmup}
\begin{figure}[htbp]
    \centering
    \begin{subfigure}[b]{0.6\textwidth}
        \includegraphics[width=\textwidth]{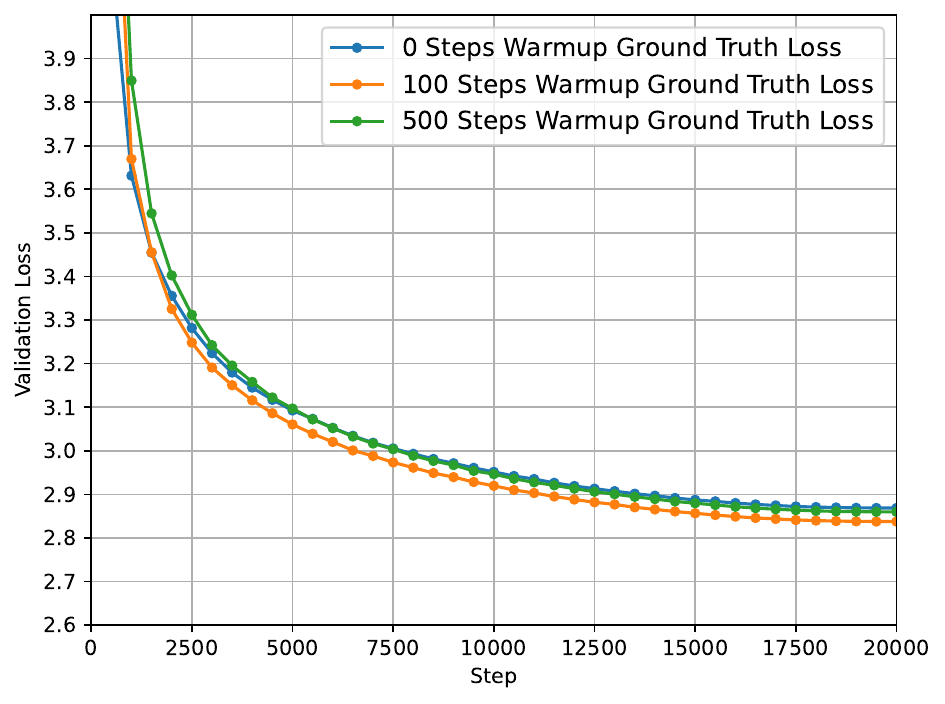}
    \end{subfigure}
    \caption{Validation loss curves corresponding to different warmup steps. We use cosine LRS with 20K total steps.}
\label{fig:warmup_diff}
\end{figure}

We investigate the impact of LR warmup steps on loss curves. Specifically, different values of warmup steps are experimented. As shown in Fig.~\ref{fig:warmup_diff}, we find that a warmup step value of 500 accelerates convergence and achieves the lowest validation loss compared to 100 or no warmup.
This finding is aligned with previous works~\citet{warmup-evidence,kosson2024analyzing}. 

Based on these findings, 
we use 500 warmup steps in our main experiments when training models from scratch.
Note that LR warmup when training a randomly initialized model is different from LR re-warmup in continual training. In fact, the re-warmup process can be seen as a special types of LRS where $S_2<0$ in our equation.

\section{Experimental Setups}
\label{apx:exp-sets}
In this work, we use multiple experimental setups to validate the effectiveness of our equation across a variety of conditions. For clarity, we summarize all experimental setups in Table~\ref{tab:exp-sets}. The majority of our experiments use Setting A. Additionally we also successfully fit our equation to the loss curves of BLOOM \citep{workshop2022bloom} and OLMo \citep{groeneveld2024olmo}. The experimental settings used to produce these loss curves are significantly different from ours. Please refer to their papers for details on their specific experimental setups.

\begin{table}[p]
\begin{center}
\caption{Experimental settings adopted in this work. Model size denotes the number of non-embedding parameters. Our datasets include Fineweb~\citep{penedo2024fineweb} and RedPajama-CC~\citep{together2023redpajama}. 
* denotes pre-training multilingual datasets including mixture of sources such as common crawls, books, arxiv, code, etc.
We use AdamW Optimizer~\citep{adam-optim,loshchilov2017decoupled}, denoted as AO.
Most experiments adopt Llama-3's tokenizer~\citep{dubey2024llama}.
Ext Llama-2's is extended from Llama-2's tokenizer ~\citep{touvron2023llama} by adding vocabulary.}
\label{tab:exp-sets}
\begin{tabular}{llll}
\toprule
\textbf{Setups} & \textbf{Setting A (main)} & \textbf{Setting B} & \textbf{Setting C}
\\
\midrule
\textbf{Model Size} & $594$M & $293$M & multiple \\
\textbf{Train Dataset} & Fineweb & Finweb & Mixture-train* \\ 
\textbf{Val Dataset} & RedPajama-CC & RedPajama-CC & Mixture-valid* \\
\textbf{Total Steps} & $60$K & $120$K & $143$K \\
\textbf{Maximal LR} & $2\times10^{-4}$ & $2\times10^{-4}$ & $1.381\times10^{-3}$ \\
\textbf{Warmup Steps} & $500$ & $100$ & $500$ \\
\textbf{Batch Size (tokens) } & $4$M & $2$M & $4$M \\
\textbf{Sequence Length} & $4096$ & $4096$ & $4096$ \\
\textbf{Tokenizer} & Llama-3's & Llama-3's & Ext Llama-2's \\
\textbf{$\beta_1$,$\beta_2$ in AO} & $0.9$, $0.9$5 &  $0.9$, $0.95$ & $0.9$, $0.95$\\
\textbf{Weight Decay}  & $0.1$ & $0.1$ & $0.1$\\
\textbf{Gradient Clip}  & $1.0$ & $1.0$ & $1.0$\\
\bottomrule  % WARNING: it cannot change to top/mid rule for unrecognized fatal bug
\toprule
\textbf{Setups} & \textbf{Setting D (MoE)} & \textbf{Setting E (1.4T tokens)} &  \\
\midrule 
\textbf{Model Size} & $8\times106$M & $1704$M &  \\
\textbf{Train Dataset} & Fineweb & Mixture-train* &  \\ 
\textbf{Val Dataset} & RedPajama-CC & Mixture-valid* &  \\
\textbf{Total Steps} & $60$K & $350$K & \\
\textbf{Maximal LR} & $2\times10^{-4}$ & $6\times10^{-4}$ & \\
\textbf{Warmup Steps} & $500$ & $1000$ & \\
\textbf{Batch Size (tokens) } & $4$M & $4$M & \\
\textbf{Sequence Length} & $4096$ & $8192$ & \\
\textbf{Tokenizer} & Llama-3's & Llama-3's &  \\
\textbf{$\beta_1$,$\beta_2$ in AO} & $0.9$, $0.9$5 &  $0.9$, $0.95$ & \\
\textbf{Top-$k$ Experts} & $2$ & - &  \\
\textbf{Auxiliary Loss} & $0.01$ & - &  \\
\textbf{Weight Decay}  & $0.1$ & $0.1$ & \\
\textbf{Gradient Clip}  & $1.0$ & $1.0$ & \\
\bottomrule
\end{tabular}
\end{center}
\end{table}

\section{Results on Extensive Experiments Setups}
\subsection{Another Set of Training Hyper-parameters}
\label{apx:exp-2}
\begin{figure}[tbp]
    \centering
    \begin{subfigure}[b]{0.32\textwidth}
        \includegraphics[width=\textwidth]{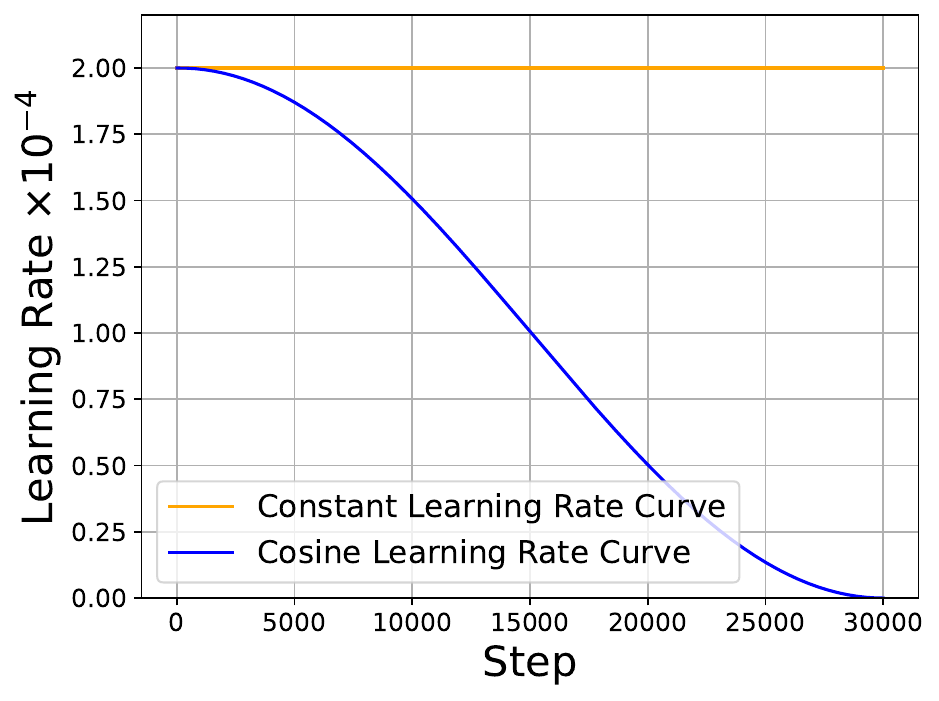}
        \caption{Learning rate.}
    \end{subfigure}
    \hfill
    \begin{subfigure}[b]{0.32\textwidth}
        \includegraphics[width=\textwidth]{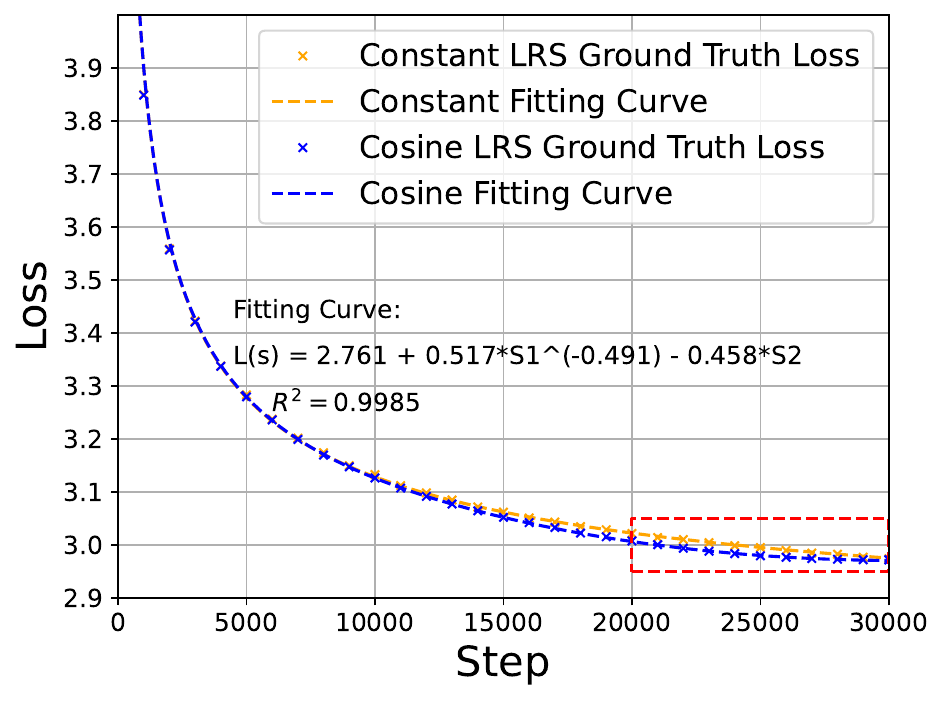}
        \caption{Zoomed-out loss fitting.}
    \end{subfigure}
    \hfill
    \begin{subfigure}[b]{0.32\textwidth}
        \includegraphics[width=\textwidth]{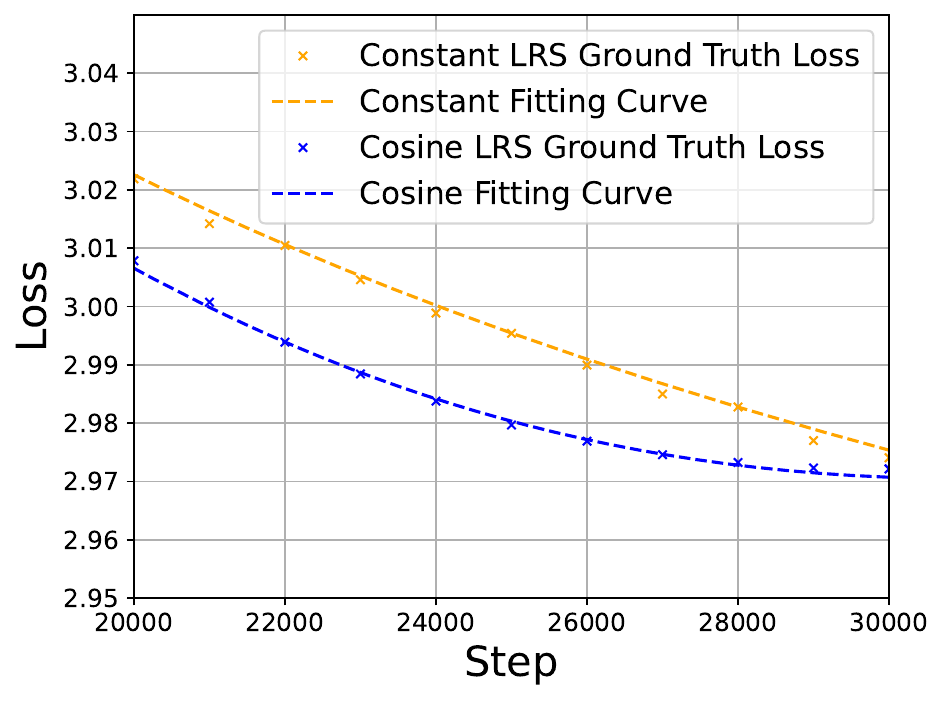}
        \caption{Zoomed-in loss fitting.}
    \end{subfigure}
    \caption{Using Eq.\ref{eq:scaling} to \textbf{fit} full loss curves yield by cosine and constant LRS. Total steps=30K, $\eta_{min}=2\times 10 ^{-4}$, $\eta_{min}=0$. We omit the warmup steps in the figure. The fitted equation is $L = 2.761 + 0.517\cdot S_1^{-0.491} - 0.458\cdot S_2$. Refer to setting B in Table~\ref{tab:exp-sets} for detailed experimental setups.}
\label{fig:fit_12w}
\end{figure}

\begin{figure}[tbp]
    \centering
    
    \begin{subfigure}[b]{\textwidth}
        \centering
        \includegraphics[width=0.32\textwidth]{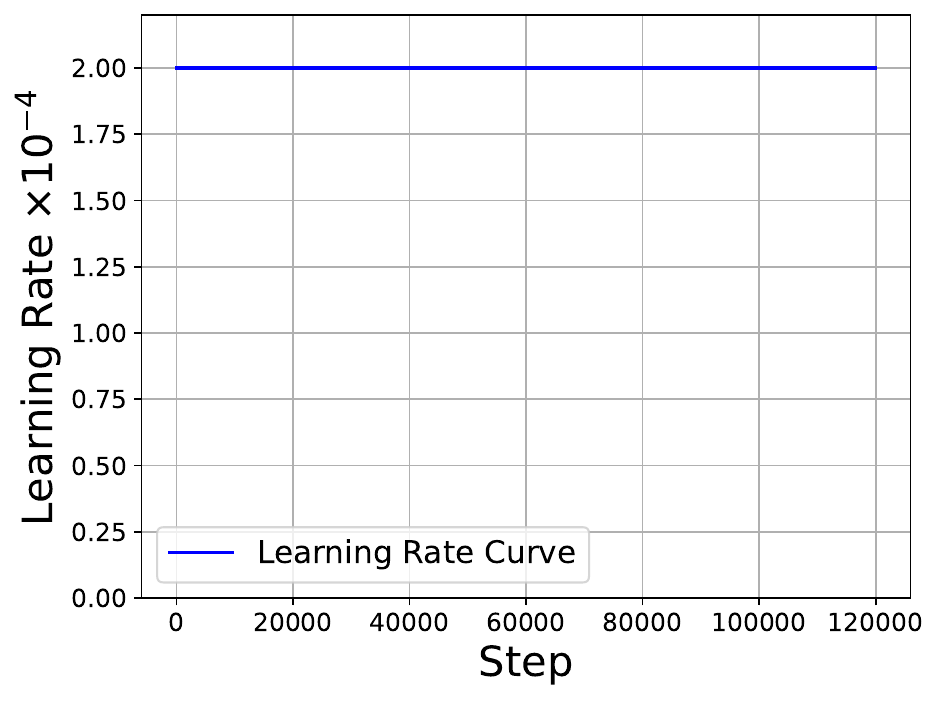}
        \includegraphics[width=0.32\textwidth]{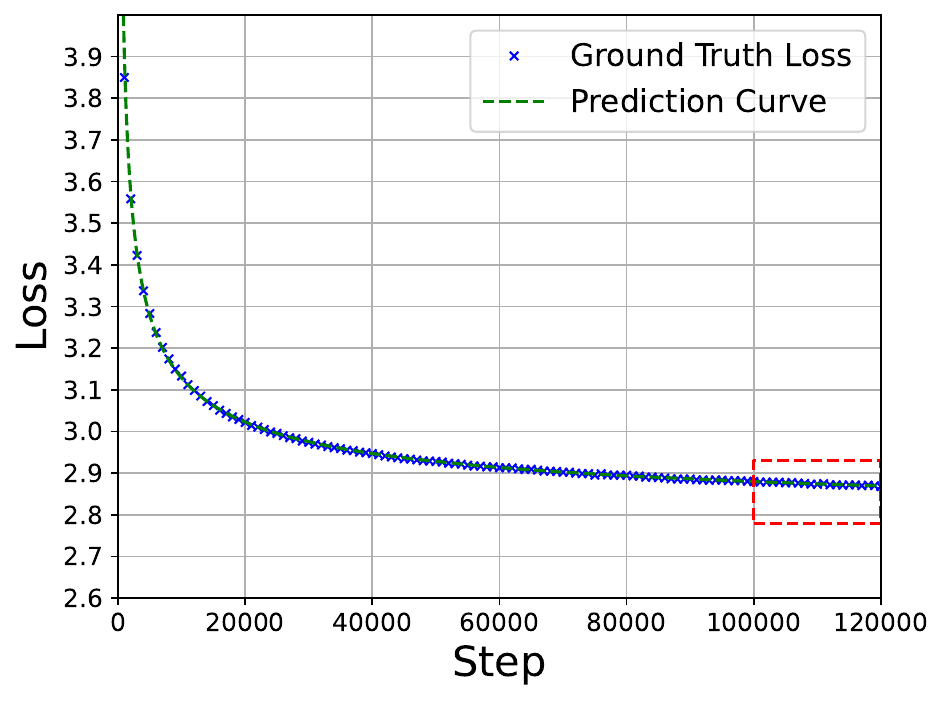}
        \includegraphics[width=0.32\textwidth]{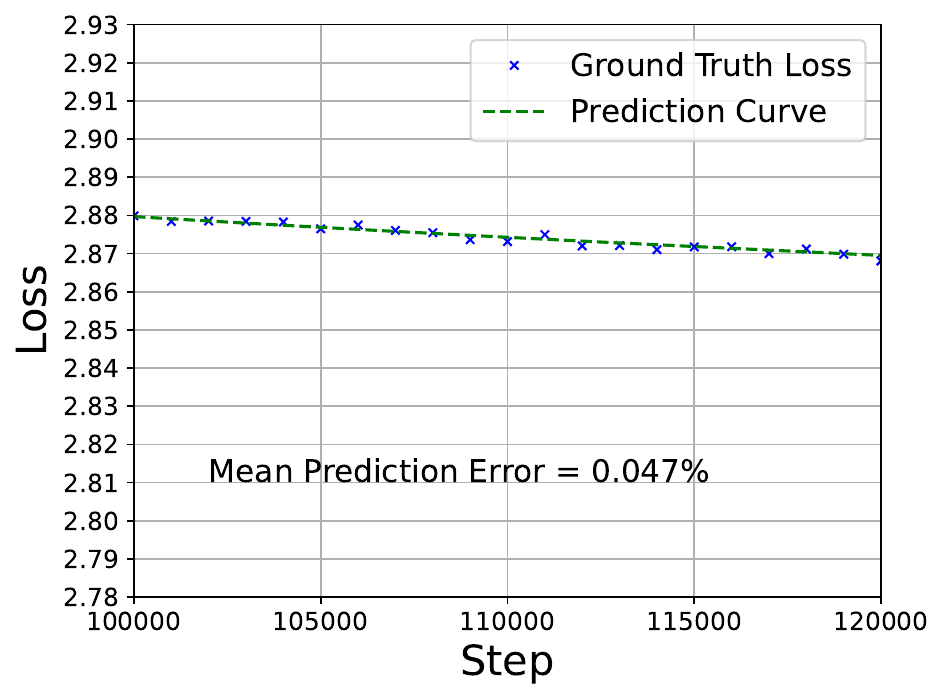}
        \caption{Full curve prediction of constant LRS.}
        \label{fig:prediction-12w-constant}
    \end{subfigure}

    \begin{subfigure}[b]{\textwidth}
        \centering
        \includegraphics[width=0.32\textwidth]{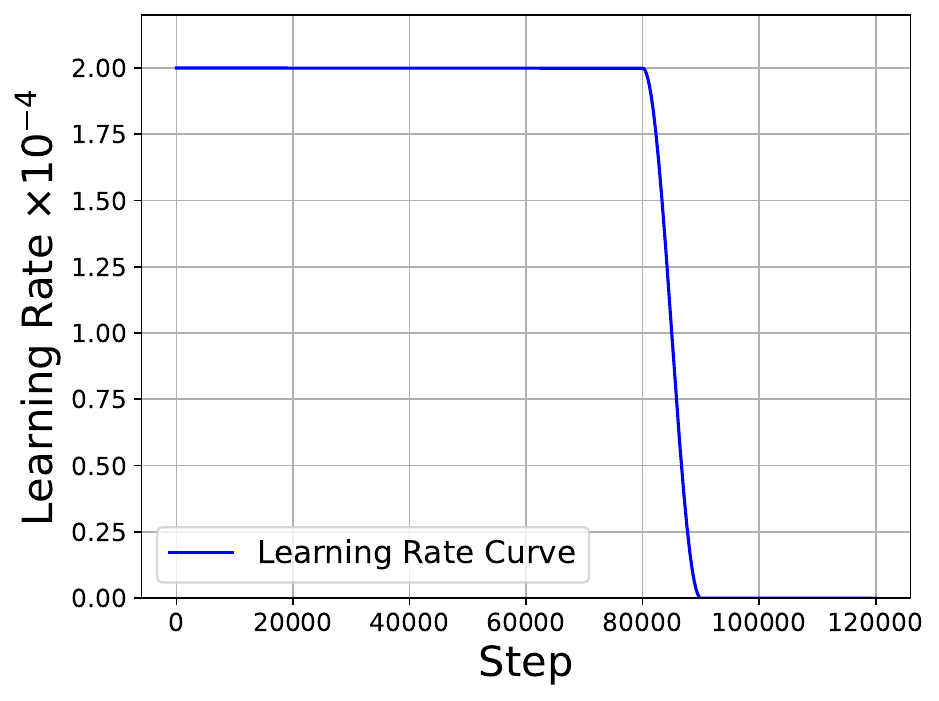}
        \includegraphics[width=0.32\textwidth]{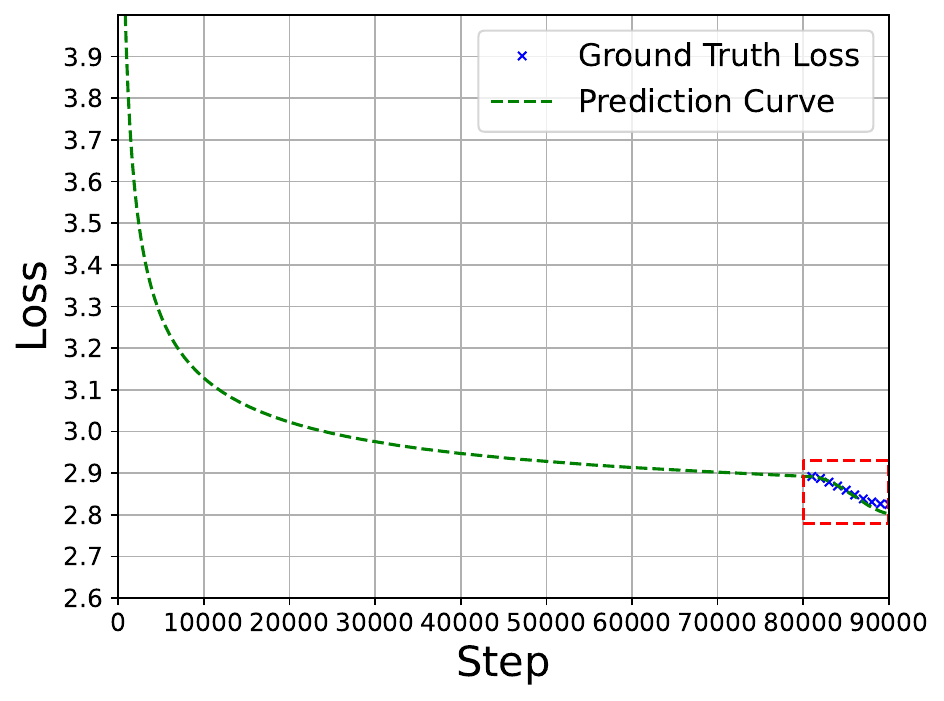}
        \includegraphics[width=0.32\textwidth]{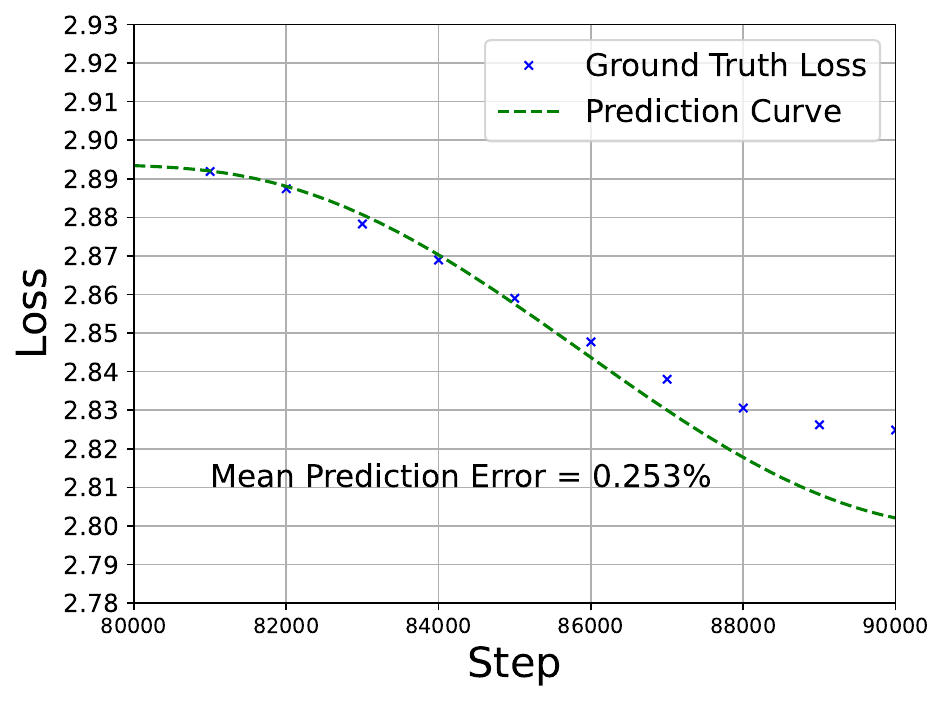}
        \caption{Full curve prediction of WSD LRS (90K total steps; 10 \% cosine annealing to $\eta_{min} = 0$).}
        \label{fig:prediction-9w-wsd}
    \end{subfigure}

    \begin{subfigure}[b]{\textwidth}
        \centering
        \includegraphics[width=0.32\textwidth]{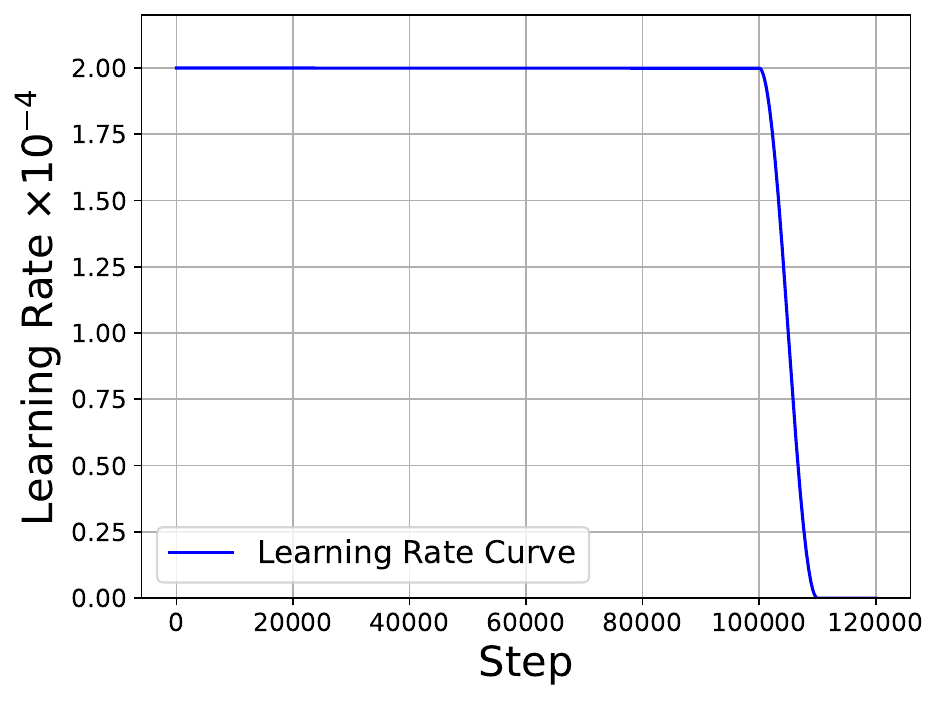}
        \includegraphics[width=0.32\textwidth]{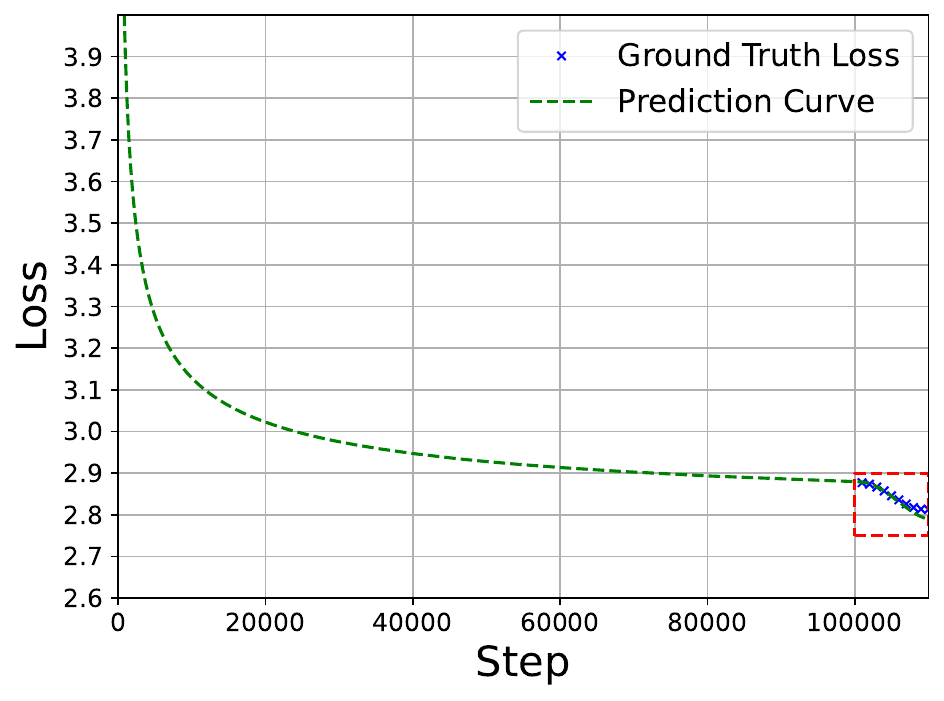}
        \includegraphics[width=0.32\textwidth]{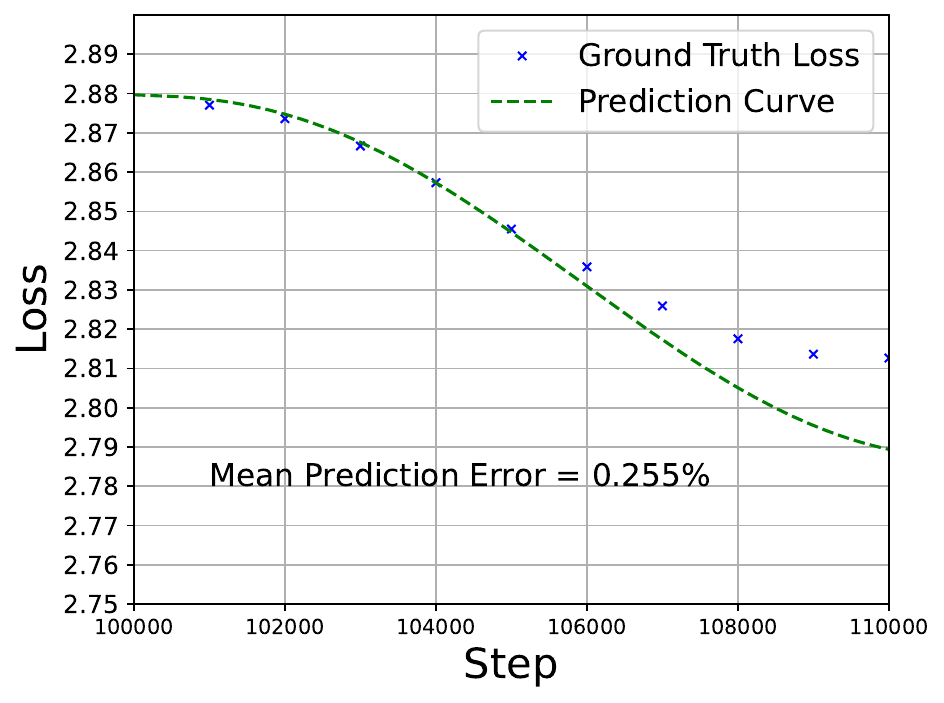}
        \caption{Full curve prediction of WSD LRS (110K total steps; 10 \% cosine annealing to $\eta_{min} = 0$).}
        \label{fig:prediction-11w-wsd}
    \end{subfigure}
    
    \caption{Using the fitted equation in Fig.~\ref{fig:fit_12w} to \textbf{predict} full loss curves for unseen LRS with 120K total steps. The left, medium, and right columns represent LR curve, the predicted loss curves, and a zoomed-in view of loss curve, respectively. The red rectangle indicates the zoomed-in zone. Warmup steps (100) are not shown in this figures. 
    Notable, all LRS and loss curves shown here were \textbf{unseen} during the fitting process. The mean prediction errors across different LRS is as low as $\sim 0.2\%$. Refer to setting B in Table~\ref{tab:exp-sets} for detailed experimental setups.}
\label{fig:predict_12w}
\end{figure}

Fig.~\ref{fig:fit} and Fig.~\ref{fig:prediction} show that our equation can work very well on our main experimental setup detailed in Table~\ref{tab:exp-sets}.
To validate the effectiveness of our scaling law formulation on different experimental settings, we report the results on Setting $B$ (see Table~\ref{tab:exp-sets}). The fitting results are shown in Fig.~\ref{fig:fit_12w}, and the prediction results are shown in Fig.~\ref{fig:predict_12w}. The results suggest that our scaling law with LR annealing works well across different experimental setups.

\newpage

\subsection{Experiments on the Mixture of Experts Architecture}
\label{apx:moe}
\begin{figure}[p]
    \centering
    \begin{subfigure}[b]{0.32\textwidth}
        \includegraphics[width=\textwidth]{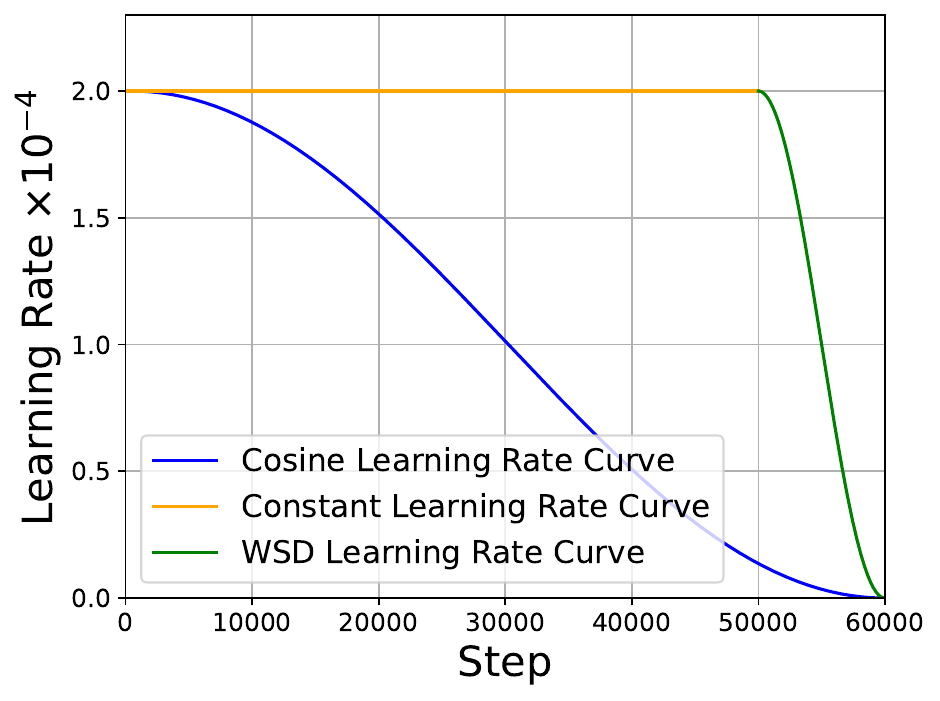}
        \caption{Learning rate.}
    \end{subfigure}
    \hfill
    \begin{subfigure}[b]{0.32\textwidth}
        \includegraphics[width=\textwidth]{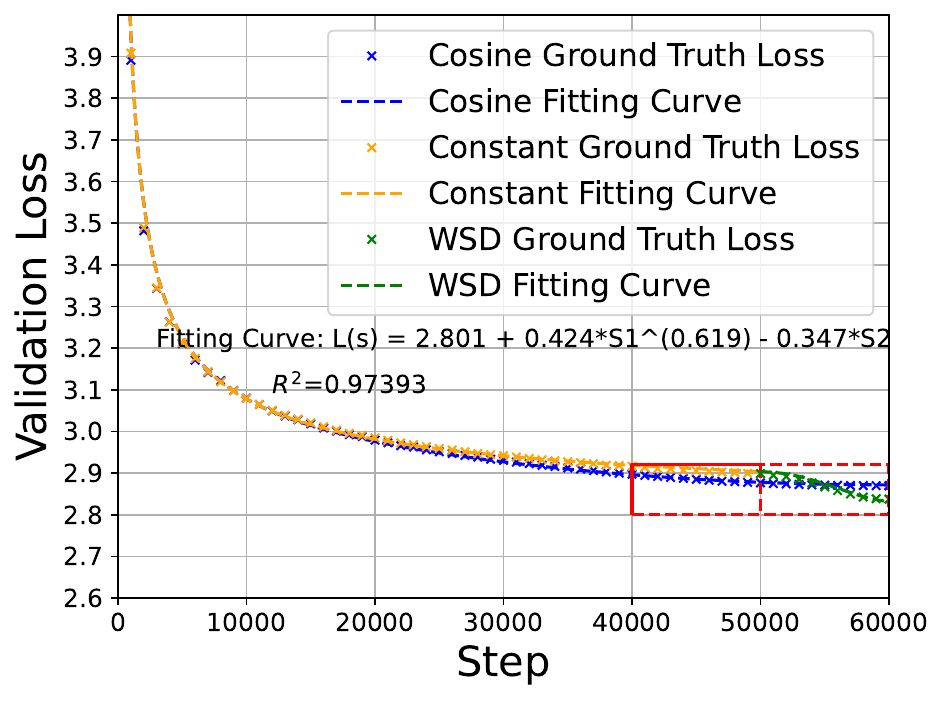}
        \caption{Zoomed-out loss fitting.}
    \end{subfigure}
    \hfill
    \begin{subfigure}[b]{0.32\textwidth}
        \includegraphics[width=\textwidth]{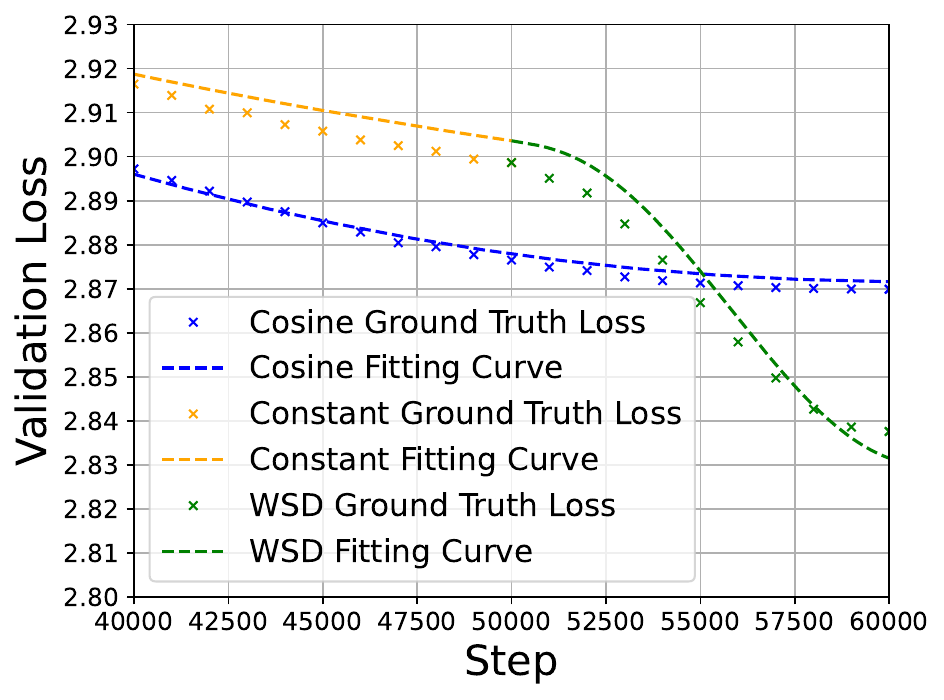}
        \caption{Zoomed-in loss fitting.}
    \end{subfigure}
    \caption{Full loss curve \textbf{fitting} for MoE models. After fitting, we get a \textbf{universal} loss equation $L = 2.801 + 0.424\cdot S_1^{-0.619} - 0.347\cdot S_2$. Refer to setting D in Table~\ref{tab:exp-sets} for detailed experimental setups.}
\label{fig:moe_fit}
\end{figure}

\begin{figure}[tbp]
    \centering
    
    \begin{subfigure}[b]{\textwidth}
        \centering
        \includegraphics[width=0.32\textwidth]{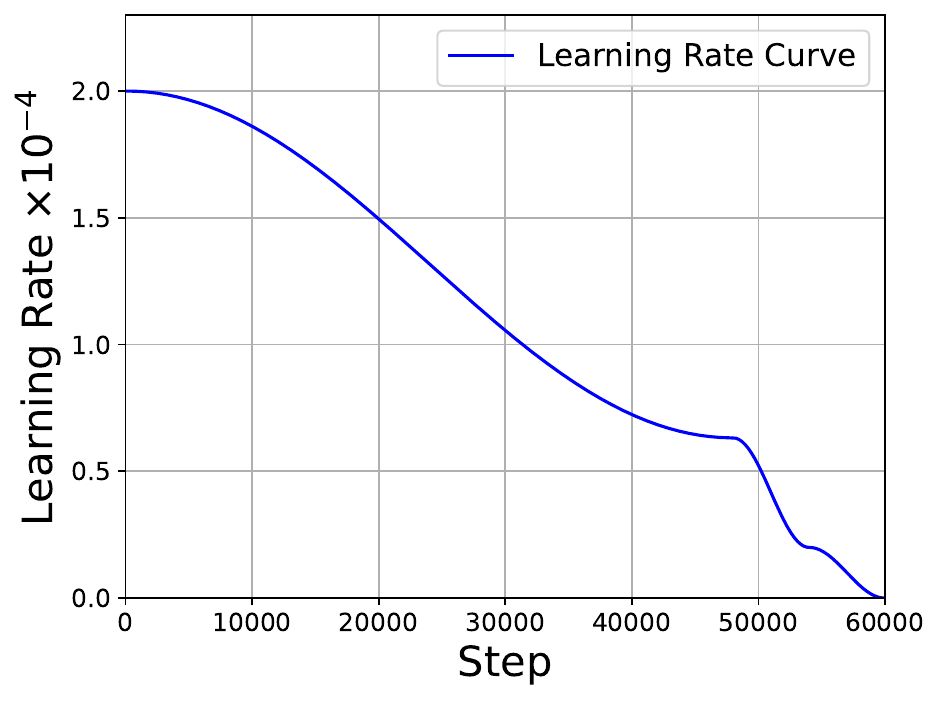}
        \includegraphics[width=0.32\textwidth]{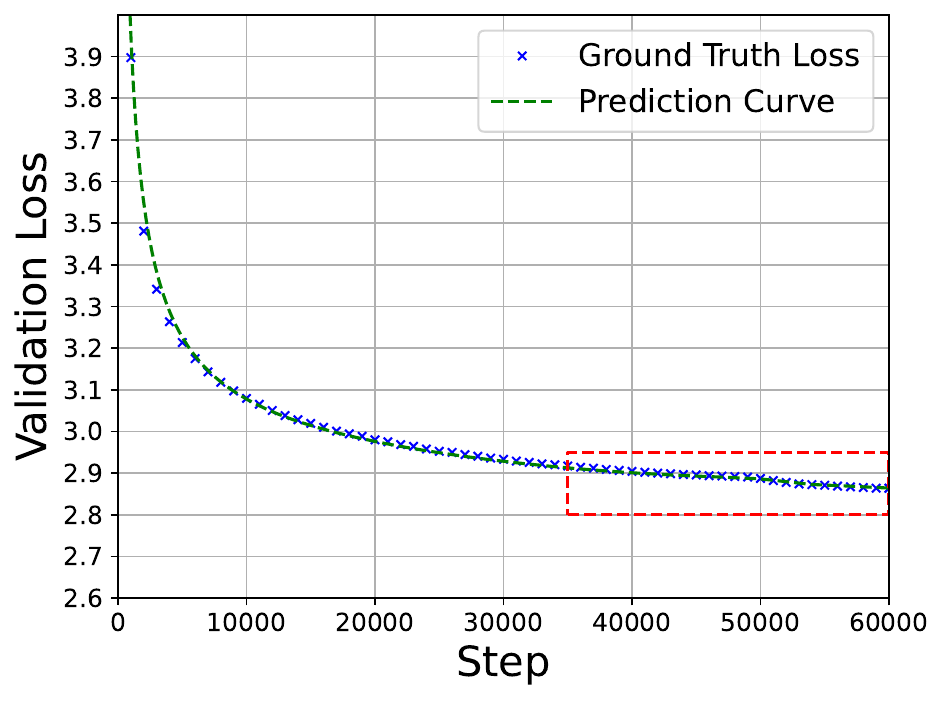}
        \includegraphics[width=0.32\textwidth]{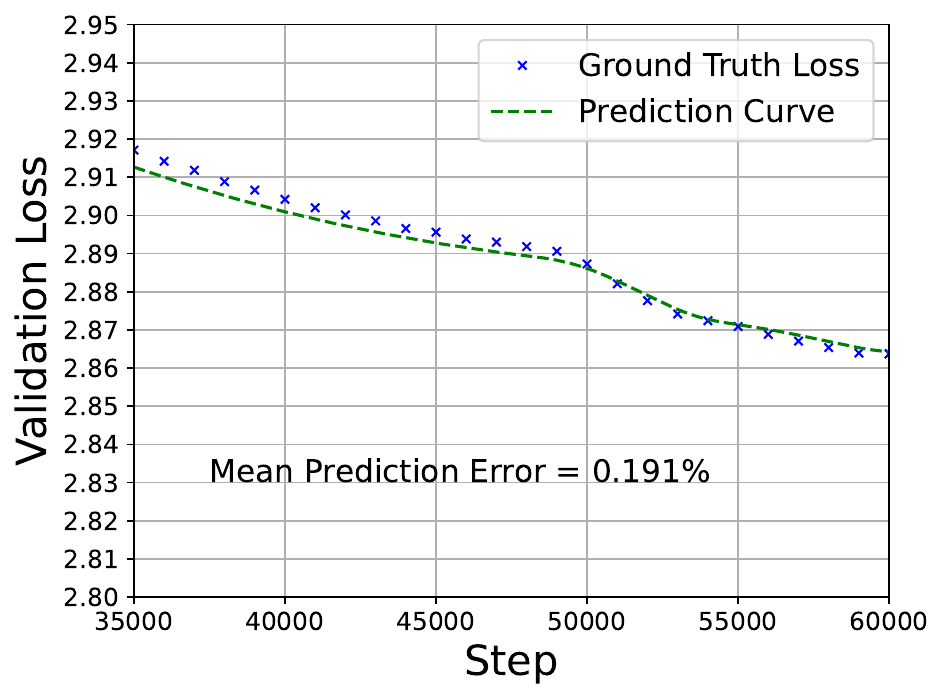}
        \caption{Full curve prediction of multi-step cosine LRS (80\% + 10\% + 10\%)~\citep{deepseek-ai2024deepseek}}
        \label{fig:moe_prediction-multi-cosine}
    \end{subfigure}

    \begin{subfigure}[b]{\textwidth}
        \centering
        \includegraphics[width=0.32\textwidth]{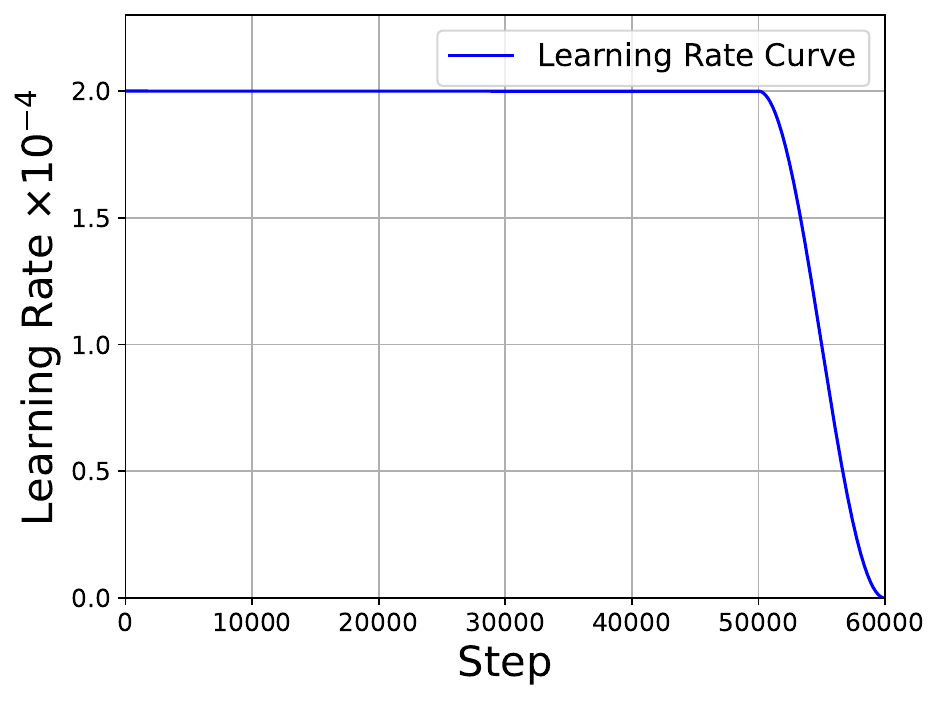}
        \includegraphics[width=0.32\textwidth]{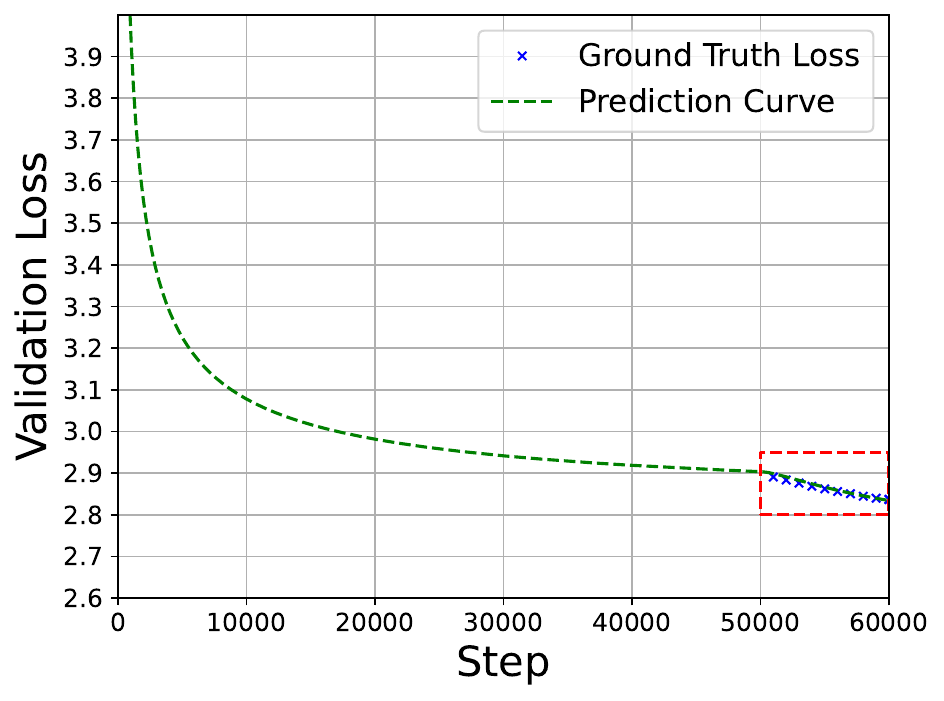}
        \includegraphics[width=0.32\textwidth]{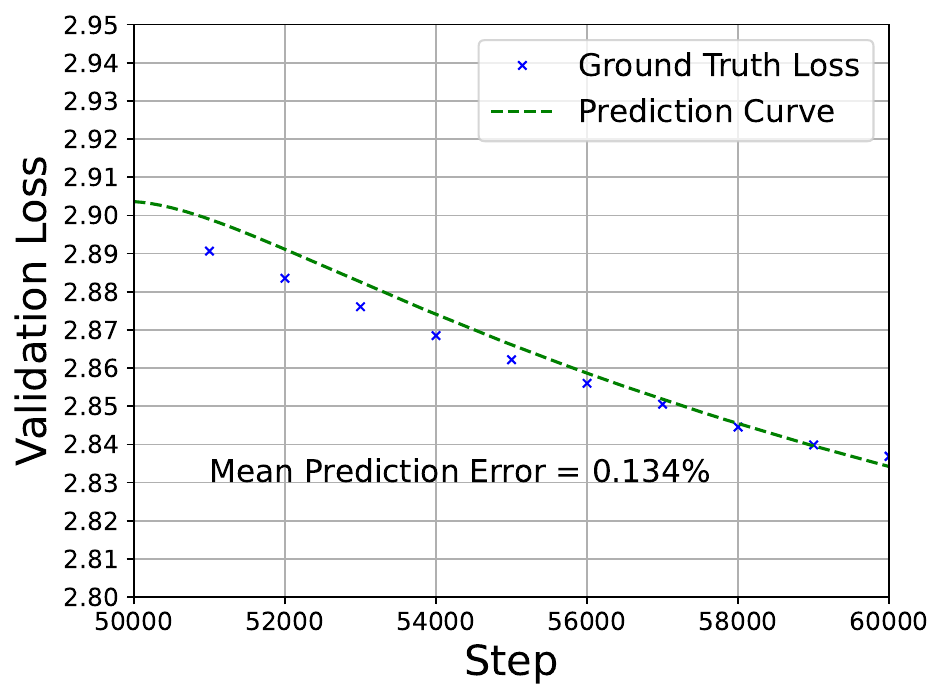}
        \caption{Full curve prediction of WSD LRS (17\% exponential annealing to $\eta_{min} = 0$)~\citep{hu2024minicpm}.}
        \label{fig:moe_prediction-wsd}
    \end{subfigure}
    
     \begin{subfigure}[b]{\textwidth}
        \centering
        \includegraphics[width=0.32\textwidth]{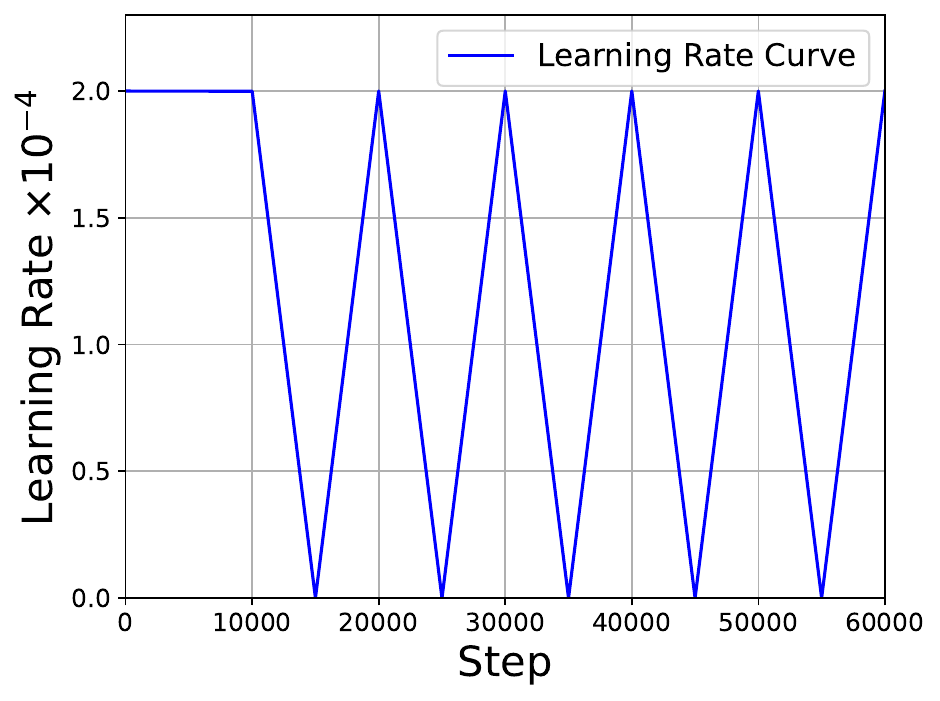}
        \includegraphics[width=0.32\textwidth]{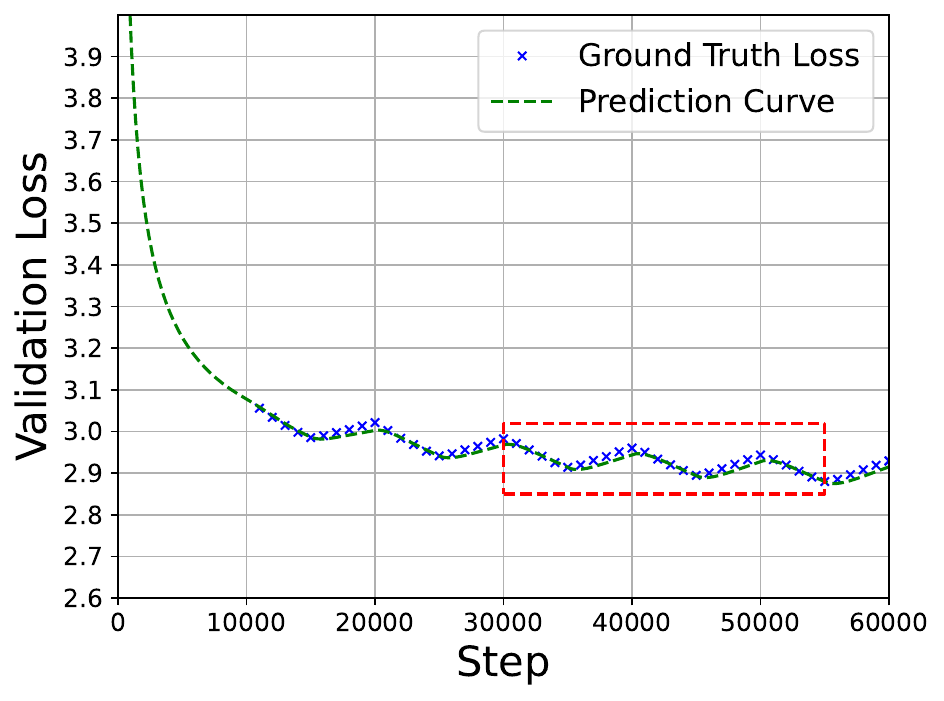}
        \includegraphics[width=0.32\textwidth]{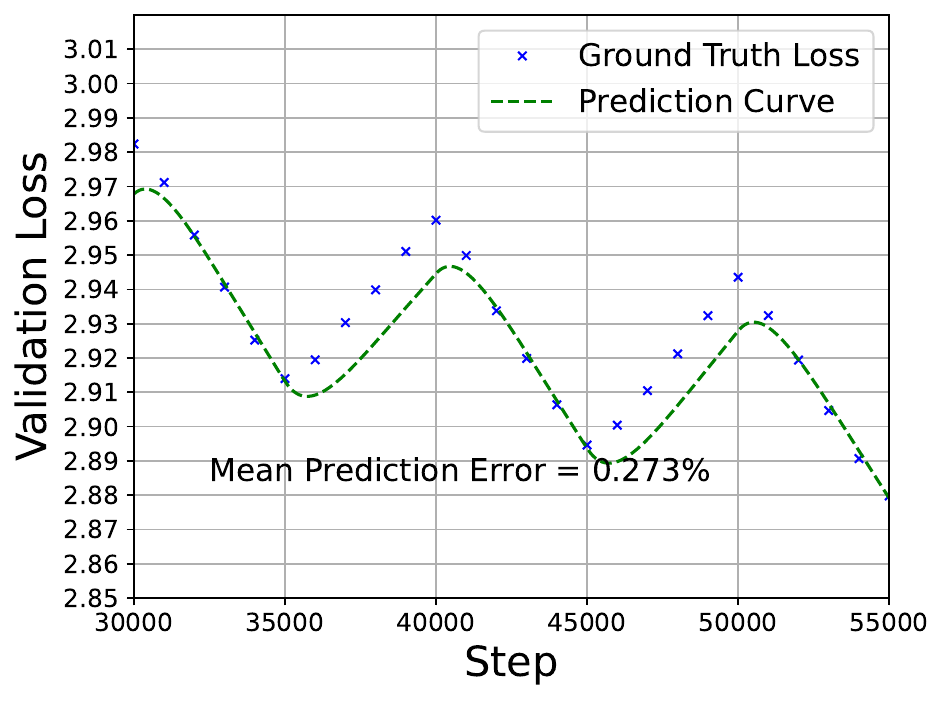}
        \caption{Full curve prediction of Cyclic LRS~\citep{smith2017cyclical} including multiple unseen LR re-warmup stages.}
        \label{fig:moe_prediction-strange}
    \end{subfigure}
    
    \caption{Using the fitted equation in Fig.~\ref{fig:moe_fit} to \textbf{predict} full loss curves for unseen LRS (MoE model). The left, medium, and right columns represent LR curve, the predicted loss curves and a zoomed-in view of loss curves, respectively. 
    The red rectangle indicates the zoomed-in zone. Warmup steps (500) are not shown in this figure. 
    Noteable, all LRS and loss curves shown here were unseen during the fitting process. The mean prediction errors across different LRS is as low as $\sim 0.2\%$. Refer to setting D in Table~\ref{tab:exp-sets} for detailed experimental setups.
    }
\label{fig:moe_prediction}
\end{figure}

% Fig.~\ref{fig:fit} and Fig.~\ref{fig:prediction} show that our equation can work very well on the dense Llama-like architecture~\citep{vaswani2017attention,touvron2023llama}.
We validate our equation on the Mixture of Experts (MoE) architecture.
The experimental setting is shown as setting $D$ in Table~\ref{tab:exp-sets}.
We add the widely-used auxiliary loss for load balancing among experts~\citep{fedus2021switch}. Moreover, we change the LRS and total steps to 60K WSD with 10K annealing steps in fitting, testing whether our scaling law is effective
under various circumstances.
The fitting results shown in Fig.~\ref{fig:moe_fit} and the prediction results shown in Fig.~\ref{fig:moe_prediction} indicates that our equation still works well on the MoE architecture.

\subsection{Scaling Up for Much Longer Steps}
\label{apx:long-exp}
In this section, we validate the effectiveness of our equation in predicting loss curves over significantly long total steps. This scalability is particularly useful in large-scale training.
Specifically, we fit our equation on loss curves yield by the constant and cosine LRS with 30K steps and predict the loss curve for the WSD LRS with 350K steps. The model has 1.7 billion parameters, and the training uses 1,400 billion tokens. More detailed experimental setup is shown as the setting $E$ in Table~\ref{tab:exp-sets}. 

\begin{figure}[p]
    \centering
    \begin{subfigure}[b]{0.32\textwidth}
        \includegraphics[width=\textwidth]{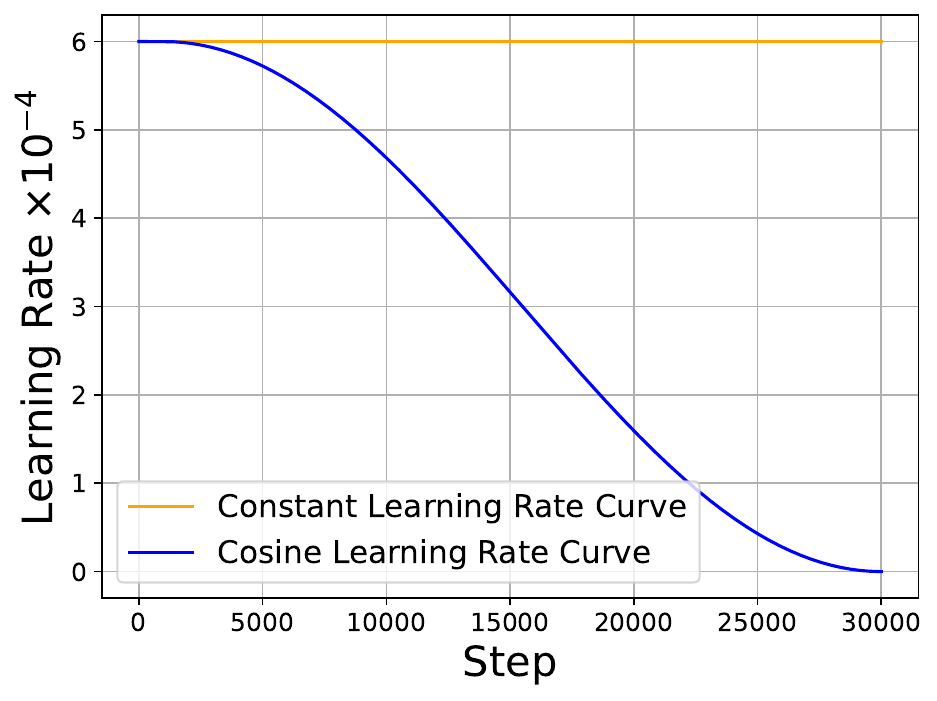}
        \caption{Learning rate.}
    \end{subfigure}
    \hfill
    \begin{subfigure}[b]{0.32\textwidth}
        \includegraphics[width=\textwidth]{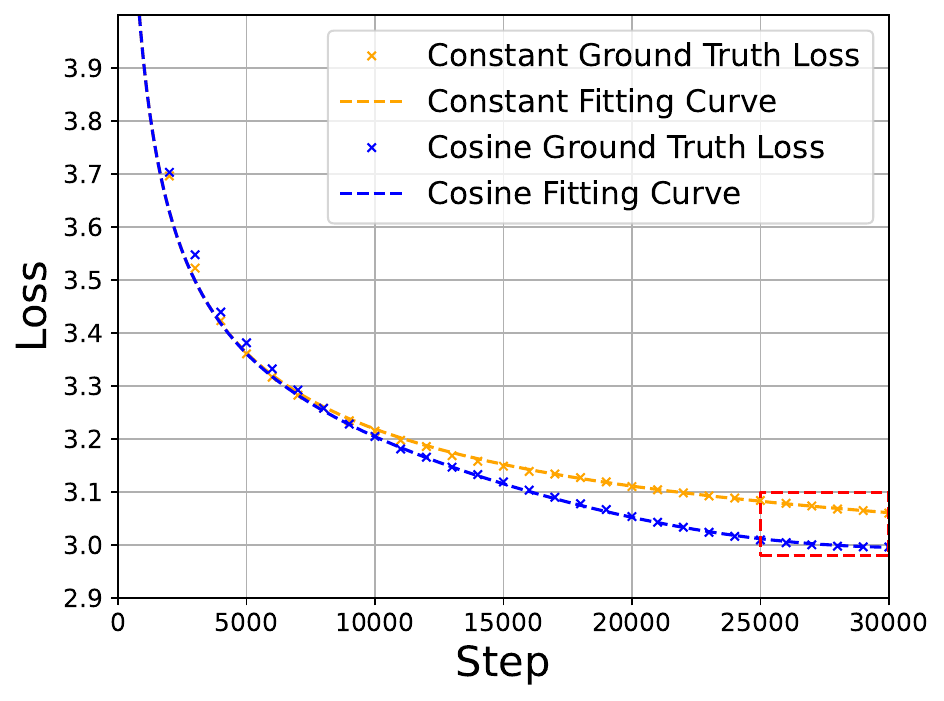}
        \caption{Zoomed-out loss fitting.}
    \end{subfigure}
    \hfill
    \begin{subfigure}[b]{0.32\textwidth}
        \includegraphics[width=\textwidth]{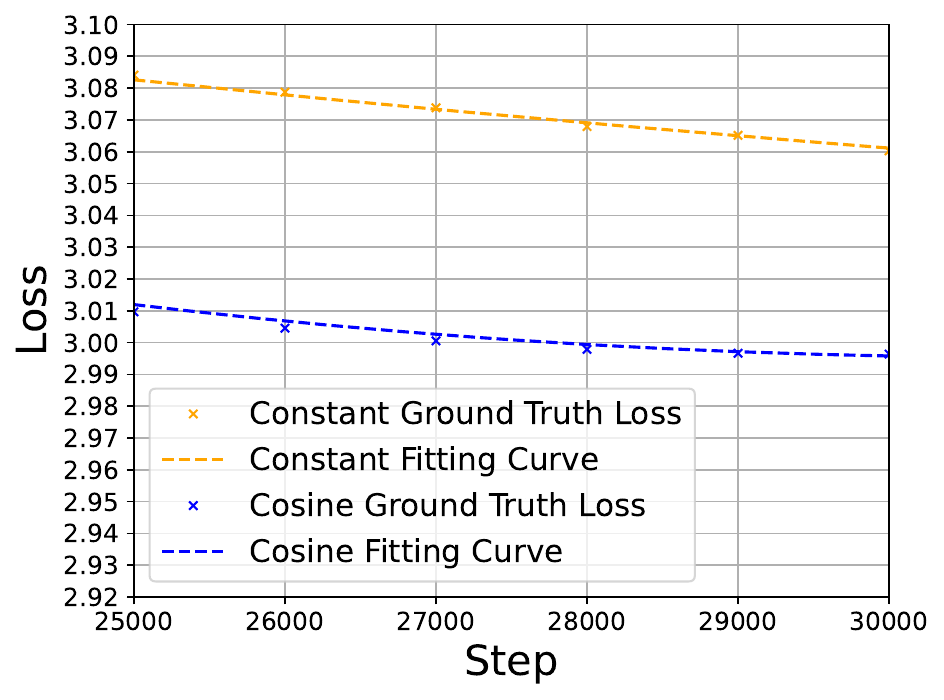}
        \caption{Zoomed-in loss fitting.}
    \end{subfigure}
    \caption{Full loss curve \textbf{fitting} for the constant and cosin LRS on 30K steps. After fitting, we get a \textbf{universal} loss equation $L = 2.788 + 0.906\cdot S_1^{-0.416} - 0.254\cdot S_2$. Refer to setting E in Table~\ref{tab:exp-sets} for detailed experimental setups. }
\label{fig:35_fit}
\end{figure}

\begin{figure}[tbp]
    \centering
    \begin{subfigure}[b]{0.32\textwidth}
        \includegraphics[width=\textwidth]{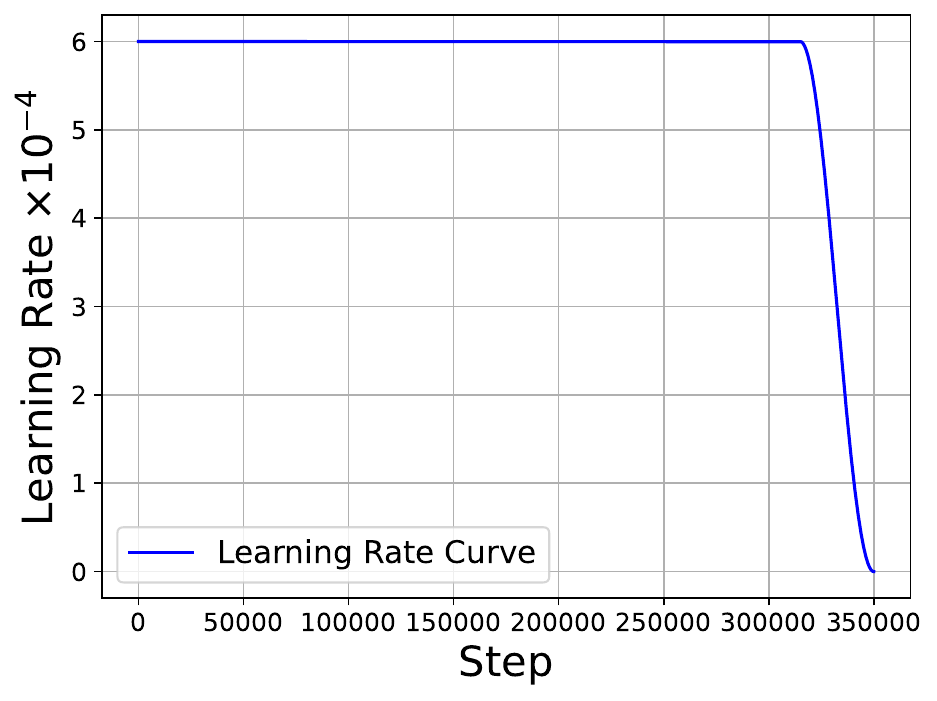}
        \caption{Learning rate.}
    \end{subfigure}
    \hfill
    \begin{subfigure}[b]{0.32\textwidth}
        \includegraphics[width=\textwidth]{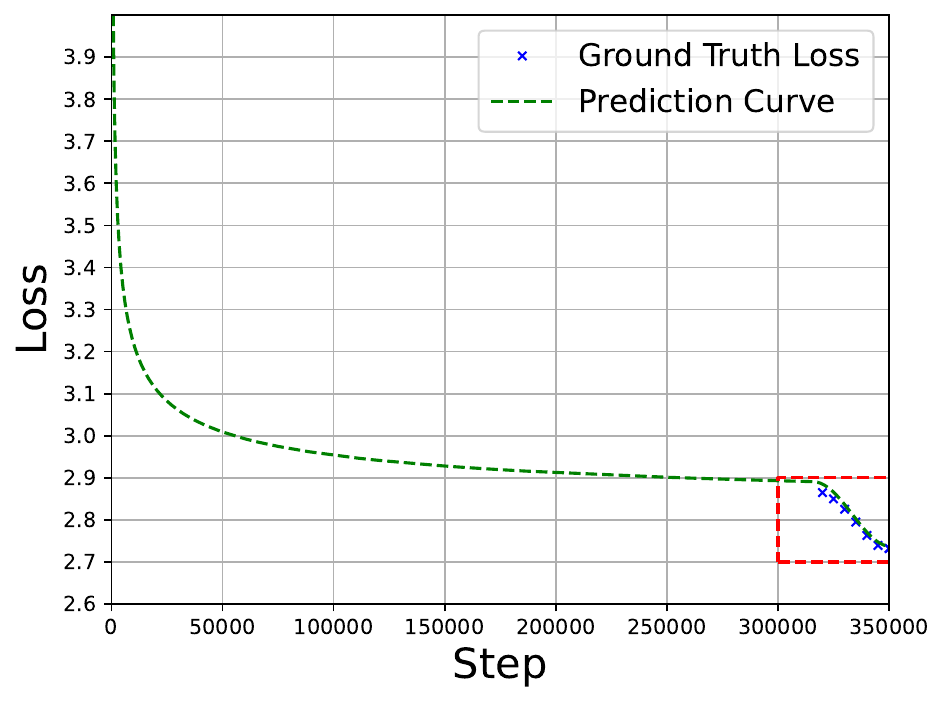}
        \caption{Zoomed-out loss prediction.}
    \end{subfigure}
    \hfill
    \begin{subfigure}[b]{0.32\textwidth}
        \includegraphics[width=\textwidth]{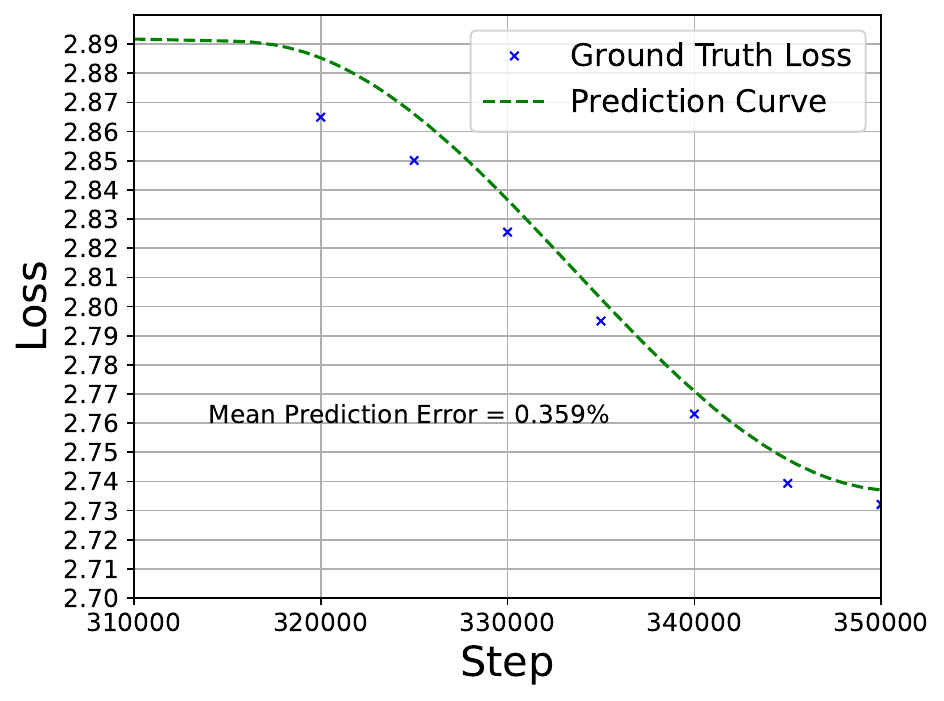}
        \caption{Zoomed-in loss prediction.}
    \end{subfigure}
    \caption{Using the fitted equation in Fig.~\ref{fig:35_fit} to \textbf{predict} the full loss curve equation under the WSD LRS (10\% cosine annealing ratio to $\eta_{min}=0$). Our equation accurately predict the loss curve in the annealing stage after the 10x longer steps. Refer to setting E in Table~\ref{tab:exp-sets} for detailed experimental setups.}
\label{fig:35_predict}
\end{figure}

The fitting and prediction results are shown in Fig.~\ref{fig:35_fit} and Fig.~\ref{fig:35_predict} respectively. It shows that we accurately predict the loss curve for the annealing stage after 10x longer steps in advance.

\subsection{Open-sourced Models}
\label{apx:open-source}
\begin{figure}[tbp]
    \centering
    \begin{subfigure}[b]{0.48\textwidth}
        \includegraphics[width=\textwidth]{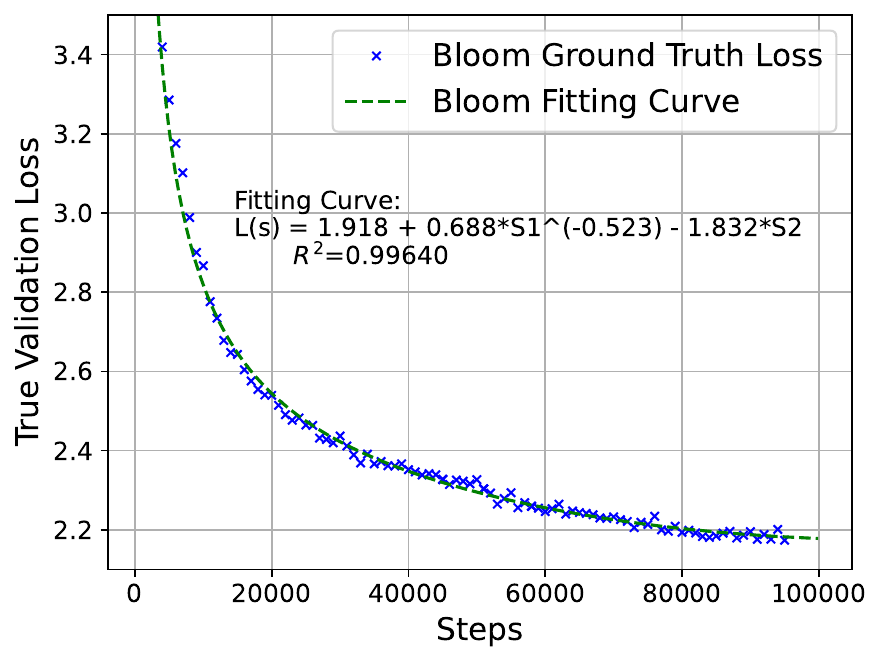}
        \caption{Full loss curve fitting on BLOOM-176B.}
        \label{fig:bloom}
    \end{subfigure}
    \hfill
    \begin{subfigure}[b]{0.48\textwidth}
        \includegraphics[width=\textwidth]{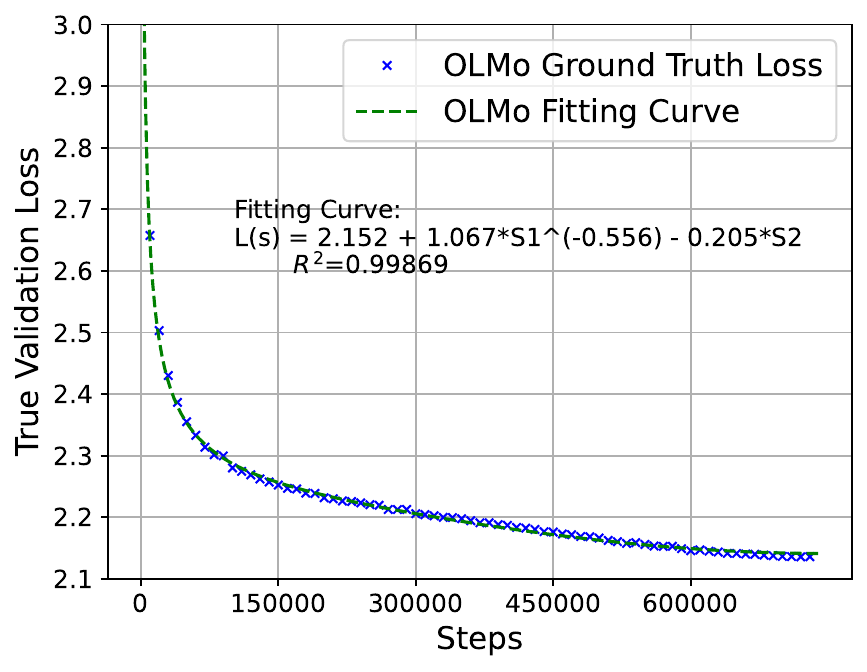}
        \caption{Full loss curve fitting on OLMo-1B (2T tokens).}
        \label{fig:olmo}
    \end{subfigure}
    \caption{Fitting loss curves for open sourced models. We extract the curve of BLOOM from \url{https://huggingface.co/bigscience/bloom/tensorboard}, and choose the column \texttt{lm-loss-validation/valid/lm loss validation} as the validation loss.
    We extract the curve of OLMo from {\url{https://wandb.ai/ai2-llm/OLMo-1B?nw=nwuserdirkgr}, and choose the column \texttt{eval/pile/CrossEntropyLoss} as the validation loss.
    Both models adopt cosine LRS.}
    }
\label{fig:open-source-fit}
\end{figure}

To further validate our proposed scaling law, we apply our equation on modeling the loss curves of open-sourced language models, including BLOOM-176B~\citep{workshop2022bloom} and OLMo-1B~\citep{groeneveld2024olmo}. As shown in Fig.~\ref{fig:open-source-fit}, our equation fits very well on loss curves of open-sourced models, even when the model size scales up to 176B (e.g. BLOOM) and token number scales up to 2000B over 740K steps (e.g. OLMo). 

\subsection{Our $N$-extended Scaling Law on Another Experiments Setups}
\label{apx:exp-2-N}
\begin{figure}[tbp]
\centering
\includegraphics[width=0.7\textwidth]{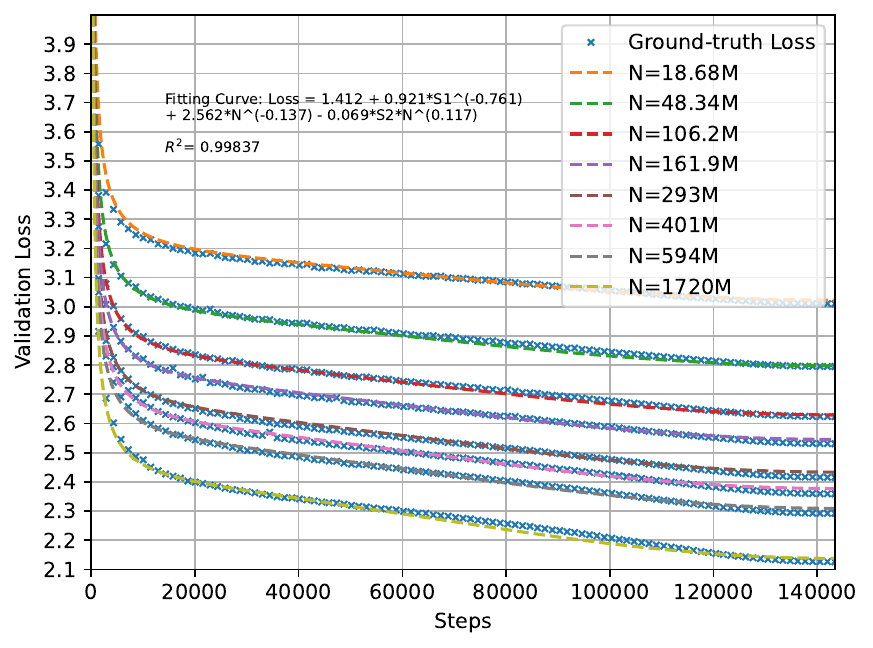}
\caption{Curve fitting on cosine LRS (143K steps to $\eta_{min}=0$) for various different model sizes using our scaling law extended to model size $N$. Refer to setting C in Table~\ref{tab:exp-sets} for detailed experimental setups.}
\label{fig:extend-N-apx}
\end{figure}

Fig.~\ref{fig:scaling-N-fit} shows that our $N$-extended equation (Eq.~\ref{eq:scaling-N}) works well on our main experimental setup.
Similarly, to validate the effectiveness of our $N$-extended scaling equation on various different experimental settings, we change our setup from setting $A$ to setting $C$ (refer to Table~\ref{tab:exp-sets}). The fitting results are shown in Fig.~\ref{fig:extend-N-apx}. The results suggest that our $N$-extended scaling law with LR annealing works well across different experimental setups.

\section{WSD Scheduler and Annealing Functions}

\begin{figure}[tbp]
    \centering
    \begin{subfigure}[b]{0.48\textwidth}
        \includegraphics[width=\textwidth]{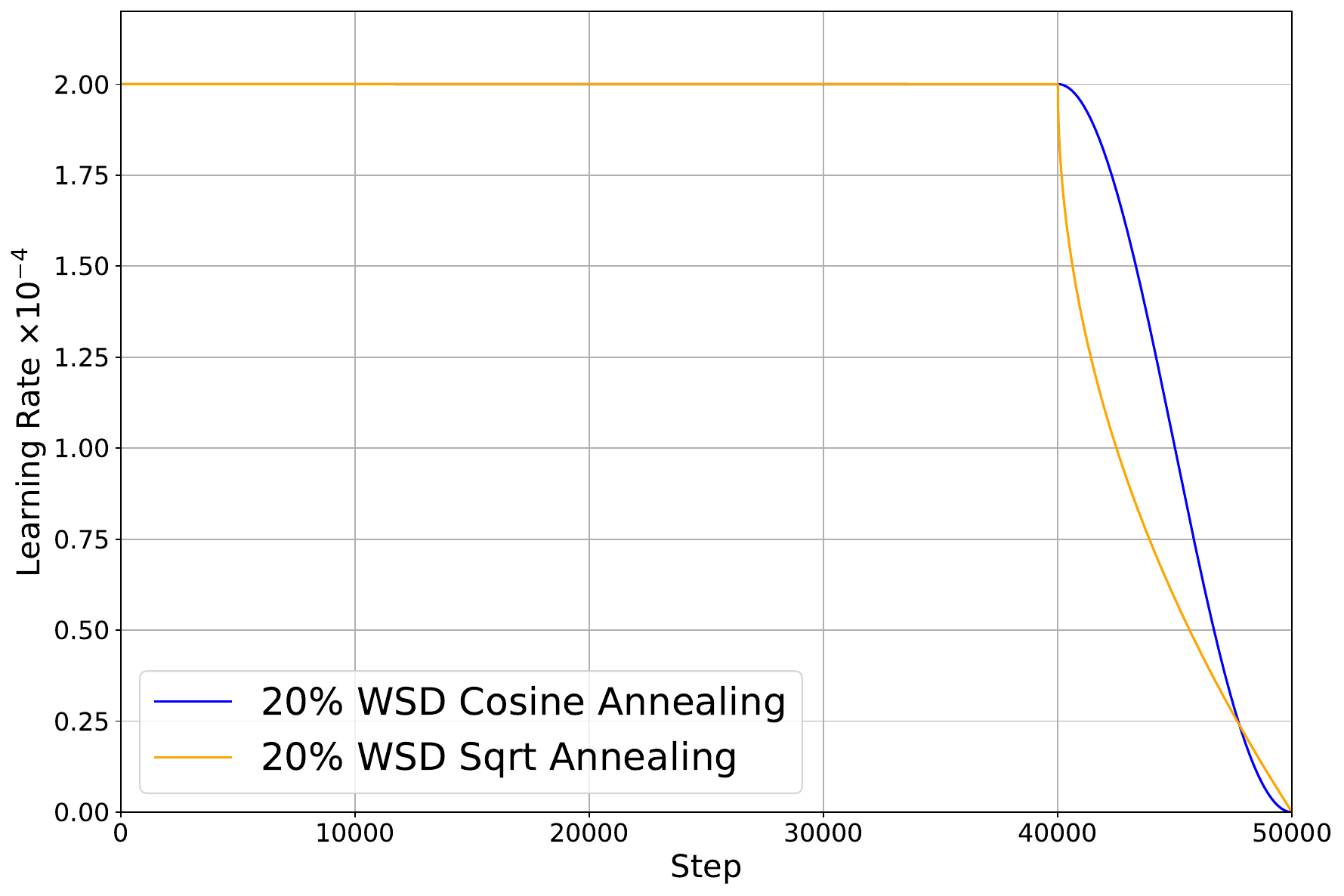}
        \caption{LR curve of WSD (20\% 1-sqrt/cosine annealing).}
    \end{subfigure}
    \hfill
    \begin{subfigure}[b]{0.48\textwidth}
        \includegraphics[width=\textwidth]{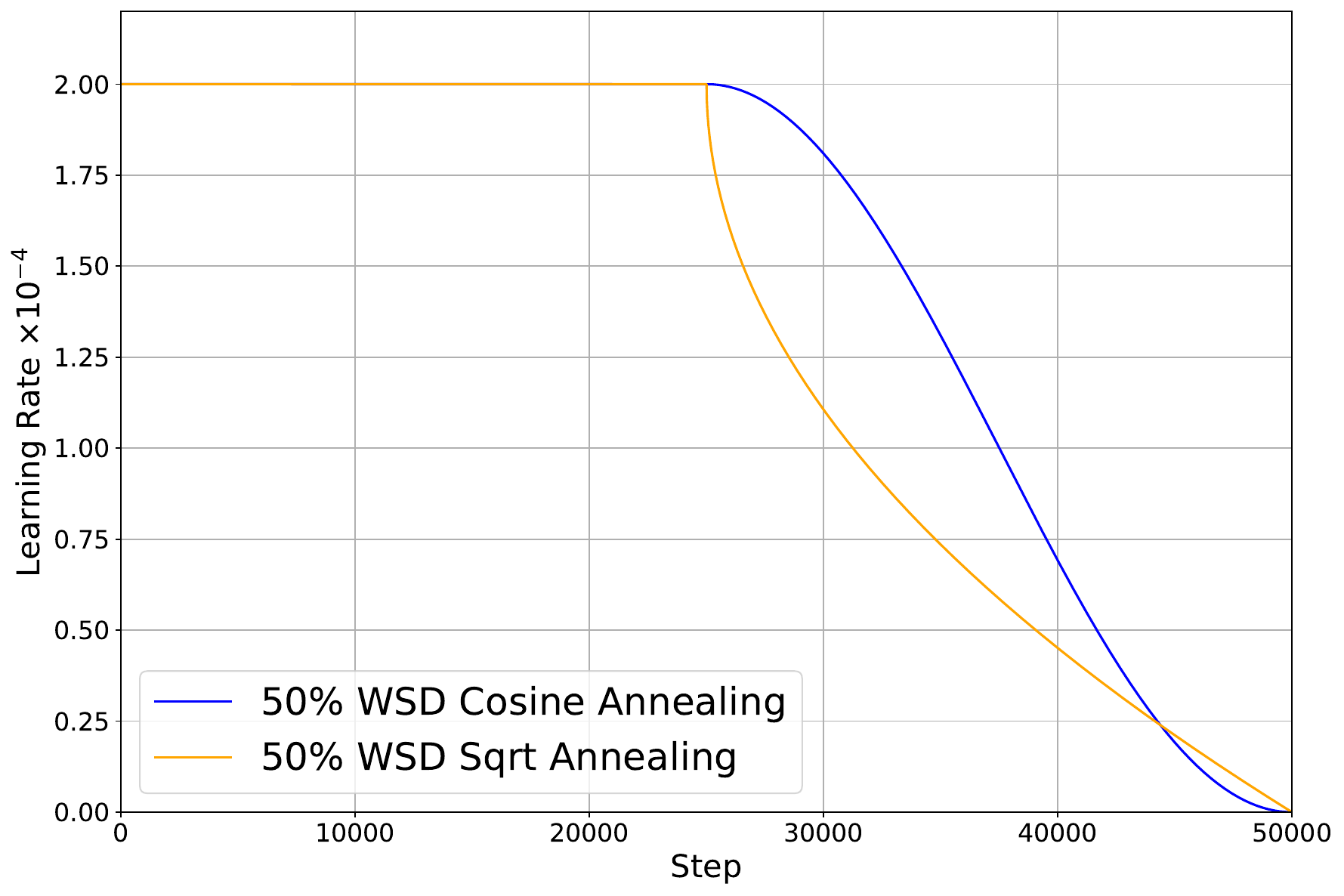}
        \caption{LR curve of WSD (50\% 1-sqrt/cosine annealing).}
    \end{subfigure}
    \caption{The learning rate curves of 20\% (left) and 50\% (right) annealing ratio in WSD LRS, with cosine and 1-sqrt annealing method.}
\label{fig:lr_1-sqrt}
\end{figure}

\citet{hu2024minicpm} proposes a warmup-stable-decay (WSD) LRS including three learning rate stages, which could help get a lower validation loss compared to the typical cosine LRS. The format is like 
\begin{equation}
\label{eq:wsd}
\begin{aligned}
WSD(s)=\begin{cases}&
\frac{s}{T_{\text {warmup}}}\eta_{max},\quad s\leq T_{\text {warmup}}\\&
\eta_{max},\quad T_{\text {warmup}}<s\leq T_{\text {stable}}\\&
\eta_{min} + f(s)\cdot(\eta_{max}-\eta_{min}),\quad T_{\text {stable}}<s\leq T_{\text {total}}
\end{cases}
\end{aligned}
\end{equation}
Where $0 \leq f(s) \leq 1$ is typically a decreasing function about step $s$, and $\eta_{max}$ is the maximal learning rate.~\citet{hägele2024scaling} consolidate the effectiveness of the WSD scheduler with numerious empirical experiments. Moreover,~\citet{hägele2024scaling} also find that using 1-sqrt annealing and a moderate
annealing ratio (e.g. 20\%) can further decrease the final loss.
\label{apx:1-sqrt}
The 1-sqrt annealing is defined as:
\begin{equation}
\begin{aligned}
&f(s)=1-\sqrt{\frac{s-T_{\text {stable }}}{T_{\text {total}} - T_{\text {stable }}}} \\ 
\end{aligned}
\end{equation}

Also, ~\citet{hägele2024scaling} mention the 1-square annealing method as a baseline, which is defined as:
\begin{equation}
\begin{aligned}
&f(s)=1-\left({\frac{s-T_{\text {stable }}}{T_{\text {total}} - T_{\text {stable }}}}\right)^2
\end{aligned}
\end{equation}

We draw the LR curve of the WSD LRS proposed by \citet{hu2024minicpm} (20\% and 50\% 1-sqrt annealing) along with the WSD LRS with a cosine annealing in Fig.~\ref{fig:lr_1-sqrt}.
\end{document}